\documentclass[preprints,article,accept,pdftex,moreauthors]{Definitions/mdpi} 
\firstpage{1} 
\pubvolume{1}
\issuenum{1}
\articlenumber{0}
\pubyear{2025}
\copyrightyear{2025}
\datereceived{ } 
\daterevised{ } % Comment out if no revised date
\dateaccepted{ } 
\datepublished{ } 
\hreflink{https://doi.org/} % If needed use \linebreak
\Title{Multi-robot LiDAR SLAM: a practical case study in underground tunnel environments}

\TitleCitation{Multi-robot LiDAR SLAM: a practical case study in underground tunnel environments}

\Author{Federica Di Lauro $^{1,2,*}$\orcidA{}, Domenico G. Sorrenti $^{1}$ \orcidB{} and Miguel Angel Sotelo $^{2}$ \orcidC{}}

\AuthorNames{Federica Di Lauro, Domenico G. Sorrenti and Miguel Ángel Sotelo}

\isAPAStyle{%
       \AuthorCitation{Lastname, F., Lastname, F., \& Lastname, F.}
         }{%
        \isChicagoStyle{%
        \AuthorCitation{Lastname, Firstname, Firstname Lastname, and Firstname Lastname.}
        }{
        \AuthorCitation{Di Lauro, F.; Sorrenti, D.G.; Sotelo, M.A.}
        }
}

\address{%
$^{1}$ \quad Università degli Studi di Milano-Bicocca\\
$^{2}$ \quad Universidad de Alcalá}

\corres{Correspondence: f.dilauro2@campus.unimib.it}

\abstract{
Multi-robot SLAM aims at localizing and building a map with multiple robots, interacting with each other. In the work described in this article, we analyze the pipeline of a decentralized LiDAR SLAM system to study the current limitations of the state of the art, and we discover a significant source of failures, i.e., that the loop detection is the source of too many false positives. We therefore develop and propose a new heuristic to overcome these limitations. The environment taken as reference in this work is the highly challenging case of underground tunnels. We also highlight potential new research areas still under-explored.}

\keyword{SLAM; Point Cloud Registration; LiDAR; Multi-robot SLAM}

\begin{document}

\section{Introduction}

Simultaneous Localization and Mapping (SLAM) is a foundational problem in robotics and autonomous systems, where an agent must construct a map of an unknown environment while concurrently estimating its own pose within that map. SLAM plays a critical role in enabling autonomous navigation and decision-making for mobile robots operating in complex and dynamic environments~\cite{cadena_past_2016}.

While traditional SLAM systems focus on a single robot, there is growing interest in multi-robot SLAM, where multiple robots collaborate to build and/or maintain a shared map, this of course requires the robots to localize themselves. This collaborative approach offers significant advantages, including improved coverage, faster map construction, redundancy, and robustness against individual failures. Multi-robot SLAM algorithms can be classified as centralized approaches when the system relies on a central base station to gather information from the robots and merge it, or decentralized when each robot builds its map by also exchanging information in a peer-to-peer manner~\cite{lajoie_towards_2022}.

Moreover, SLAM in perceptually challenging environments, such as underground tunnels, poses difficulties to current systems. A typical SLAM pipeline includes an incremental part, built around the estimation of the motion of the robot between a sensor activation and the subsequent one (aka odometry), as well as a mechanism for integrating such sensor data. Another component of a typical SLAM pipeline is a module to detect loop closures, i.e., the robot passing again at the same place. Lastly, a module for exploiting the loop closures, in order to temper the unavoidable geometric distorsions arising in the incremental phase. 
Oftentimes, in the case of usage of LiDARs, the incremental phase is called tout-court LiDAR odometry, even though the estimation of the rototranslation between two consecutive activations is just the first of the activities implied, before the fusion of the point clouds

A lot of work has been done with respect to LiDAR odometry estimate in challenging environments, with recent studies mainly focused on the degeneracy of ICP optimization~\cite{tuna_x-icp_2024, tuna_informed_2025}, ICP being the long-live state-of-the-art algorithm for merging, i.e., registering, point clouds (PCs) from LiDAR activations. However, little has been done w.r.t. the task of  place recognition, loop closure and global point cloud registration in the same kind of environments.

The most famous example of a situation where multi-robot SLAM in an underground tunnel scenario is necessary is represented by the DARPA SubT Challenge~\cite{ackerman_robots_2022}. During this event, multiple teams deployed their multi-robot SLAM solution in a tunnel-like scenario similar to the one proposed by us later on in this article.
Ebadi et. al~\cite{ebadi_present_2024} present the infrastructure used by each team in the final DARPA event, highlighting the strengths and weaknesses of each team. Focusing on CERBERUS, the winning team,~\cite{tranzatto_cerberus_2022} we can highlight the following weak points and open problems:

\begin{itemize}
    \item The multi-robot SLAM algorithm used is a centralized one, which can be limiting due to the need of always having communication between each robot and the base station, and that can also be a bottleneck in the case of a large number of robots and high amount of data.
    \item Wrong loop detection hypotheses would completely break the mapping system. For this reason a human operator was manually inspecting map merging and discarding wrong results. This approach however means that the system cannot be deployed in an unmanned way.
    \item It also relied on the information that all the robot started from the same location and could localize themselves in a shared frame, due to the presence of QR codes at the entrance. This last point is a major limitation, w.r.t. real life applications, that applies to the contributions from all teams participating in the DARPA challenge.
\end{itemize}

These issues, and in particular the last two, are still a major limitation in deploying fully unmanned systems. In this work, in order to overcome the first limitation above, we develop a multi-robot pipeline, inspired by Swarm-SLAM~\cite{lajoie_swarm-slam_2024}, which is a decentralized multi SLAM pipeline that relies on pose graph optimization, and test it on a new dataset gathered specifically for this task. In particular, these are our main contributions:

\begin{itemize}
    \item We gather a dataset in a tunnel-like structure, exploiting realistic simulations, thanks to a real mesh, i.e., acquired in real world by using real world instrumentation.
    \item We analyze a standard SLAM pipeline and highlight the current limitations, with respect to the decentralized map merging between robots. We find that, even though outlier rejection and robust optimization have evolved and made SLAM less affected by outliers, they seem to be not enough to guarantee sufficient robustness. The road is still open to explore new tasks, such as keyframe selection and identification of the alignability of a pair of point clouds.
    \item We propose a simple yet effective method to partially overcome the limitations presented above, and show how carefully selected keyframes greatly reduce the chances of catastrophic failures.
\end{itemize}

By examining the critical challenges of multi-robot LiDAR SLAM and highlighting unresolved questions, this work provides a new direction for achieving practical effectiveness in real-world, complex environments.

\section{Methods}

\subsection{Dataset description}

We propose a new dataset acquired in underground conditions, with multiple trajectories recorded and with robots starting from different entrances, to overcome one significant limitation of the DARPA SubT dataset, that is the single entrance for all robots.

\begin{figure}
    \centering
    \includegraphics[width=0.3\linewidth]{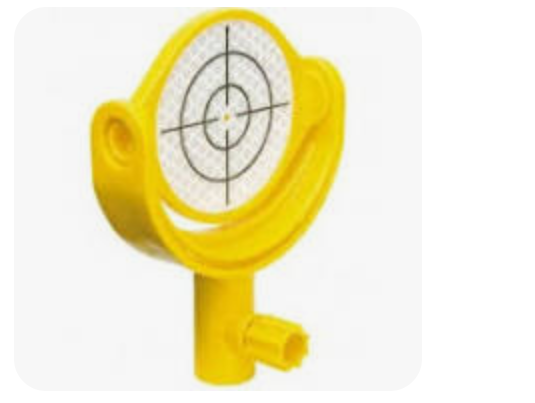}
    \caption{Example of a bireflex target, used for aligning the scans to generate the mesh environment}
    \label{fig:bireflex}
\end{figure}

The simulated environment is based on a high-fidelity mesh reconstructed from real LiDAR scans of the Zentrum am Berg underground infrastructure research facility\footnote{https://www.zab.at/en/about-us/the-facility}.
The data gathered consists of 261 single point clouds, acquired with a RIEGL VZ-400i scanner, providing a ranging accuracy of 5~mm.
Since the standard registration of point clouds of a tunnel does not really work well, because the single point clouds do not show much difference since they more or less represent always a tube, the registration have been eased by adding constraints based on the usage of ad-hoc mounted bireflex targets, show in in fig \ref{fig:bireflex}. These targets, widely used in geomatic surveys in auto-similar conditions like those in underground tunnels, produce a very distinguishable profile in the intensity channel of the point cloud, and this information is then used to help to align the point clouds. The mesh used for the simulations has been obtained from such well-aligned point clouds.

\begin{figure}[ht]
    \centering
    \includegraphics[width=\linewidth]{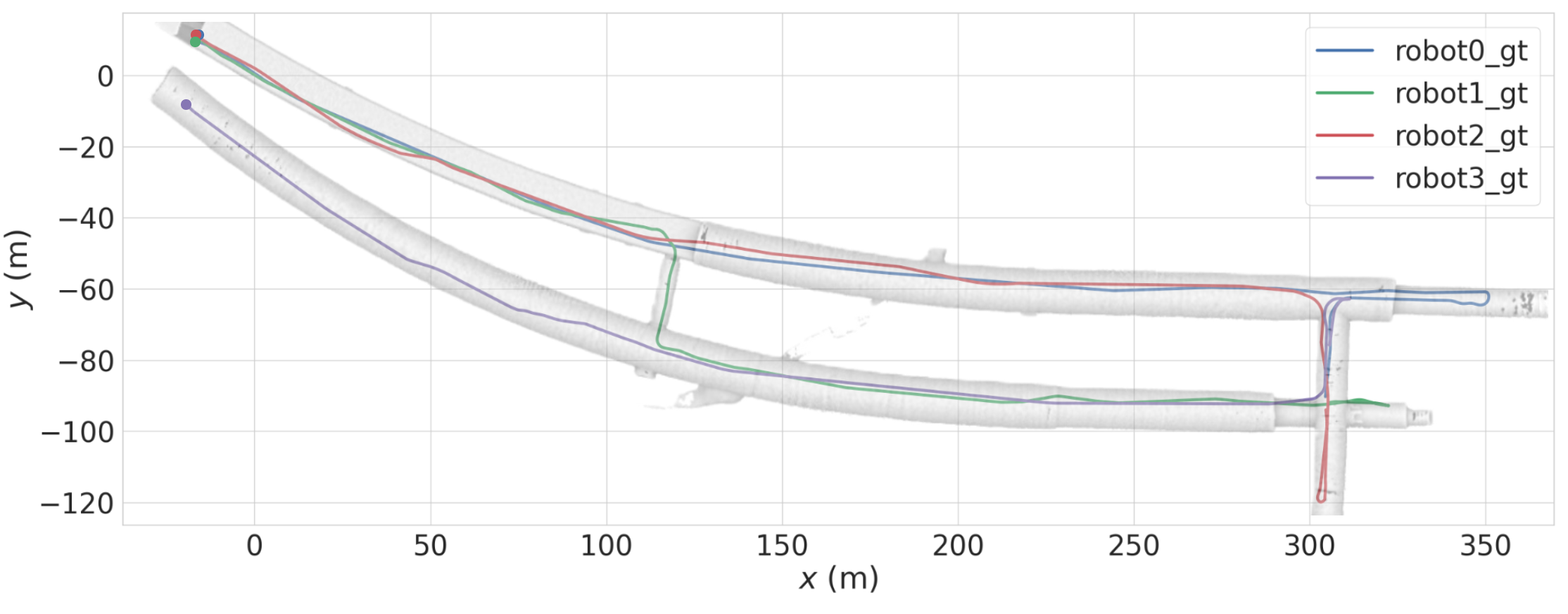}
    \caption{Ground truth trajectories of the four robots, displayed over the mesh of the environment. Starting points shown with colored a dot.}
    \label{fig:gt_traj}
\end{figure}

The datasets used in our experiments were generated simulating four-wheeled robots equipped with Ouster OS2-128 LiDAR sensor, operating within the environment described above. 
To emulate a multi-robot SLAM scenario, four distinct robot trajectories within this environment have been recorded. Both the environment model and the corresponding ground-truth trajectories are illustrated in Figure~\ref{fig:gt_traj}.

\begin{table}[ht]
    \centering
    \begin{tabular}{cccc}
        \toprule
        Robot & Path Length (m) & Time (s) & Avg Speed (m/s) \\
        \midrule
        0 & 453.31 & 238.65 & 1.89\\
        1 & 405.34 & 219.85 & 1.84\\
        2 & 416.52 & 222.59 & 1.87\\
        3 & 390.73 & 210.45 & 1.85\\
        \bottomrule
    \end{tabular}
    \caption{Total path length, trajectory duration, and average speed for each robot. The mesh of the floor appears to be quite smooth, so that we could set such high speeds.}
    \label{tab:gt_trajectories}
\end{table}

\subsection{Single robot LiDAR Odometry}

For odometry estimation, we compare  KISS-ICP~\cite{vizzo_kiss-icp_2023} and Kinematic-ICP~\cite{guadagnino_kinematic-icp}.
KISS-ICP was chosen for the small set of parameters to be tuned and its high performances on a big set of diverse datasets, obtained without any ad-hoc tuning. Kinematic ICP was chosen for its robustness in handling degenerate geometric configurations and again for the simplicity of the parameter tuning. 

Concerning the estimate of the rototranslation between consecutive LiDAR activations, KISS-ICP only uses the LiDAR scans to estimate it, while Kinematic ICP employs LiDAR Scans combined with wheel odometry, to be more robust in degenerate cases.

\subsection{Multi-robot SLAM implementation}

We developed a custom multi-robot SLAM pipeline, tailored for LiDAR-based perception, using well-established concepts and software in the literature and public domain.

Using the LiDAR odometry estimate, we generate a so-called ''KeyFrame`` each time the robot has moved at least 0.5 meters, i.e., the ''KeyFrame`` is the pose estimated and the corresponding point cloud at that time. This approach, with different distance thresholds for keyframe generation, is the most common approach in LiDAR SLAM systems, such as in LIO-SAM~\cite{LIO-SAM} and hdl\_graph\_slam~\cite{hdl_slam}, together with approaches that use instead a fixed time threshold between LiDAR scans for generating keyframes, such as LEGO-LOAM~\cite{LEGO-LOAM}.

To facilitate loop closure detection, which in a multi-robot case means it has to take place also across robots, we compute a global descriptor for each keyframe. Due to the low generalization capabilities of DNN-based descriptors for point clouds, which was demonstrated both by Fontana et. al \cite{fontana_assessing_2025} and Seo et al. \cite{seo_buffer-x_2025} on the point cloud registration task, we decided to use the well-known handcrafted Scan Context~\cite{kim_scan_2018} descriptor. 
For every keyframe from a robot, we search for the most similar descriptor among the keyframes of a second robot, discarding candidates whose similarity scores fall below a predefined threshold. When a promising match is found, we estimate the relative transformation between the corresponding point clouds using the global registration method KISS-Matcher~\cite{lim_kiss-matcher_2024}, a fast and robust approach based on matching FPFH features, and a graph-based outlier rejection scheme, which demonstrates both fast and accurate results.

The back-end of the SLAM system is built using GTSAM~\cite{gtsam}, which provides both the pose graph representation and its optimization, and is heavily inspired from Swarm-SLAM~\cite{lajoie_swarm-slam_2024}. In this approach each robot builds its own pose graph using the keyframes previously generated, with each keyframe representing a node of the graph, and the LiDAR odometry measurement between a node and the previous one representing an edge. We then add inter-robot edges, that is edges that connect nodes from different robots, using the global registration approach described above. In this way, each robot is able to independently perform the loop detection and closure, and is also able to exploit the information that was gathered by other robots.

To ensure robustness in the presence of incorrect loop closures, the main problem that is affecting current multi-robot SLAM systems, we incorporate two complementary techniques: (i) the Pairwise Consistency Measurement (PCM) \cite{mangelson_pairwise_2018} is used to pre-filter inconsistent inter-robot loop closures before optimization: this method maximizes the consistency between the single robot odometry trajectories and the loops found in the inter robot matches, by representing the consistent measurements as edges of a graph and then finding the maximum clique; and (ii) the Graduated Non-Convexity (GNC) optimizer \cite{yang_graduated_2020}, which is applied as a final step to improve robustness during the pose graph optimization process: GNC solves non-convex optimization problem by starting with a convex approximation, gradually increasing its non-convexity, and using each solution as the starting point for the next step until the original problem is reached, thus avoiding local minima. For what concerns the GNC optimization, we mark the odometry edges as inliers, so that they will not be affected by the GNC optimization.

To investigate the impact of uninformative data on loop detection and closure, and specifically for our case of ``tunnel'' structures, we introduce a simple filtering heuristic aimed at excluding keyframes that bring in a small amount of new information. Our proposed heuristic is based on computing the oriented bounding box of each keyframe point cloud; PCA is used to compute the orientation of the box~\cite{Zhou2018_open3d}. We then inspect the second-largest dimension of the bounding box. In order to detect, and discard, continuous tunnel-like keyframes, we check whether the second largest dimension of the bounding box, likely the width of the tunnel in a tunnel-only scenario, falls below a predefined threshold, tuned according to knowledge of the environment. If the condition is true, the keyframe is presumed to originate from a tunnel-only structure, with limited structural variation like, e.g., an intersection between tunnels, and is excluded from the insertion into the graph as well as from the subsequent matching and processing.

While this approach is relatively coarse and environment-dependent, it provides an initial means to explore how redundant or uninformative loop detection observations influence the performance of multi-robot SLAM, which is the objective of this work.

\section{Discussion}

\subsection{Single robot LiDAR Odometry}

The SLAM graph is first constructed using the LiDAR odometry from each robot, to obtain an incremental map estimate that will be then used used for loop detection and closure. We compare the two approaches for LiDAR odometry presented above, to show that while KISS-ICP might be very good in usual and geometrically informative environments, it can fail in degenerate situations. Instead, the use of the data coming from the robots' movements introduces constraints on the optimization process and allows Kinematic ICP to work even in tunnel like environments.

\begin{figure}[H]
    \centering
    \includegraphics[width=\linewidth]{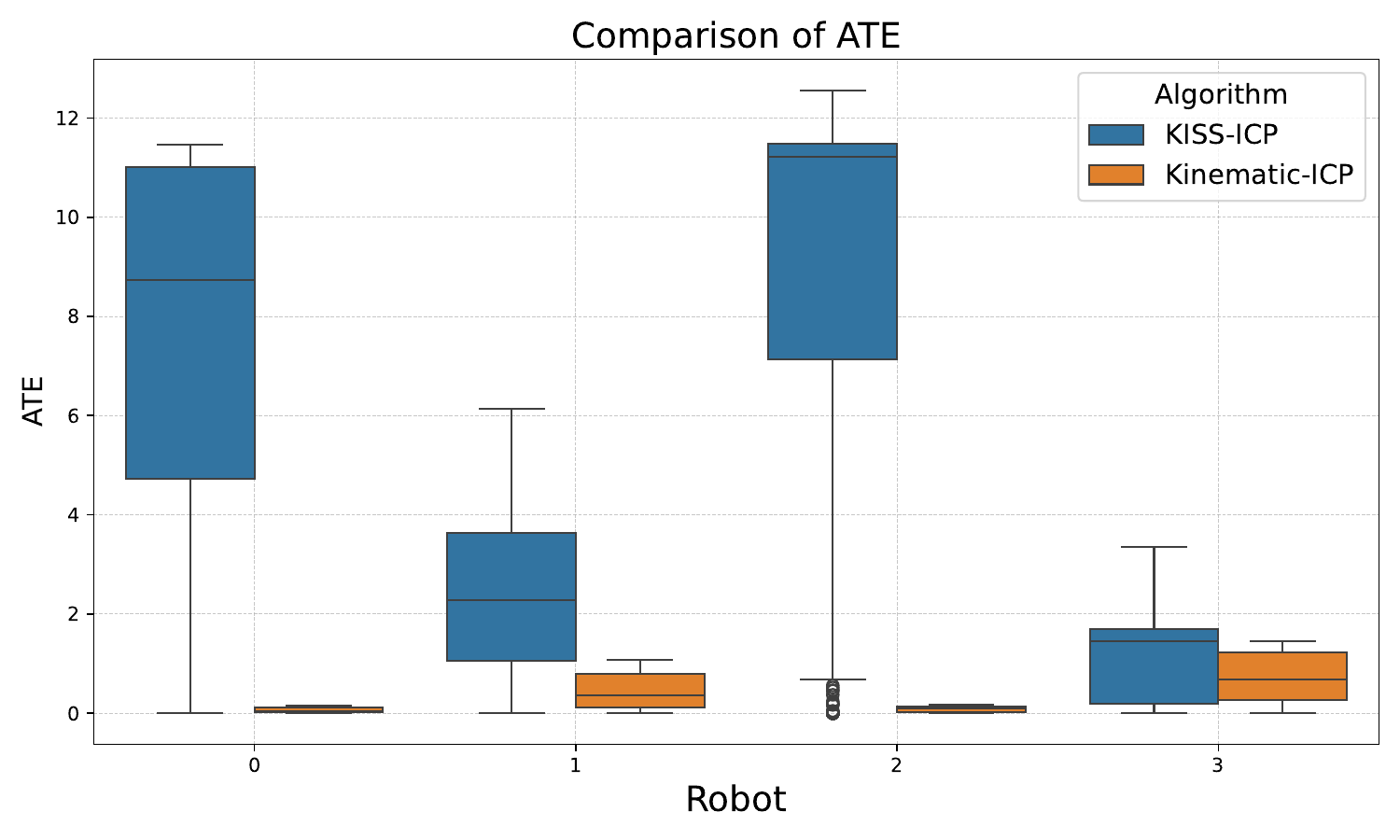}
    \caption{Comparison of Absolute Trajectory error for KISS-ICP and Kinematic ICP. On the x axis the 4 trajectories, on the y axis the ATE, in meters. The error is represented as a standard boxplot with the median error shown as line in the middle of the box.}
    \label{fig:lidar_odometry_comp}
\end{figure}

The results, for which we use the Absolute Error Trajectory, i.e. the sum of the translation errors between the Ground Truth poses and the estimated poses, show that KISS-ICP is not resilient against degenerate environments, while Kinematic ICP is robust and can effectively estimate the trajectory, as shown in fig. \ref{fig:lidar_odometry_comp}.

\subsection{Multi-robot map merging}

Here we present the results of the map merging. Given the distributed approach that we are following, we use the term "map merging" because it integrates the loop detections of each robot altogether with the actual merging across the maps built by each robot. We test the map merging for each robot pair, resulting in 6 different configurations, obtained by varying the robots involved in the pair. We also analyze the impact of using PCM filtering and the use of all keyframes, i.e., those considered by the usual heuristics creating a keyframe every 0.5m motion by a robot, compared with using only tunnel-filtered keyframes.
In tables~\ref{tab:outliers_before_pcm} and \ref{tab:outliers_after_pcm}, we show how our proposed filtering process, which we refer to as ''tunnel-filtered KF`` influences the percentage of outliers present in the final graphs. We can clearly see that our approach eliminates a large percentage of outliers, both wrong loops added because of a wrong place recognition detection, and wrong loops added because the place was correctly recognized but the point cloud registration algorithm failed. We consider the PCR failed if the translation error exceeds 1 meter or the rotation error exceeds 15 degrees.
Moreover, in table~\ref{tab:tunnel_slam_with_tunnel_filter} we show the impact of the reduction in the loop outliers and consider SLAM successful if the overall structure of the final optimized map is consistent with the expected map. Further visual information is provided in Appendix A and Appendix B.

\begin{table}[h!]
\centering
\begin{tabular}{lcccc}
\toprule
\multicolumn{1}{c}{} & 
\multicolumn{2}{c}{All KF} & 
\multicolumn{2}{c}{Tunnel-filtered KF} \\
\cmidrule(lr){2-3} \cmidrule(lr){4-5}
Robot Pair & Outlier PR & Outlier PCR & Outlier PR & Outlier PCR \\
\midrule
(0, 1) & 46.08 & 37.30 & 33.33 & - \\
(0, 2) & 32.34 & 26.59 & 16.66 & 8.33 \\
(0, 3) & 72.16 & - & - & - \\
(1, 2) & 46.70 & 37.24 & 28.57 & - \\
(1, 3) & 46.17 & 3.98 & 50.00 & - \\
(2, 3) & 77.22 & 0.55 & - & 3.84 \\\bottomrule
\end{tabular}
\caption{Percentage of outliers (classified as wrong PR or wrong PCR) among total loop detections for robot pairs, across All and Tunnel-filtered KF cases before PCM}
\label{tab:outliers_before_pcm}
\end{table}

\begin{table}[ht!]
\centering
\begin{tabular}{lcccc}
\toprule
\multicolumn{1}{c}{} & 
\multicolumn{2}{c}{All KF} & 
\multicolumn{2}{c}{Tunnel-filtered KF} \\
\cmidrule(lr){2-3} \cmidrule(lr){4-5}
Robot Pair & Outlier PR & Outlier PCR & Outlier PR & Outlier PCR \\
\midrule
(0, 1) & 9.09 & 9.09 & - & - \\
(0, 2) & 3.31 & 2.1 & - & 3.57 \\
(0, 3) & 14.29 & - & - & - \\
(1, 2) & 13.16 & 10.53 & - & - \\
(1, 3) & 2.84 & - & - & - \\
(2, 3) & 23.08 & - & - & - \\\bottomrule
\end{tabular}
\caption{Percentage of outliers (classified as wrong PR or wrong PCR) among total loop detections for robot pairs, across All and Tunnel-filtered KF cases after PCM filtering.}
\label{tab:outliers_after_pcm}
\end{table}

\begin{table}[ht]
    \centering
    \begin{tabular}{cccc}
        \toprule
        Robot pair & PCM filtering & All KF & Tunnel-filtered KFs \\
        \midrule
        (0, 1) & --         & \textbf{\texttimes} & \textbf{\texttimes} \\
        (0, 2) & --         & \textbf{\texttimes} & \textbf{\texttimes} \\
        (0, 3) & --         & \textbf{\texttimes} & \textbf{\checkmark} \\
        (1, 2) & --         & \textbf{\texttimes} & \textbf{\texttimes} \\
        (1, 3) & --         & \textbf{\texttimes} & \textbf{\texttimes} \\
        (2, 3) & --         & \textbf{\texttimes} & \textbf{\checkmark} \\
        \midrule
        (0, 1) & \checkmark & \textbf{\texttimes} & \textbf{\checkmark} \\
        (0, 2) & \checkmark & \textbf{\checkmark} & \textbf{\texttimes} \\
        (0, 3) & \checkmark & \textbf{\texttimes} & \textbf{\checkmark} \\
        (1, 2) & \checkmark & \textbf{\texttimes} & \textbf{\checkmark} \\
        (1, 3) & \checkmark & \textbf{\checkmark} & \textbf{\checkmark} \\
        (2, 3) & \checkmark & \textbf{\texttimes} & \textbf{\checkmark} \\
        \bottomrule
    \end{tabular}
    \caption{Top rows without PCM filtering, bottom ones with PCM filtering. ``All KFs'' shows the SLAM successes when all keyframes were used for loop detection, ``Tunnel-filtered KFs'' shows the SLAM successes when filtering out tunnel-like point clouds from the loop detection. We define ''success`` if all the robots have a ratio between max ATE and path length less then 1\%}
    \label{tab:tunnel_slam_with_tunnel_filter}
\end{table}

\begin{table}[ht]
    \centering
    \begin{tabular}{cccc}
        \toprule
        Robot pair & ATE 1st robot & ATE 2nd robot \\
        \midrule
        (0, 1) &  0.291 &  0.358 \\
        (0, 2) & 88.106 & 63.850 \\
        (0, 3) &  0.159 &  1.530 \\
        (1, 2) &  0.741 &  0.656 \\
        (1, 3) &  1.073 &  0.918 \\
        (2, 3) &  0.445 &  2.174 \\
        \bottomrule
    \end{tabular}
    \caption{Maximum ATE, in meters, of each robot for the experiments with tunnel-filtered KFs}
    \label{tab:ate_filtered_kfs}
\end{table}

An interesting fact that can be noticed is that un-informative keyframes influence both the number of outliers in the Place Recognition task, due to global perceptual aliasing, and in the Point Cloud Registration task, due to the low geometric structure that makes the alignment fail due to perceptual aliasing in local point cloud patches. By applying our  keyframe tunnel-filtering we retain only the point clouds that are both globally distinct from others and also present enough local distinctiveness to allow the correct estimation of the alignment.

The experimental results indicate that relying solely on a robust optimizer is generally not sufficient to ensure reliable SLAM performance, if no common frame is imposed between the robots, which is a realistic assumption. In these cases, the optimizer struggles to handle the ambiguities and noise introduced by the tunnel keyframes. On the other hand, its performance shows a noticeable improvement when applied in conjunction with keyframe tunnel-filtering, suggesting that even a basic level of data pruning, like the one proposed, can largely enhance robustness in the optimization phase.

The Pairwise Consistency Maximization (PCM) approach demonstrates strong performance overall and is notably more effective than the optimizer alone. However, it still falls short in certain challenging configurations. Specifically, when all keyframes are used without any form of filtering heuristic, PCM achieves successful results in only two out of six test cases. This indicates that while PCM significantly improves robustness, it is not entirely resilient to the negative effects of tunnel-like unfiltered data.

The most promising results are achieved when PCM is combined with tunnel-based keyframe filtering. This configuration leads to a success rate of five out of six, marking a substantial improvement over other methods. The single failure observed in this setup likely stems from the lack of sufficient high-quality loop closures, possibly due to a too aggressive exclusion of informative keyframes. Nonetheless, the overall reliability of this approach highlights the importance of combining robust verification mechanisms with intelligent data selection.

An especially noteworthy observation emerges from the configuration denoted as (0,2) with PCM filtering in Table \ref{tab:tunnel_slam_with_tunnel_filter}. In this case, the SLAM pipeline fails when tunnel-filtering is applied but succeeds when all keyframes are retained. This counterintuitive result underscores a critical limitation of our current keyframe filtering strategy: while intended to improve efficiency and robustness, the filtering method may inadvertently discard keyframes that contain essential spatial information necessary for accurate loop detection. This finding suggests a clear need for more refined and adaptive keyframe selection techniques, capable of preserving structurally significant frames, while discarding keyframes that are not useful for the loop detection.

\section{Conclusions}

This study presented a practical multi-robot LiDAR SLAM system for underground tunnel environments, where traditional localization methods struggle due to severe perceptual aliasing and lack of global positioning. We uncovered new critical challenges that are often overlooked or managed by human intervention or ad-hoc solutions such as having a predefined global reference frame for all robots. In particular, we identified that keyframe selection for loop detection plays a pivotal role in overall system performance, with poor selections leading to unstable or failed optimizations, even with robust optimizers. Furthermore, we observed that certain pairs of point clouds, especially those with minimal geometric variation, are inherently difficult to align reliably, introducing significant errors into the global map. By introducing a simple yet effective keyframe selection strategy, we mitigated some of these issues, achieving consistent results. These findings highlight the need for further research into robust keyframe management and alignment strategies, especially in distributed SLAM systems operating in difficult environments.

\vspace{6pt} 

%%%%%%%%%%%%%%%%%%%%%%%%%%%%%%%%%%%%%%%%%%
%% optional
%\supplementary{The following supporting information can be downloaded at:  \linksupplementary{s1}, Figure S1: title; Table S1: title; Video S1: title.}

%%%%%%%%%%%%%%%%%%%%%%%%%%%%%%%%%%%%%%%%%%
\authorcontributions{``Conceptualization, F.D. D.G.S. and M.S.; methodology, F.D.; software, F.D.; validation, F.D; formal analysis, F.D.; investigation, F.D.; resources, M.S.; data curation, F.D.; writing---original draft preparation, F.D.; writing---review and editing, F.D. D.G.S and M.S.; visualization, F.D.; supervision, D.G.S and M.S.; project administration, M.S.; funding acquisition, M.S. 
All authors have read and agreed to the published version of the manuscript.''}

\funding{This publication is part of the ARTIFICIAL INTELLIGENCE USING QUANTUM MEASURED INFORMATION FOR REALTIME DISTRIBUTED SYSTEMS AT THE EDGE project (PCI2022-134968-2) funded by MCIN/AEI/10.13039/501100011033 and by the European Union “NextGenerationEU”/PRTR, and has also been funded by the A-IQ-Ready project of the KDT-JU.2021 (Joint Undertaking Key Digital Technologies) Program of the European Commission under Grant Agreement: 101096658.}

% \dataavailability{Code and data are available at \url{https://github.com/invett/cslam_ros}}

\acknowledgments{The authors want to thank the team from the University of Leoben for providing the scans of ZaB tunnels and the team from Virtual Vehicles for providing the Gazebo robot simulation and the Zentrum am Berg mesh model.}

\conflictsofinterest{The authors declare no conflicts of interest} 

%%%%%%%%%%%%%%%%%%%%%%%%%%%%%%%%%%%%%%%%%%
%% Optional

%% Only for journal Encyclopedia
%\entrylink{The Link to this entry published on the encyclopedia platform.}

\abbreviations{Abbreviations}{
The following abbreviations are used in this manuscript:
\\

\noindent 
\begin{tabular}{@{}ll}
ATE & Absolute Trajectory Error \\
FPFH & Fast Point Feature Histogram \\
KF & Key Frame \\
PCM & Pair-wise Consistency Maximization \\
PCR & Point Cloud Registration \\
PR & Place Recognition \\
SLAM & Simultaneous Localization and Mapping
\end{tabular}
}

%%%%%%%%%%%%%%%%%%%%%%%%%%%%%%%%%%%%%%%%%%
%% Optional
\appendixtitles{yes} % Leave argument "no" if all appendix headings stay EMPTY (then no dot is printed after "Appendix A"). If the appendix sections contain a heading then change the argument to "yes".
\appendixstart
\appendix
\section[\appendixname~\thesection]{Pose graph visualization}

In this appendix we present some visual representation of the pose graphs constructed and the resulting optimized graph.

Moreover, in each graph we distinguish the inter robot edges in three categories: green for correct loop detection and closure, red for edges that are wrong place recognized and blue for loops that were correctly recognized but where the global registration failed.

\begin{figure}[H]
    \centering

    % Tunnels
    \begin{minipage}[t]{0.32\textwidth}
        \centering
        \includegraphics[width=\textwidth]{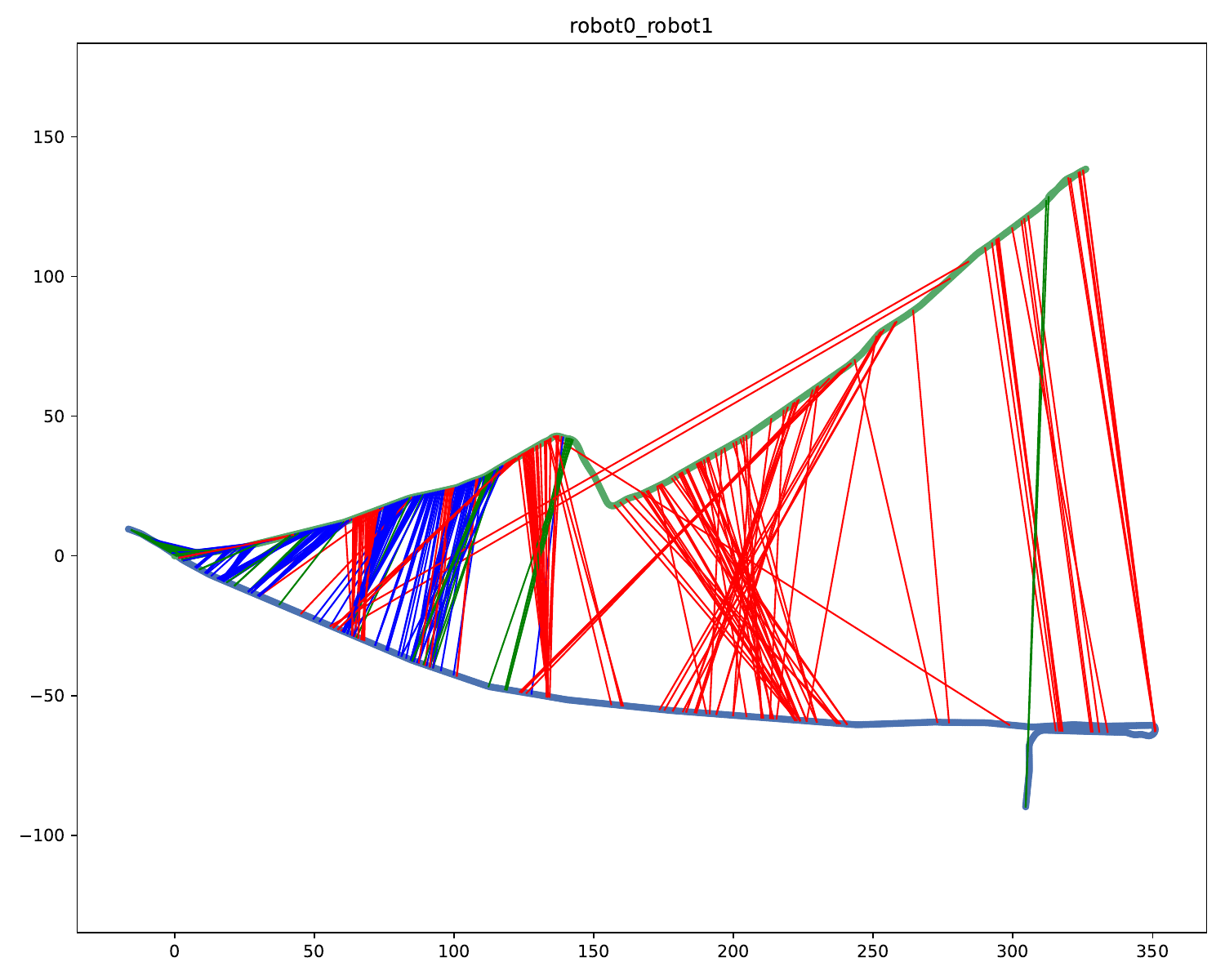}
        \par\smallskip
        \footnotesize\textbf{(a)} Initial g2o graph with all KFs
    \end{minipage}
    \hfill
    \begin{minipage}[t]{0.32\textwidth}
        \centering
        \includegraphics[width=\textwidth]{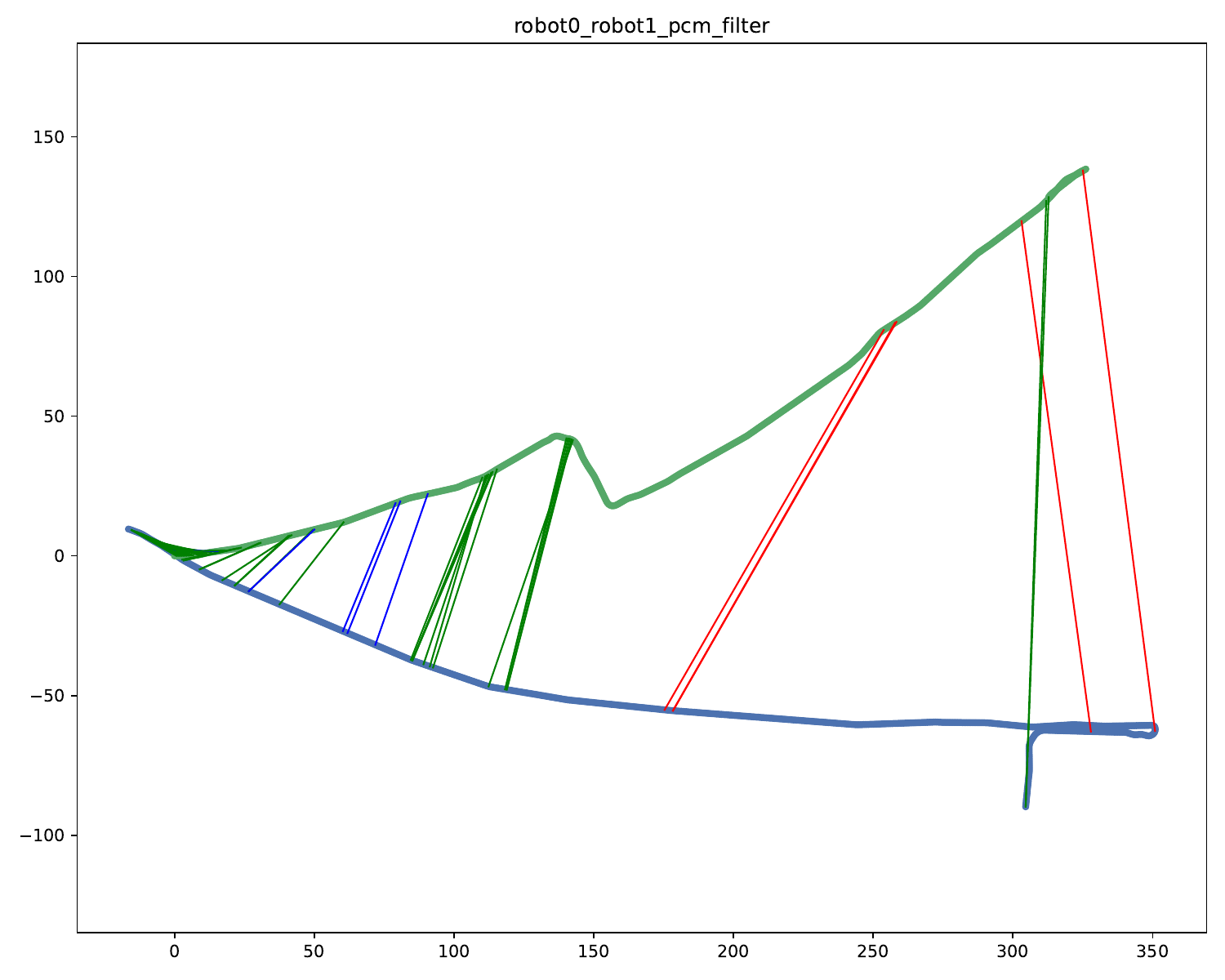}
        \par\smallskip
        \footnotesize\textbf{(b)} g2o graph with all KFs after PCM
    \end{minipage}
    \hfill
    \begin{minipage}[t]{0.32\textwidth}
        \centering
        \includegraphics[width=\textwidth]{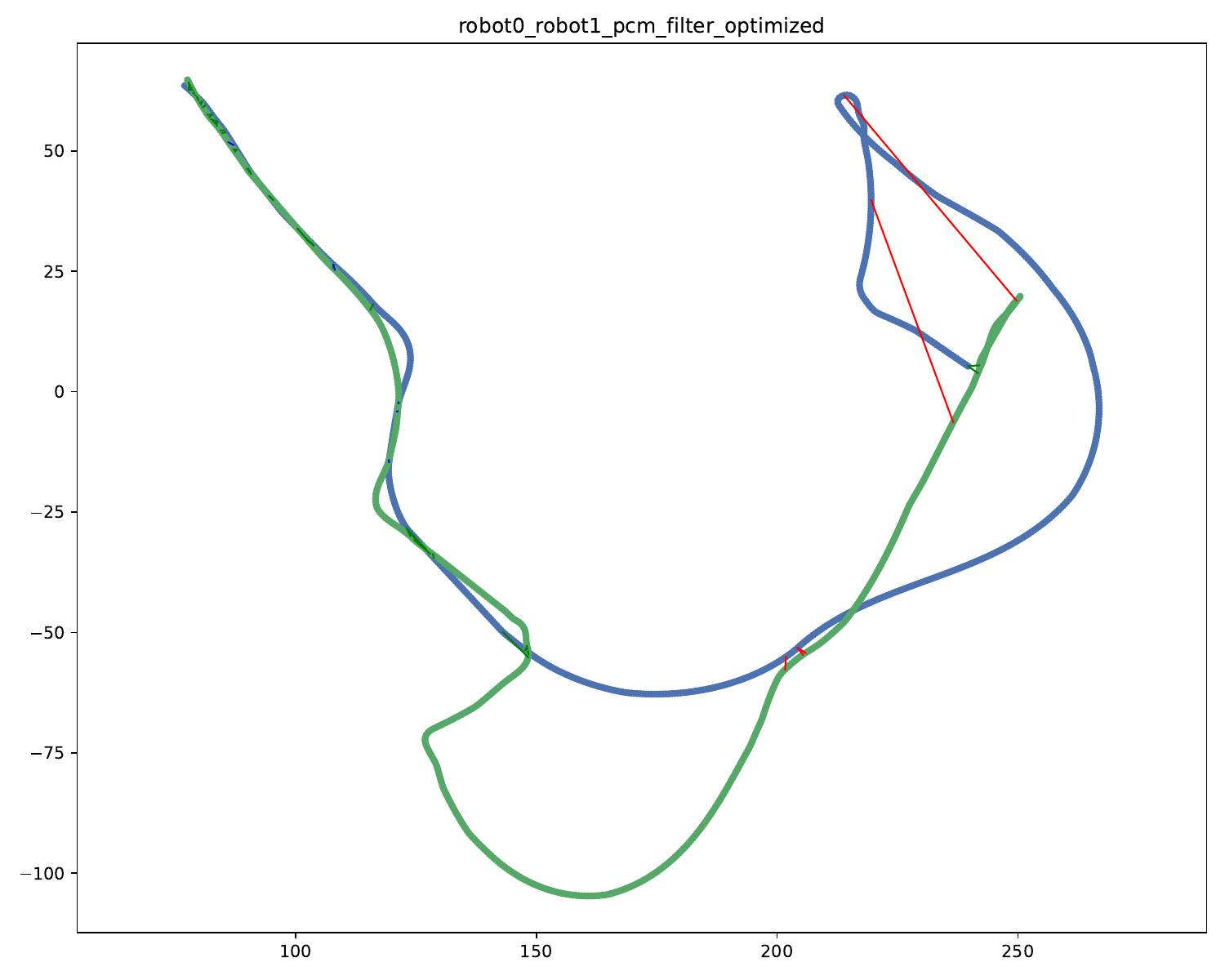}
        \par\smallskip
        \footnotesize\textbf{(c)} g2o graph with all KFs after optimization
    \end{minipage}

    % No tunnels
    \begin{minipage}[t]{0.32\textwidth}
        \centering
        \includegraphics[width=\textwidth]{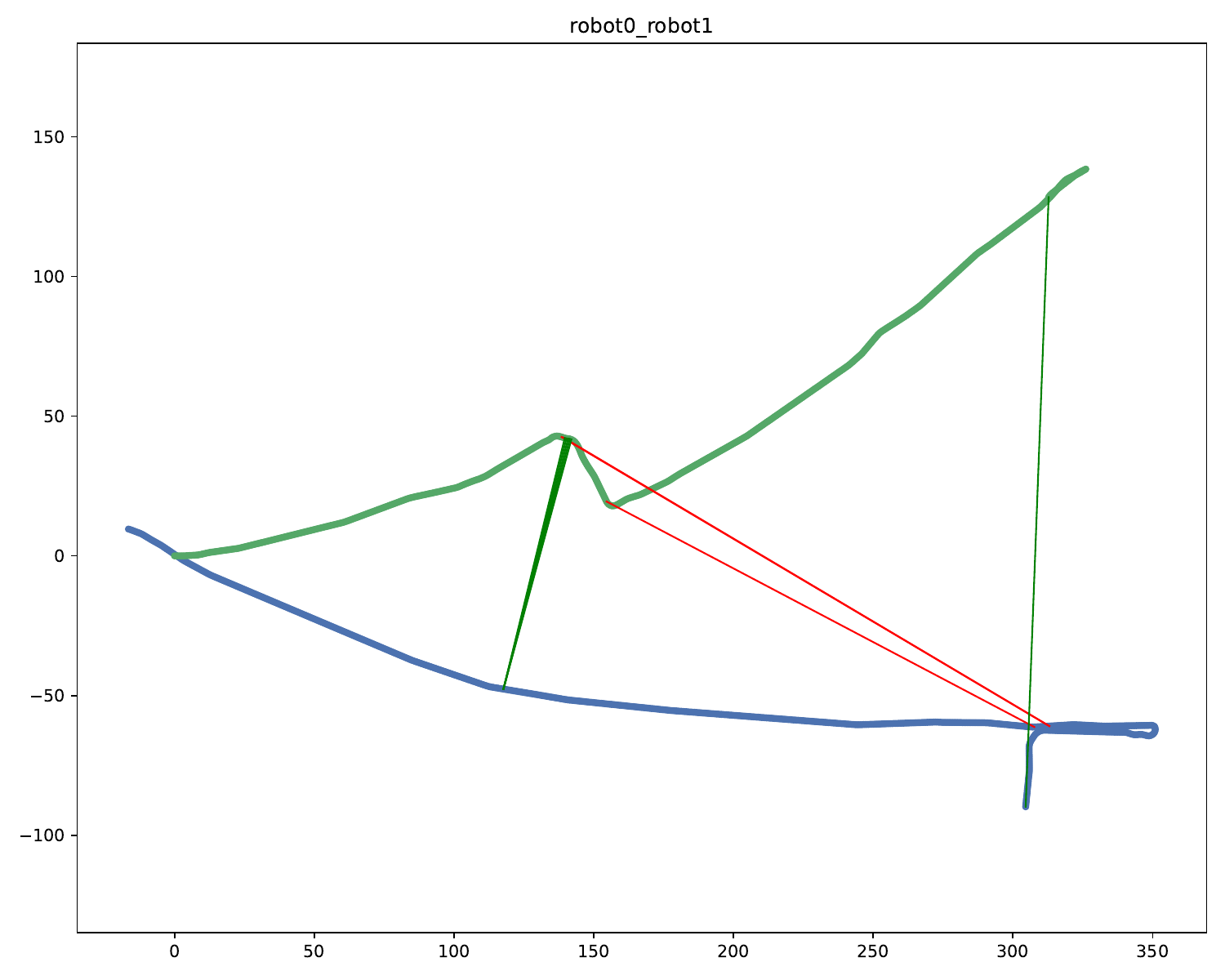}
        \par\smallskip
        \footnotesize\textbf{(d)} Initial g2o graph with tunnel-filtered KFs
    \end{minipage}
    \hfill
    \begin{minipage}[t]{0.32\textwidth}
        \centering
        \includegraphics[width=\textwidth]{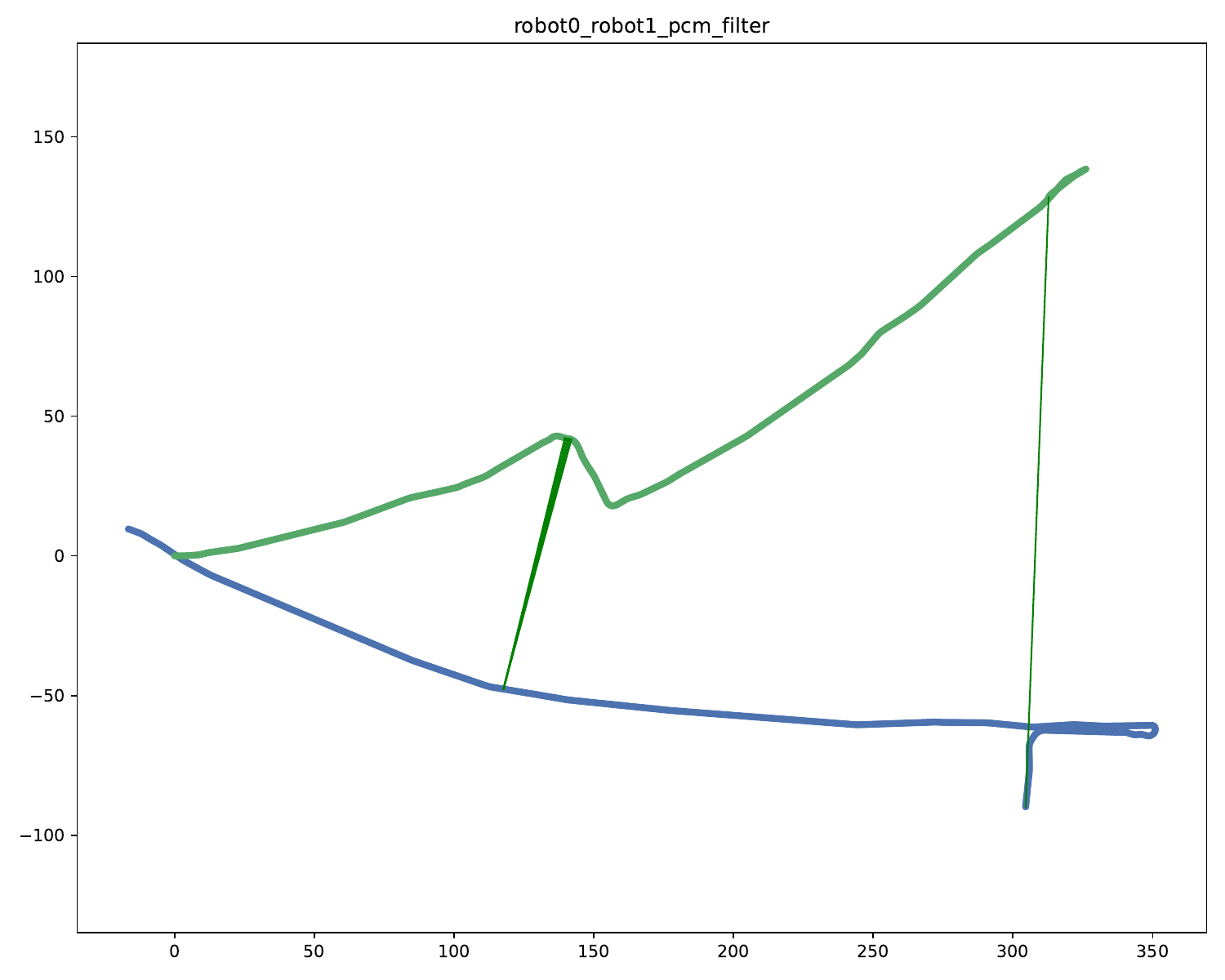}
        \par\smallskip
        \footnotesize\textbf{(e)} g2o graph with tunnel-filtered KFs after PCM
    \end{minipage}
    \hfill
    \begin{minipage}[t]{0.32\textwidth}
        \centering
        \includegraphics[width=\textwidth]{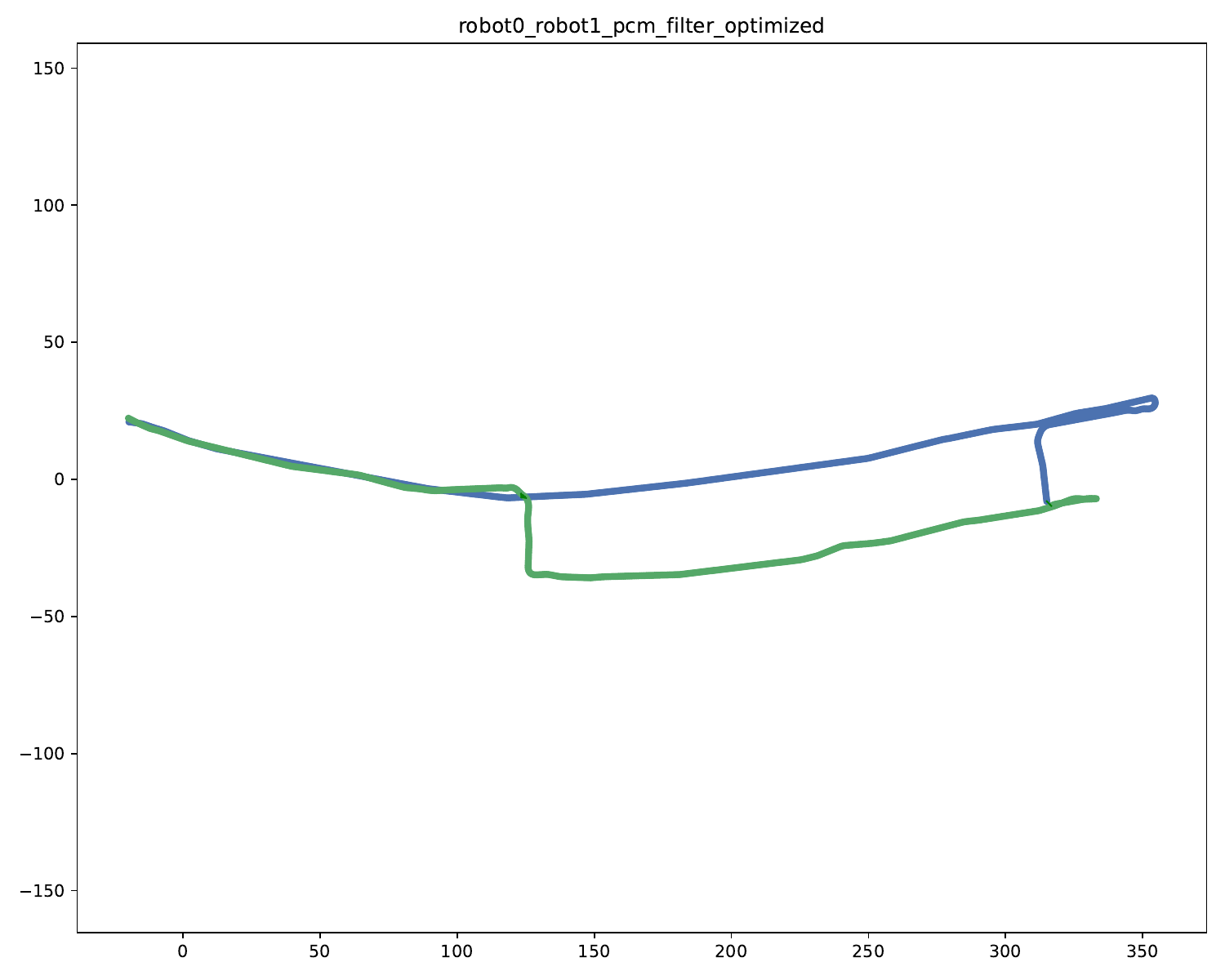}
        \par\smallskip
        \footnotesize\textbf{(f)} g2o graph with tunnel-filtered KFs after optimization
    \end{minipage}
    
    \caption{g2o graph visualization for the robot pair (0,1)}
\end{figure}

% --------- Experiment (0,2) ---------
\begin{figure}[H]
    \centering

    % Tunnels
    \begin{minipage}[t]{0.32\textwidth}
        \centering
        \includegraphics[width=\textwidth]{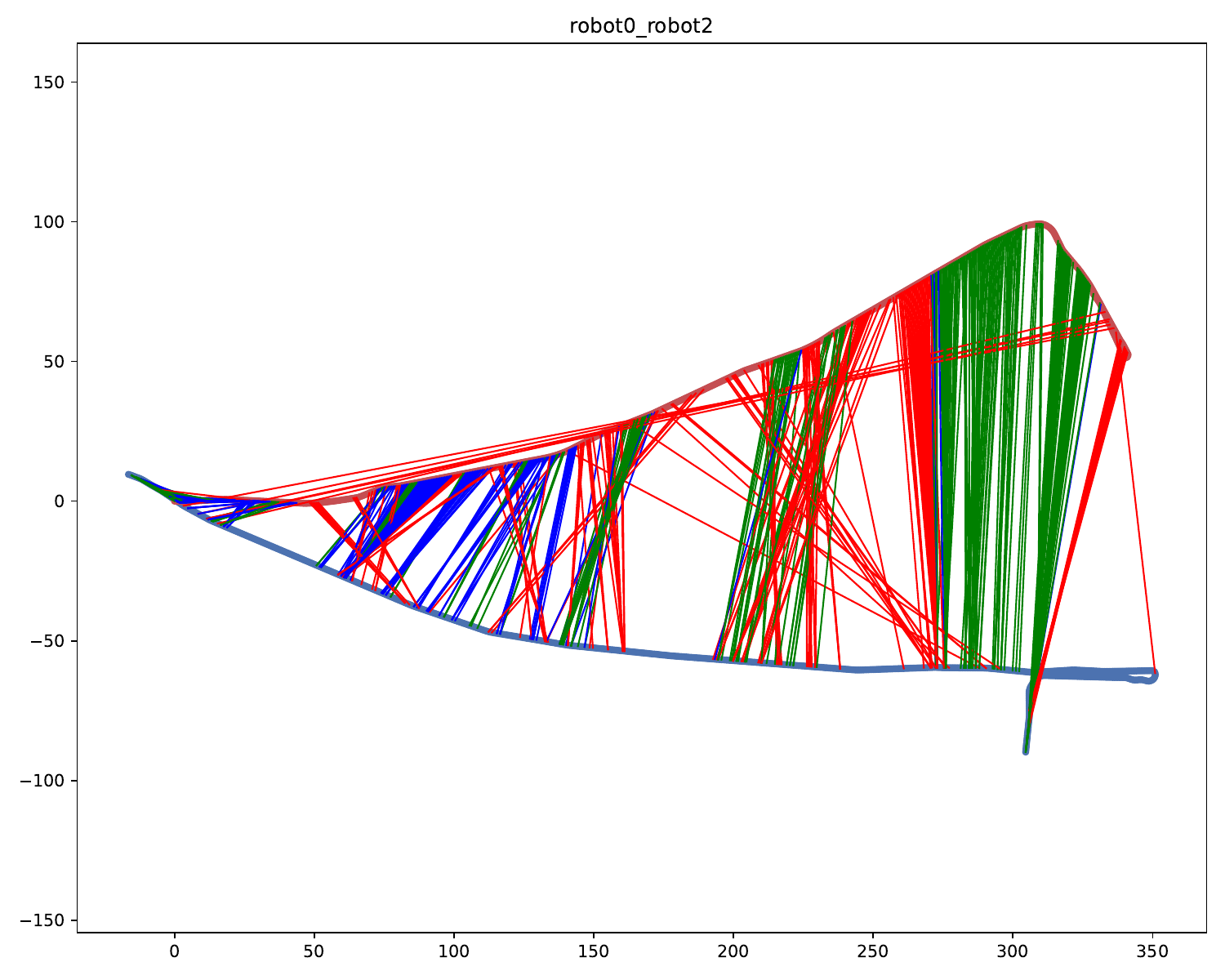}
        \par\smallskip
        \footnotesize\textbf{(a)} Initial g2o graph with all KFs
    \end{minipage}
    \hfill
    \begin{minipage}[t]{0.32\textwidth}
        \centering
        \includegraphics[width=\textwidth]{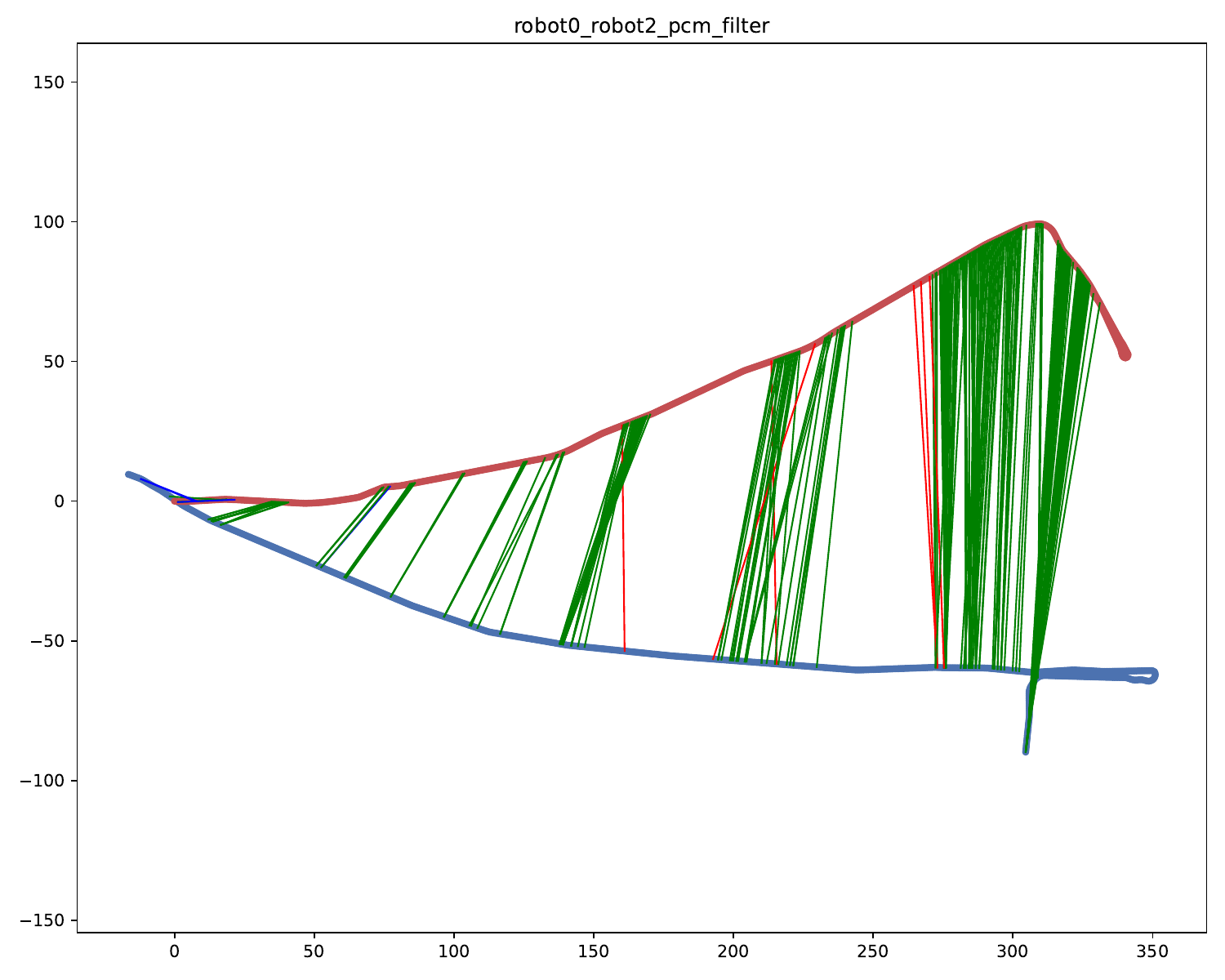}
        \par\smallskip
        \footnotesize\textbf{(b)} g2o graph with all KFs after PCM
    \end{minipage}
    \hfill
    \begin{minipage}[t]{0.32\textwidth}
        \centering
        \includegraphics[width=\textwidth]{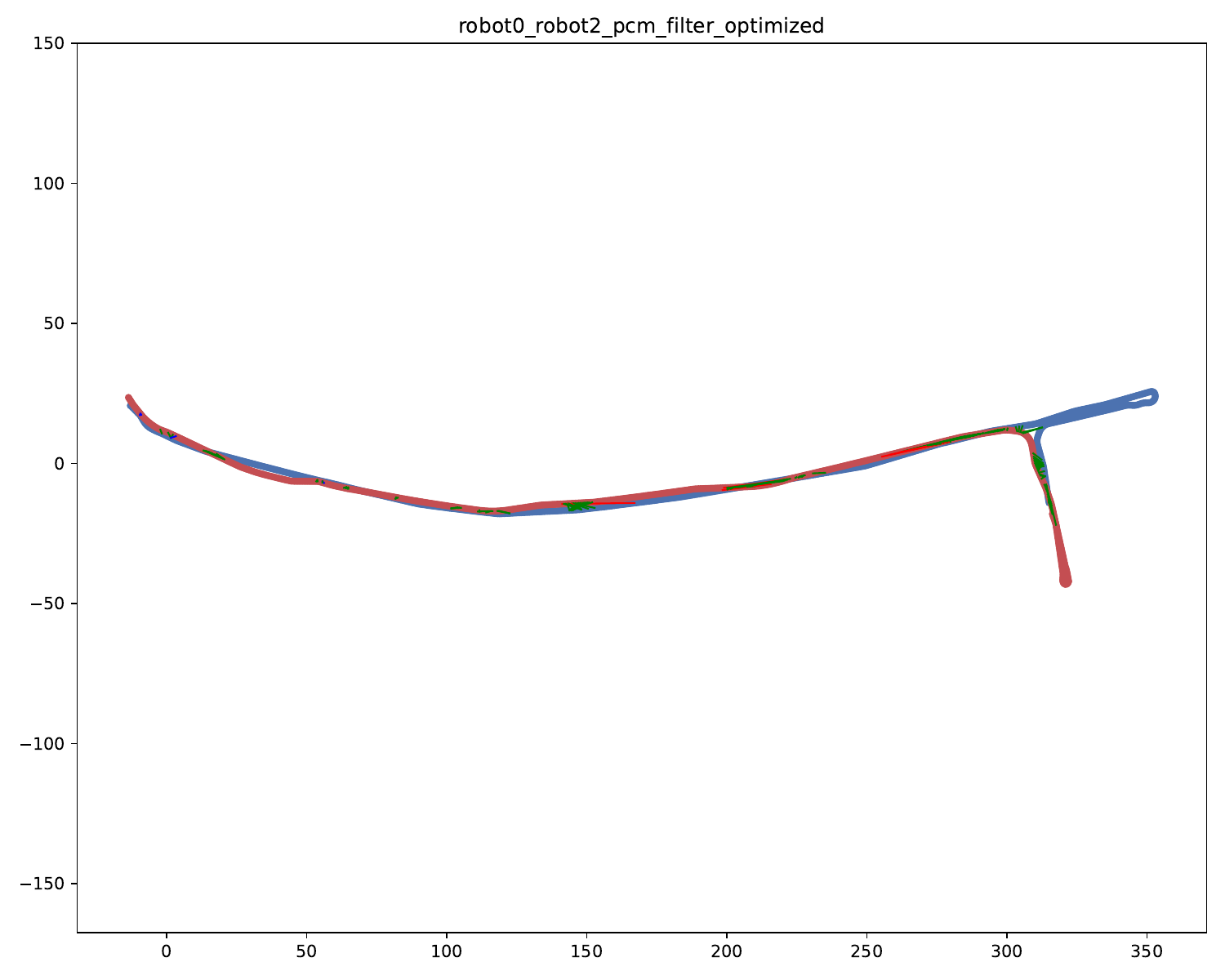}
        \par\smallskip
        \footnotesize\textbf{(c)} g2o graph with all KFs after optimization
    \end{minipage}

    % No tunnels
    \begin{minipage}[t]{0.32\textwidth}
        \centering
        \includegraphics[width=\textwidth]{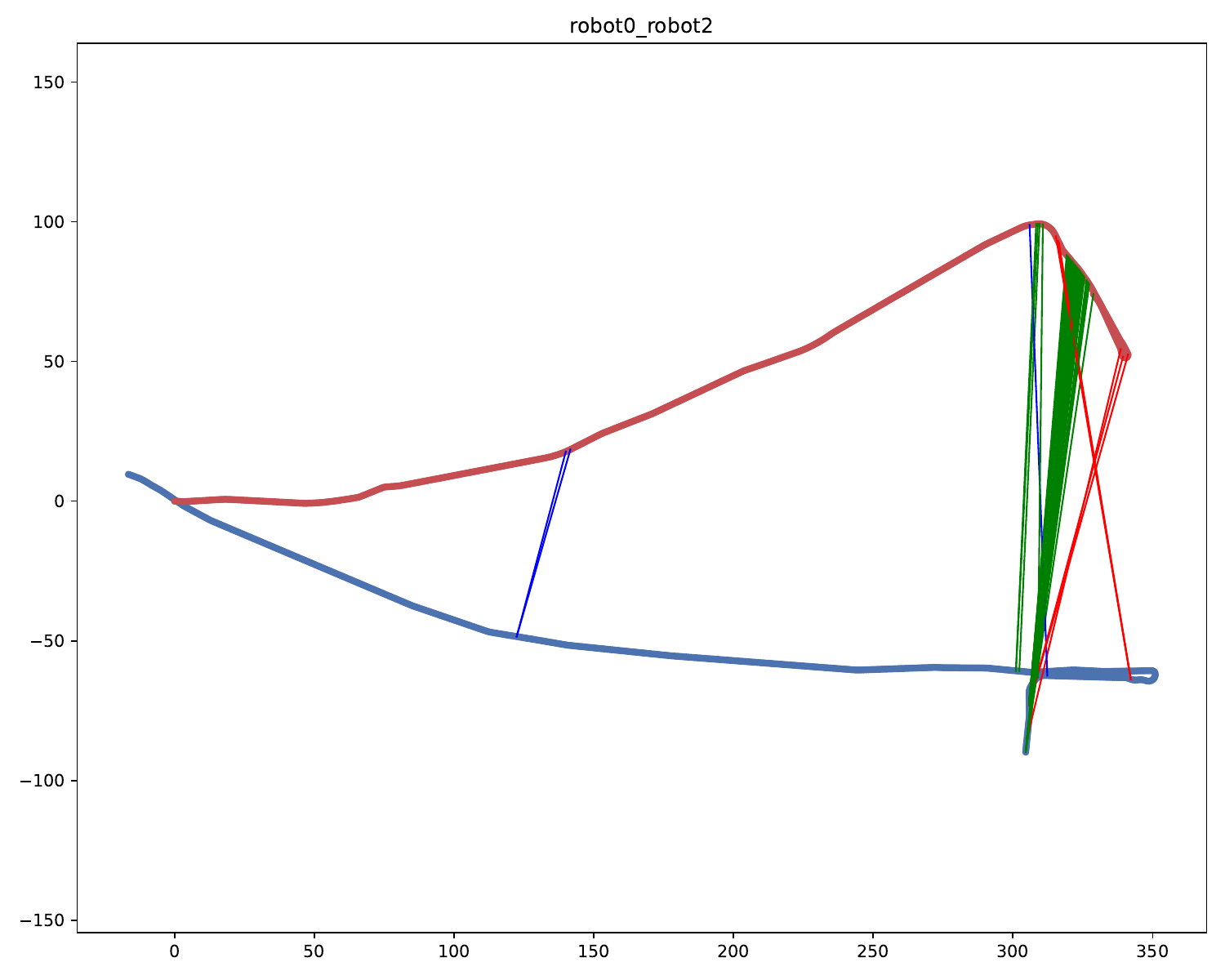}
        \par\smallskip
        \footnotesize\textbf{(d)} Initial g2o graph with tunnel-filtered KFs
    \end{minipage}
    \hfill
    \begin{minipage}[t]{0.32\textwidth}
        \centering
        \includegraphics[width=\textwidth]{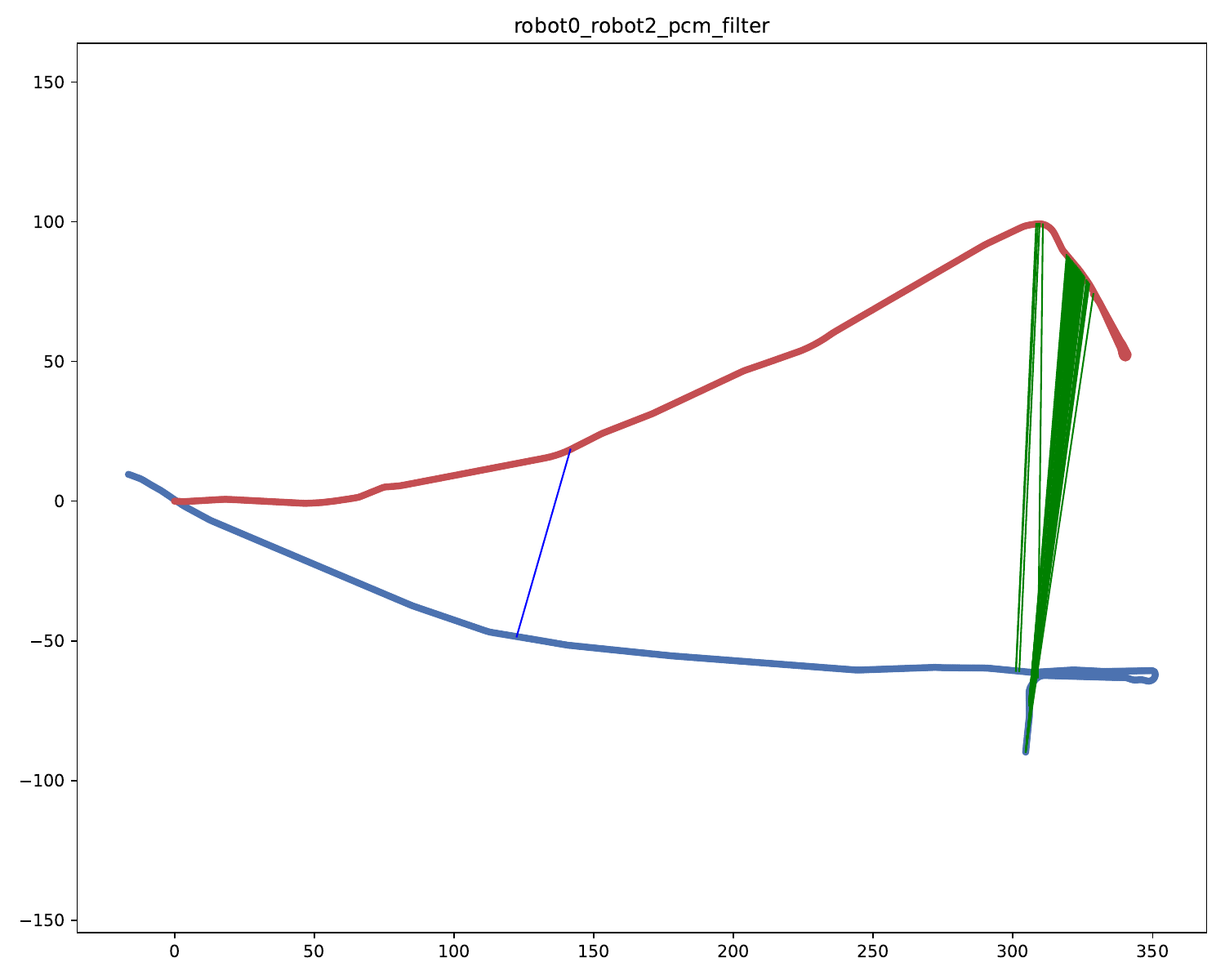}
        \par\smallskip
        \footnotesize\textbf{(e)} g2o graph with tunnel-filtered KFs after PCM
    \end{minipage}
    \hfill
    \begin{minipage}[t]{0.32\textwidth}
        \centering
        \includegraphics[width=\textwidth]{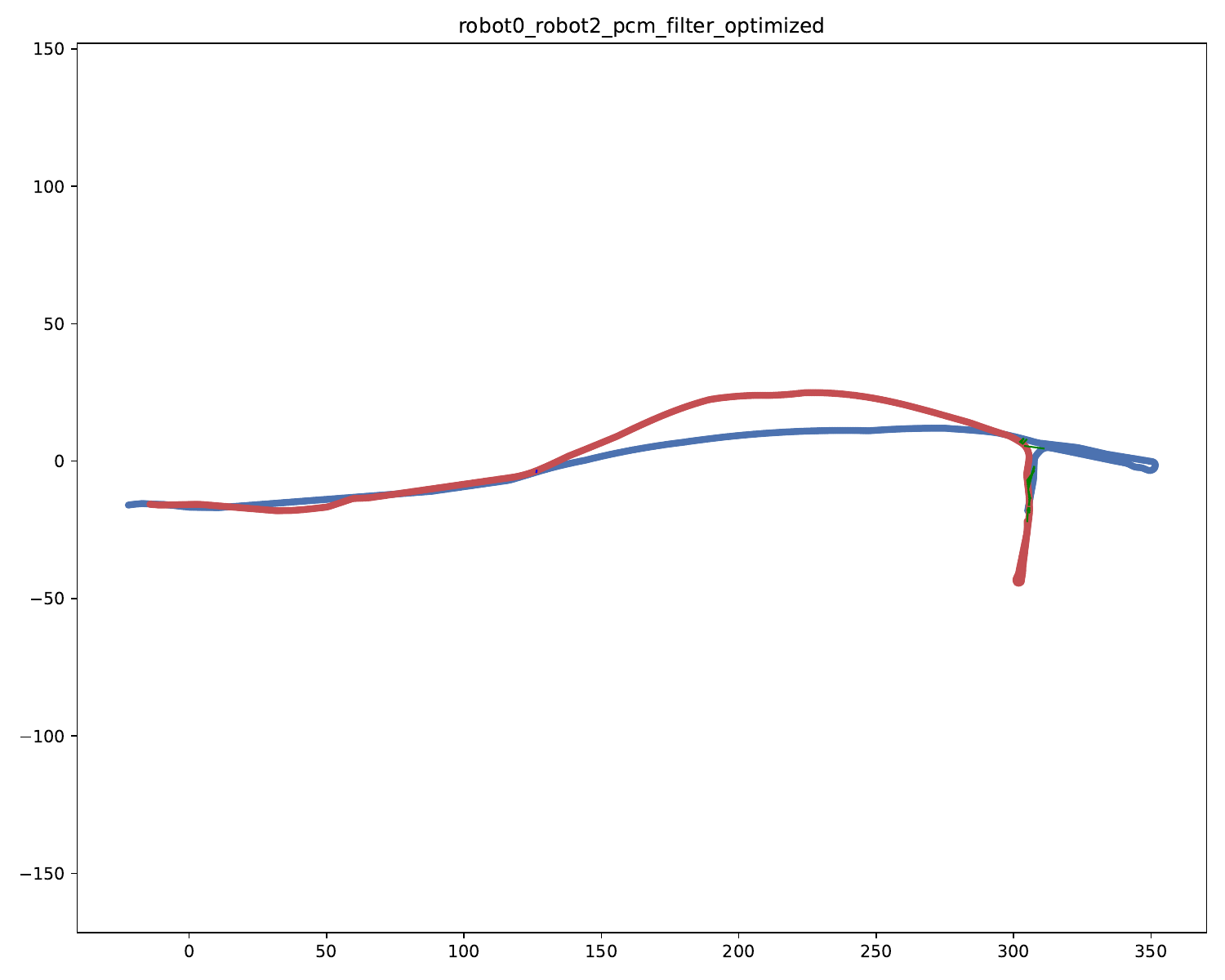}
        \par\smallskip
        \footnotesize\textbf{(f)} g2o graph with tunnel-filtered KFs after optimization
    \end{minipage}
    
    \caption{g2o graph visualization for the robot pair (0,2)}
\end{figure}

% --------- Experiment (0,3) ---------
\begin{figure}[H]
    \centering

    % Tunnels
    \begin{minipage}[t]{0.32\textwidth}
        \centering
        \includegraphics[width=\textwidth]{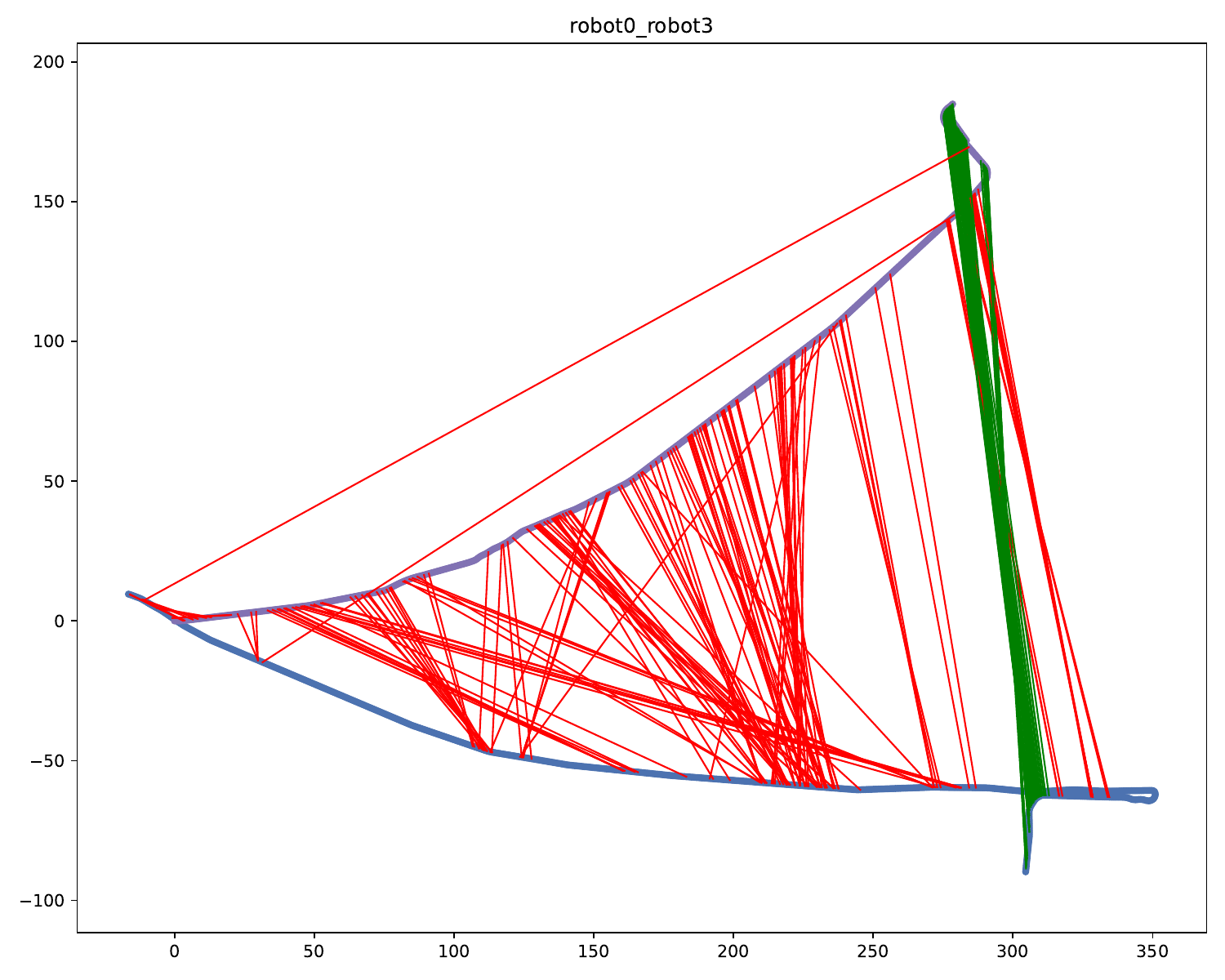}
        \par\smallskip
        \footnotesize\textbf{(a)} Initial g2o graph with all KFs
    \end{minipage}
    \hfill
    \begin{minipage}[t]{0.32\textwidth}
        \centering
        \includegraphics[width=\textwidth]{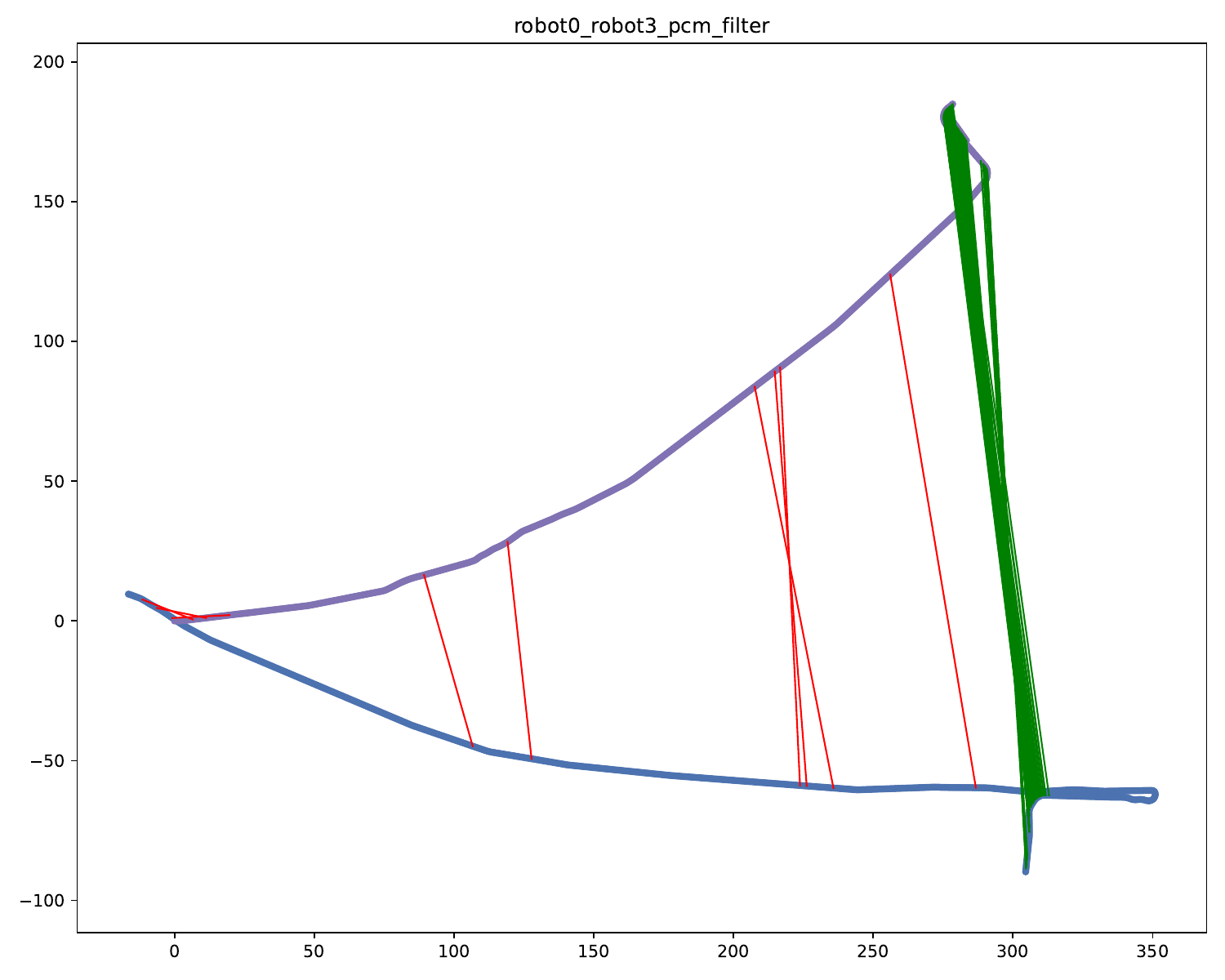}
        \par\smallskip
        \footnotesize\textbf{(b)} g2o graph with all KFs after PCM
    \end{minipage}
    \hfill
    \begin{minipage}[t]{0.32\textwidth}
        \centering
        \includegraphics[width=\textwidth]{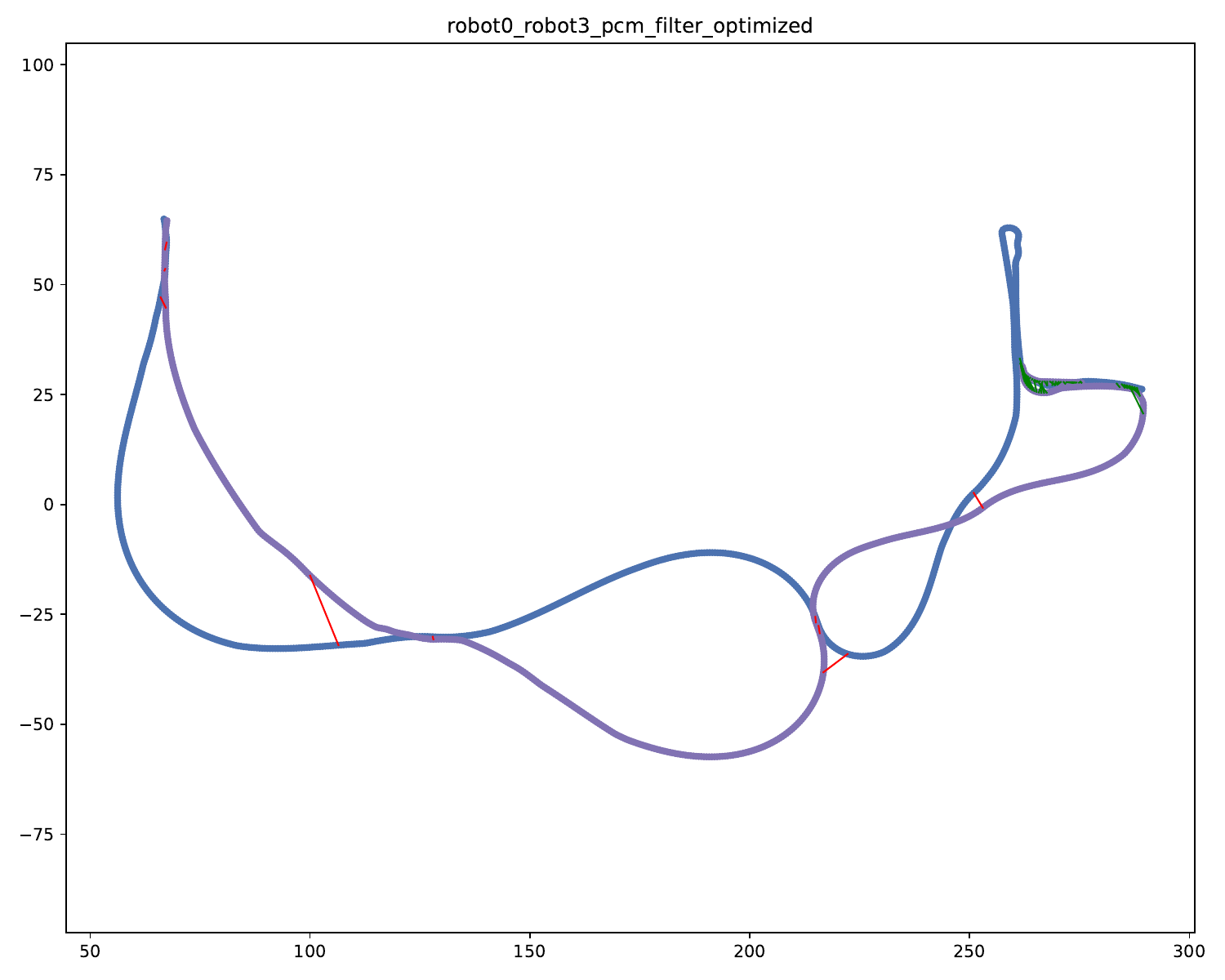}
        \par\smallskip
        \footnotesize\textbf{(c)} g2o graph with all KFs after optimization
    \end{minipage}

    % No tunnels
    \begin{minipage}[t]{0.32\textwidth}
        \centering
        \includegraphics[width=\textwidth]{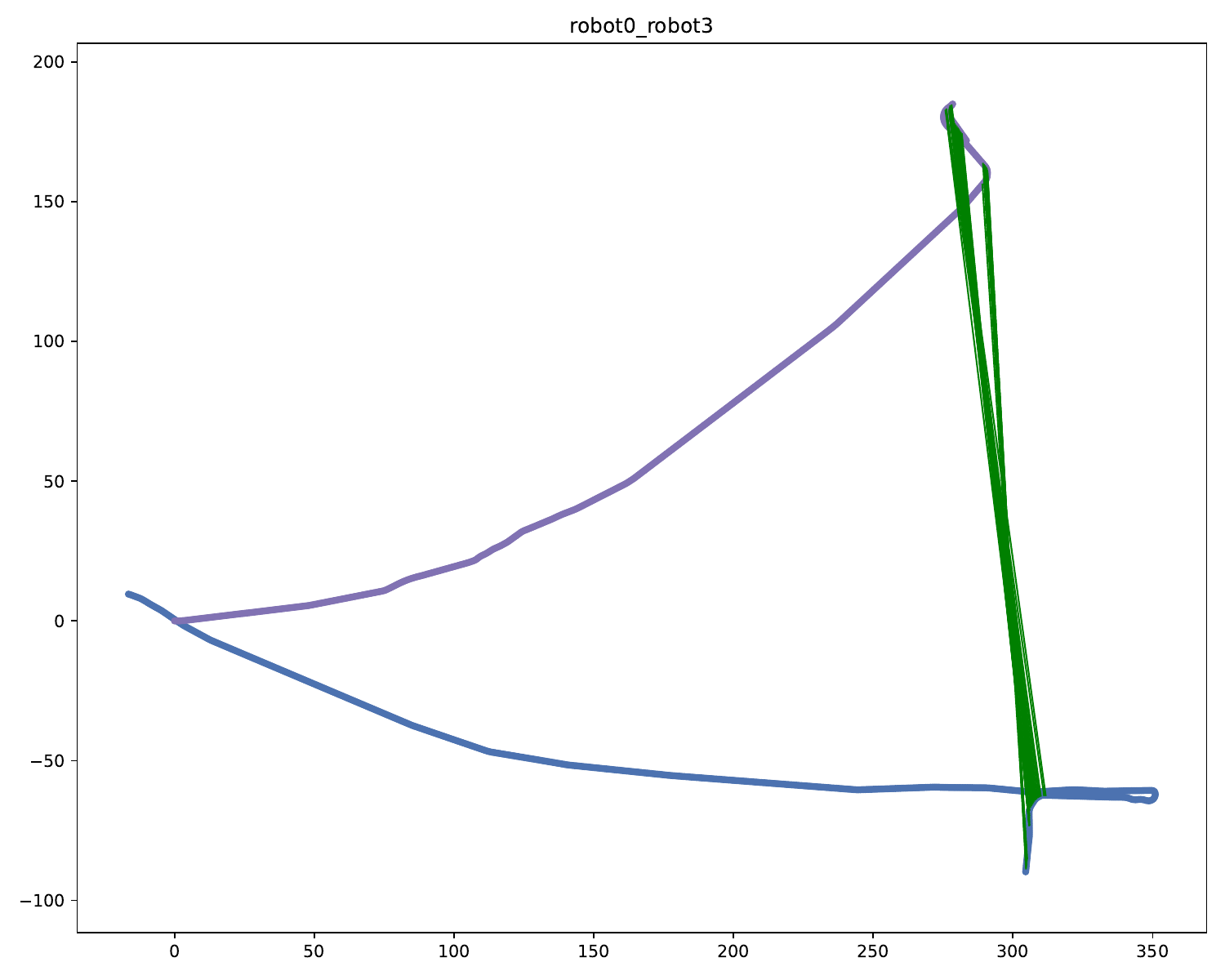}
        \par\smallskip
        \footnotesize\textbf{(d)} Initial g2o graph with tunnel-filtered KFs
    \end{minipage}
    \hfill
    \begin{minipage}[t]{0.32\textwidth}
        \centering
        \includegraphics[width=\textwidth]{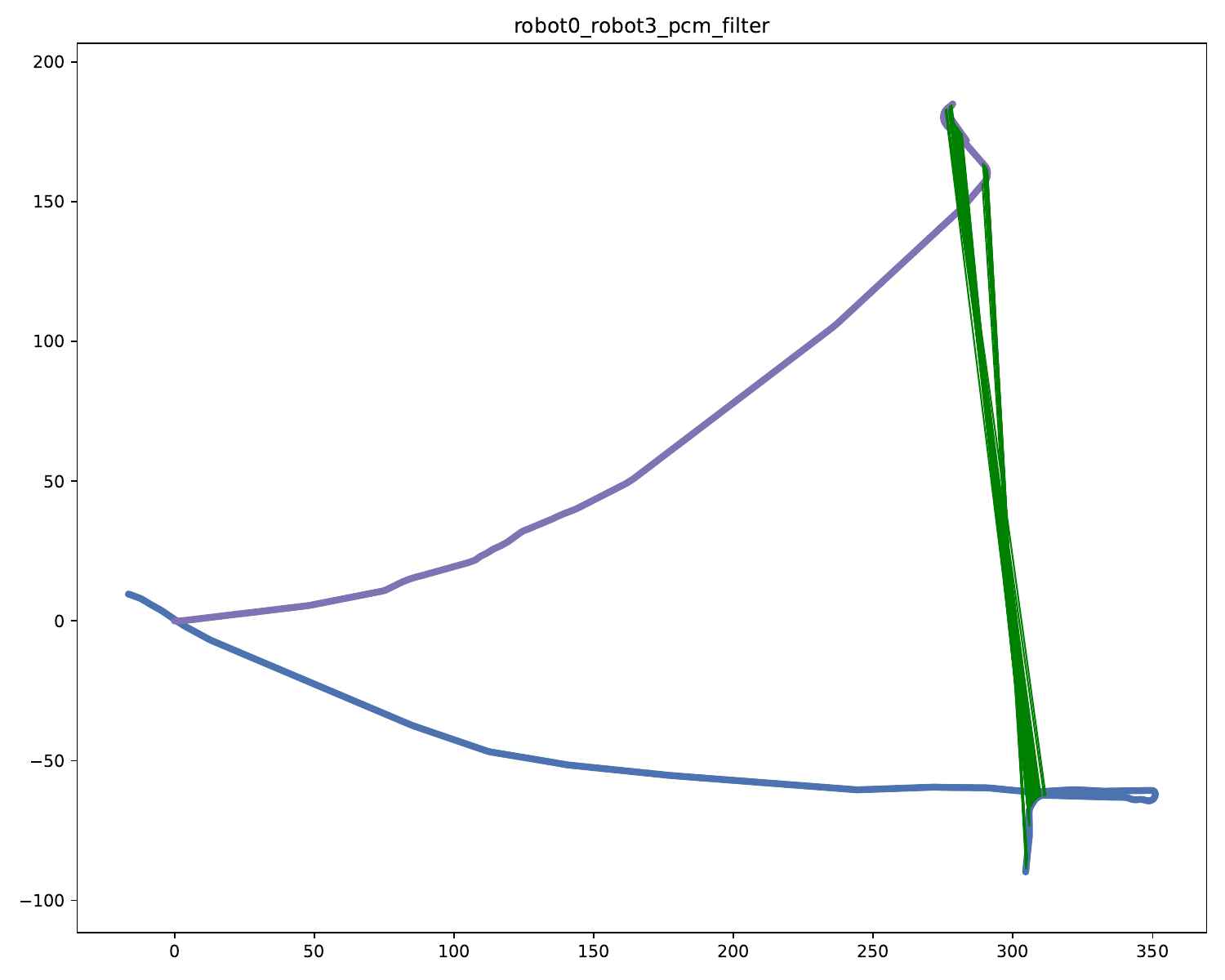}
        \par\smallskip
        \footnotesize\textbf{(e)} g2o graph with tunnel-filtered KFs after PCM
    \end{minipage}
    \hfill
    \begin{minipage}[t]{0.32\textwidth}
        \centering
        \includegraphics[width=\textwidth]{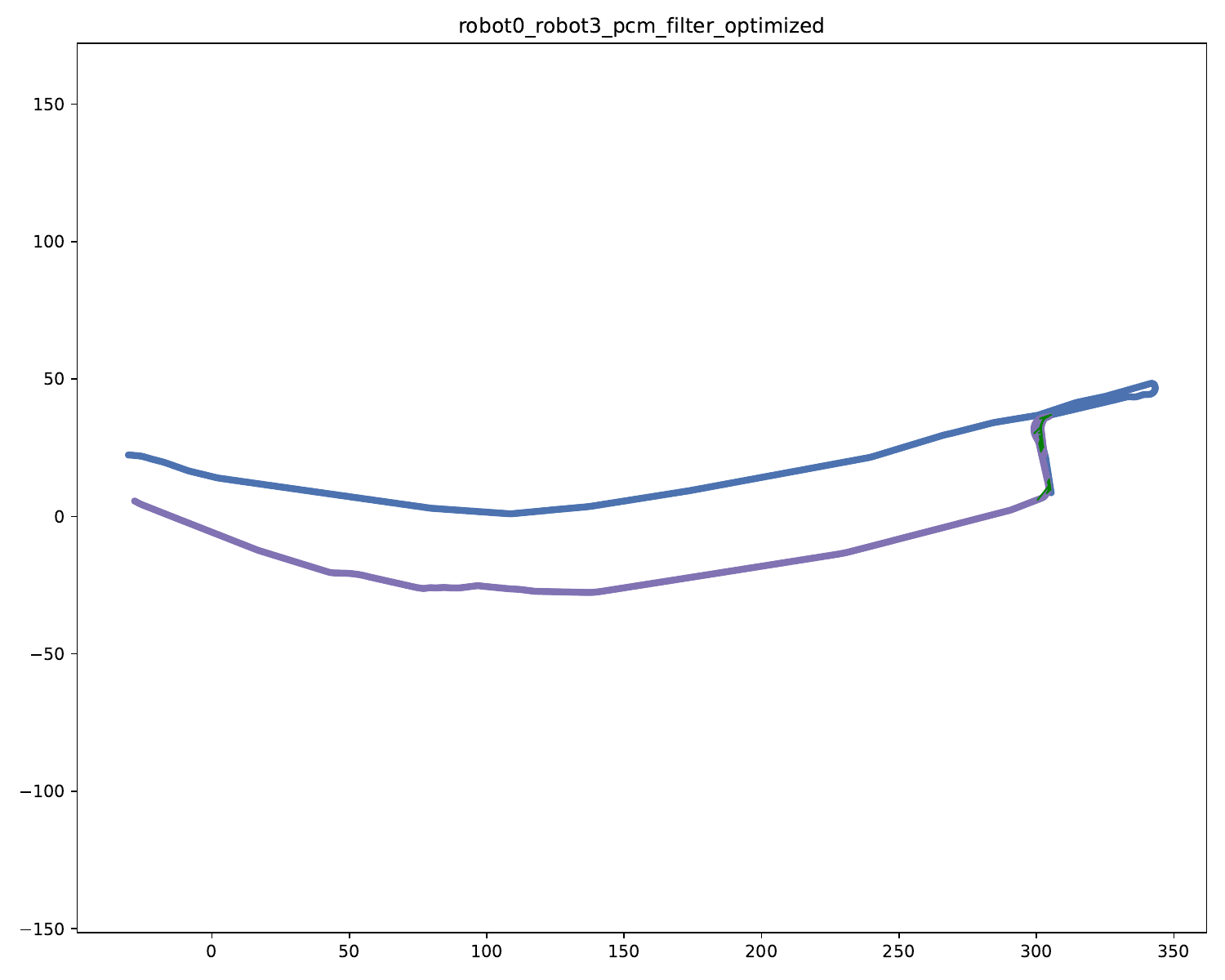}
        \par\smallskip
        \footnotesize\textbf{(f)} g2o graph with tunnel-filtered KFs after optimization
    \end{minipage}
    
    \caption{g2o graph visualization for the robot pair (0,3)}
\end{figure}

% --------- Experiment (1,2) ---------
\begin{figure}[H]
    \centering

    % Tunnels
    \begin{minipage}[t]{0.32\textwidth}
        \centering
        \includegraphics[width=\textwidth]{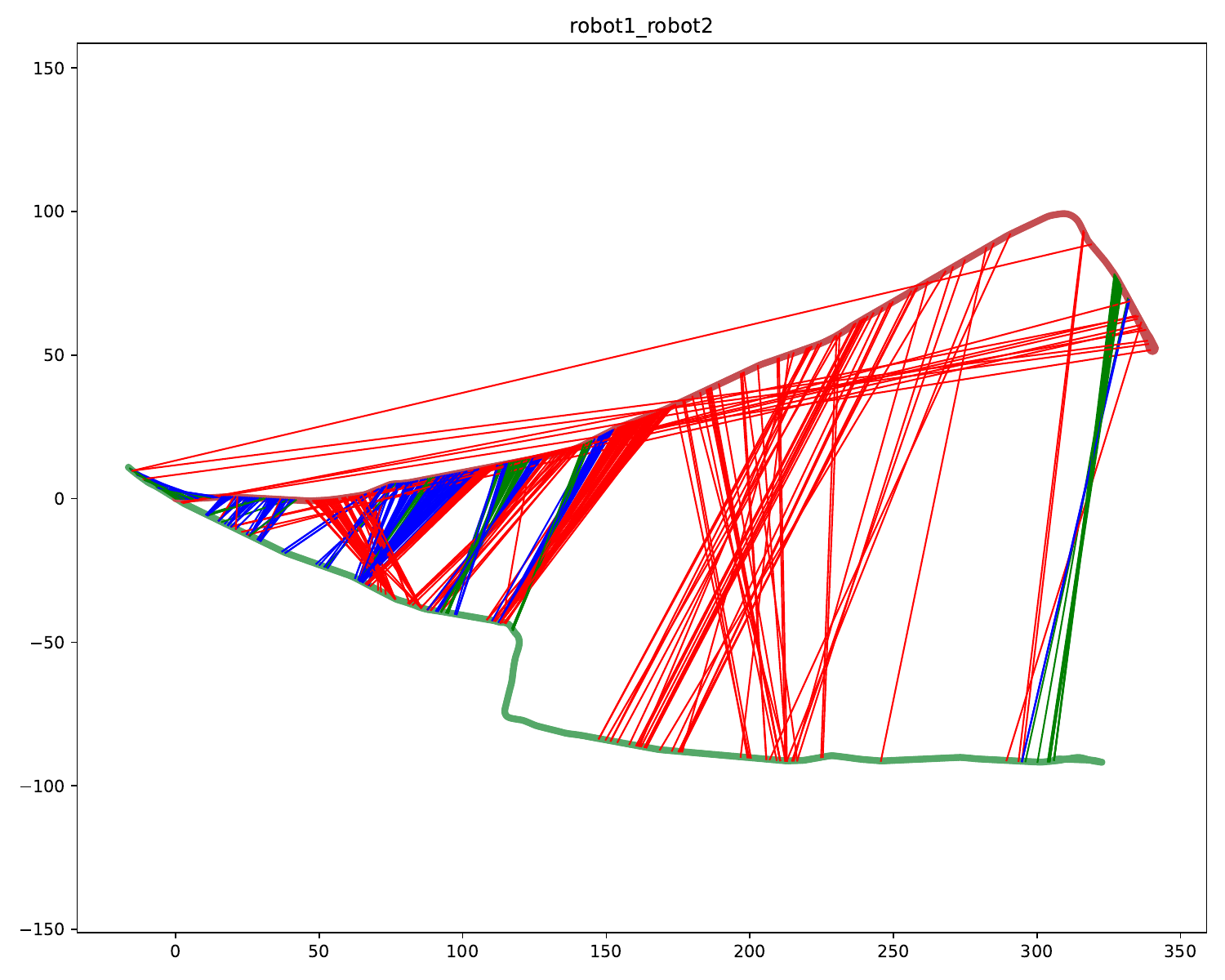}
        \par\smallskip
        \footnotesize\textbf{(a)} Initial g2o graph with all KFs
    \end{minipage}
    \hfill
    \begin{minipage}[t]{0.32\textwidth}
        \centering
        \includegraphics[width=\textwidth]{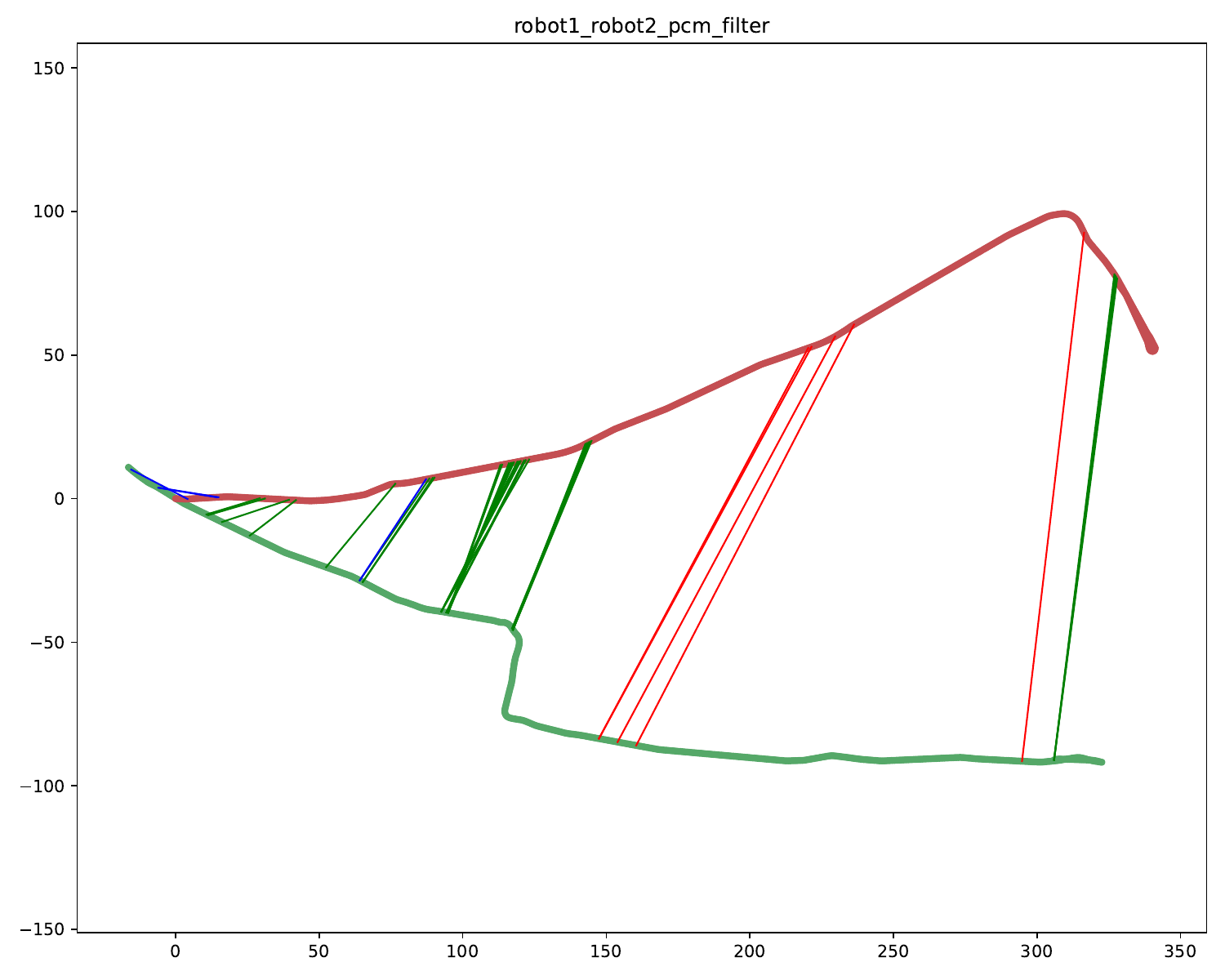}
        \par\smallskip
        \footnotesize\textbf{(b)} g2o graph with all KFs after PCM
    \end{minipage}
    \hfill
    \begin{minipage}[t]{0.32\textwidth}
        \centering
        \includegraphics[width=\textwidth]{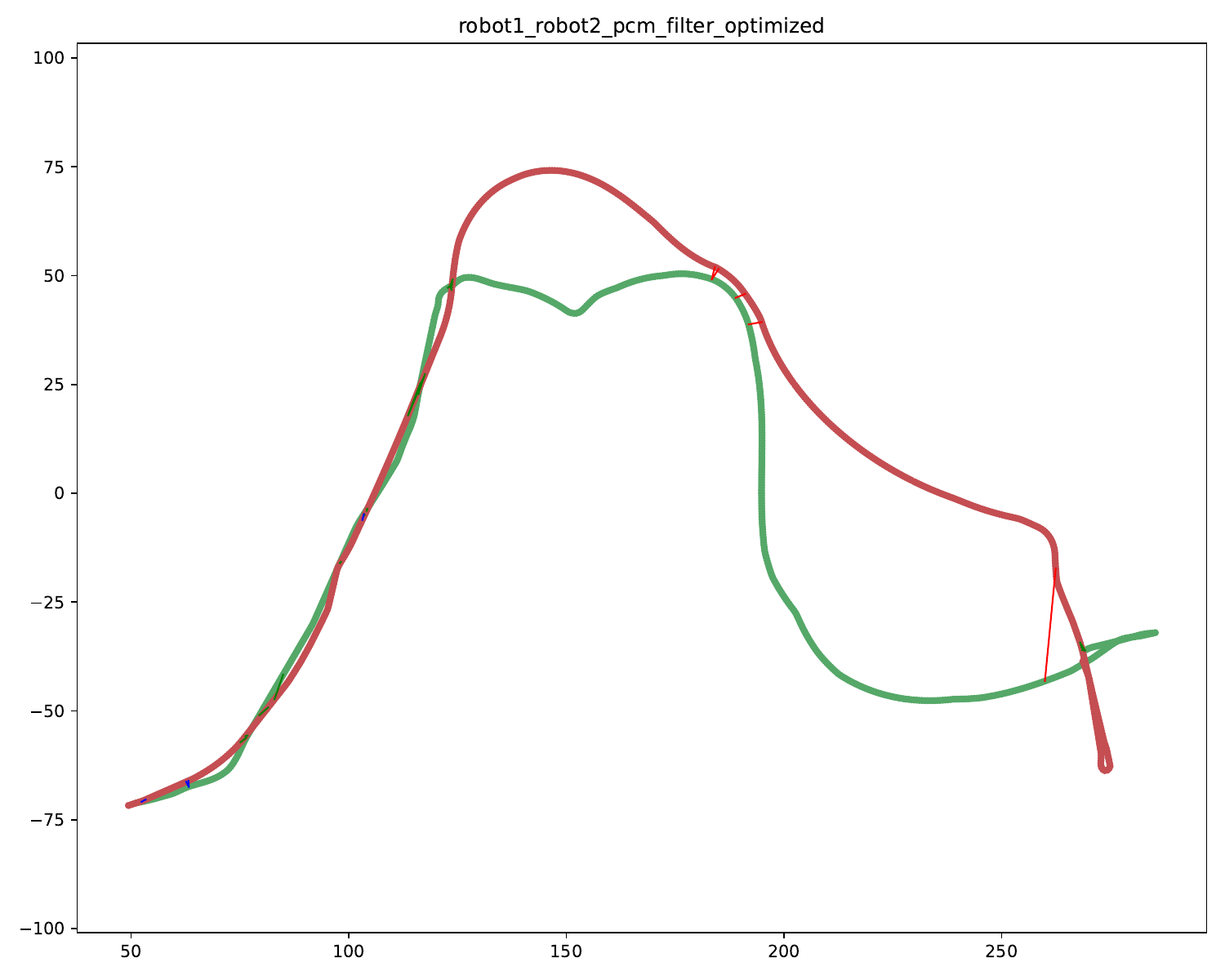}
        \par\smallskip
        \footnotesize\textbf{(c)} g2o graph with all KFs after optimization
    \end{minipage}

    % No tunnels
    \begin{minipage}[t]{0.32\textwidth}
        \centering
        \includegraphics[width=\textwidth]{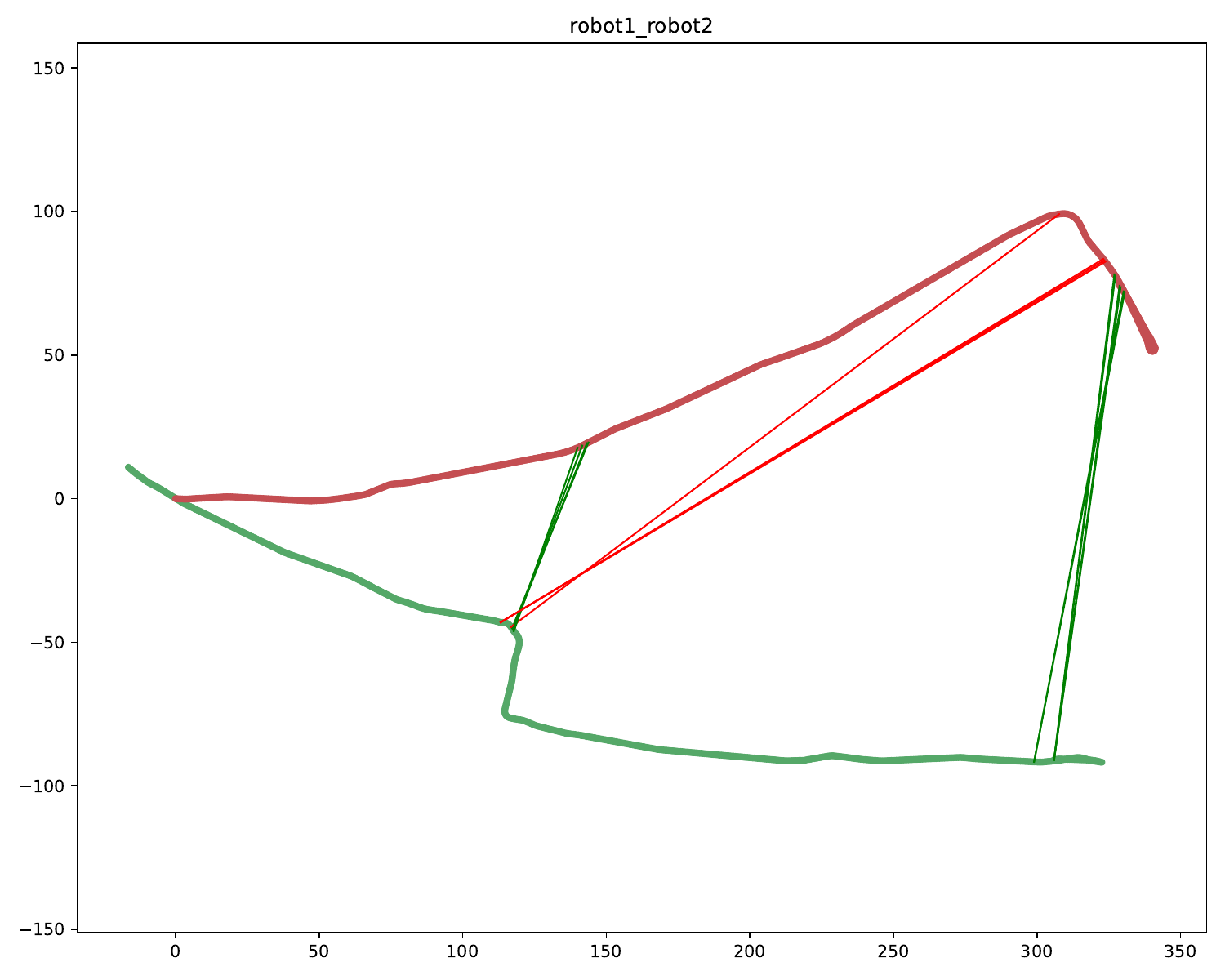}
        \par\smallskip
        \footnotesize\textbf{(d)} Initial g2o graph with tunnel-filtered KFs
    \end{minipage}
    \hfill
    \begin{minipage}[t]{0.32\textwidth}
        \centering
        \includegraphics[width=\textwidth]{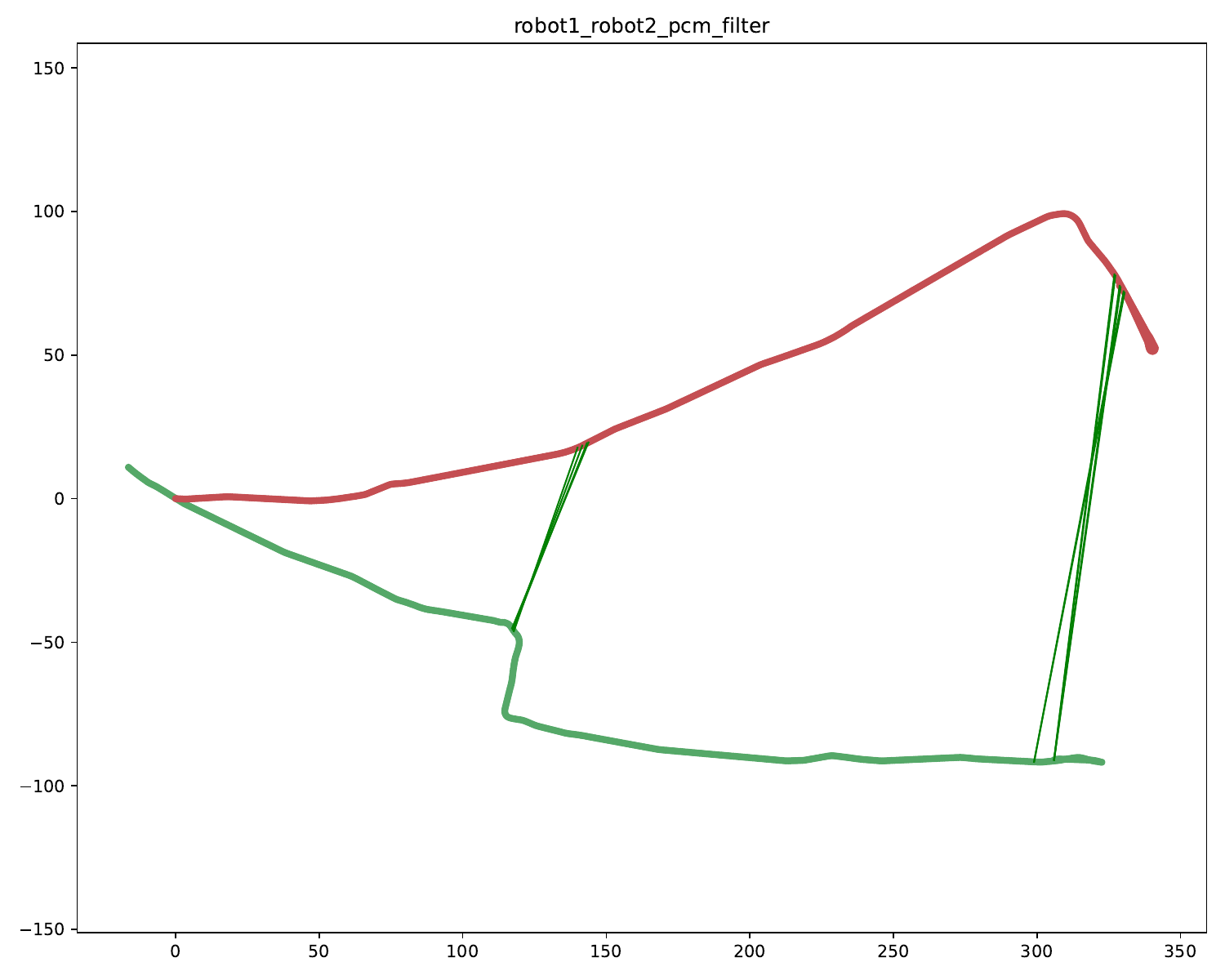}
        \par\smallskip
        \footnotesize\textbf{(e)} g2o graph with tunnel-filtered KFs after PCM
    \end{minipage}
    \hfill
    \begin{minipage}[t]{0.32\textwidth}
        \centering
        \includegraphics[width=\textwidth]{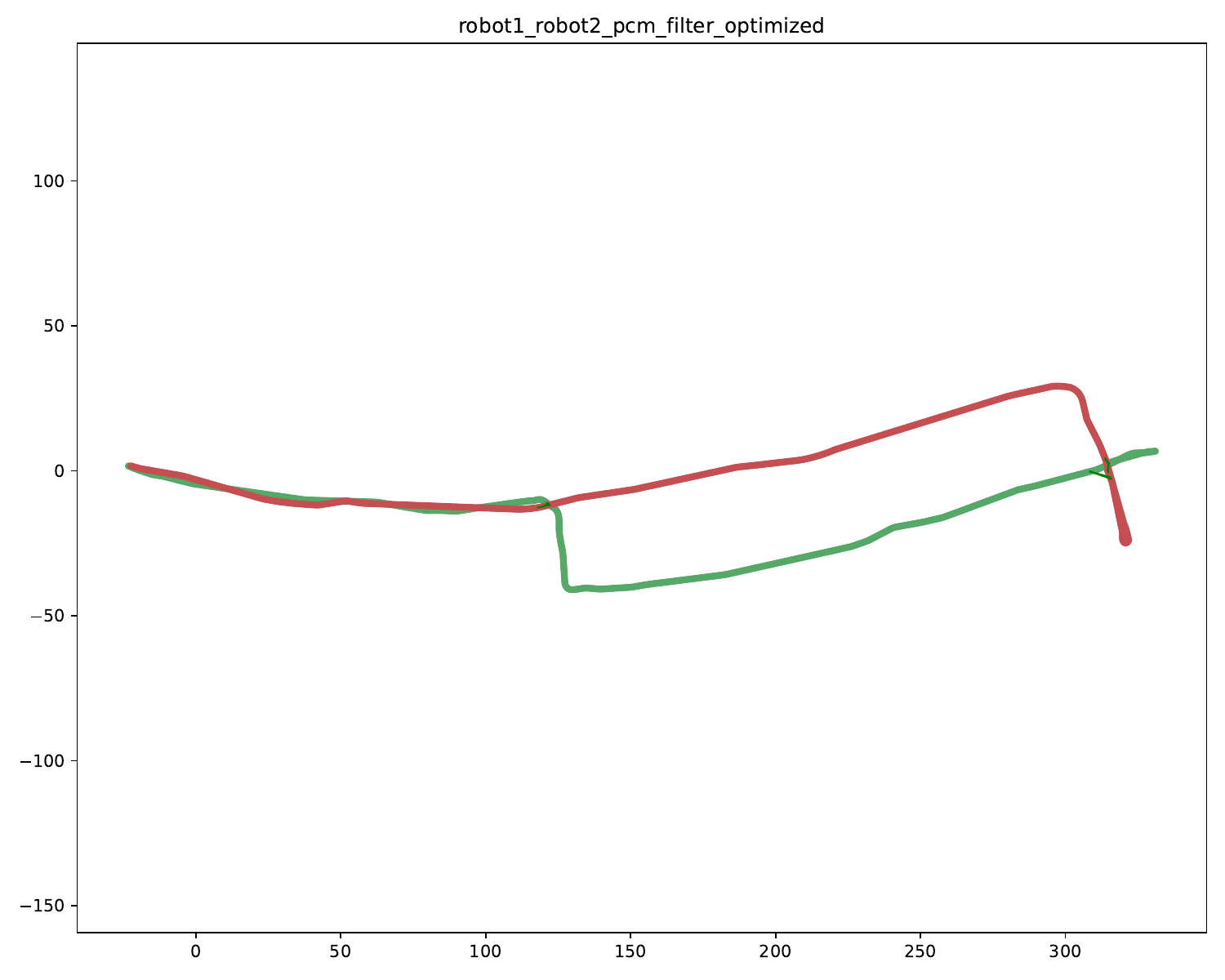}
        \par\smallskip
        \footnotesize\textbf{(f)} g2o graph with tunnel-filtered KFs after optimization
    \end{minipage}
    
    \caption{g2o graph visualization for the robot pair (1,2)}
\end{figure}

% --------- Experiment (1,3) ---------
\begin{figure}[H]
    \centering

    % Tunnels
    \begin{minipage}[t]{0.32\textwidth}
        \centering
        \includegraphics[width=\textwidth]{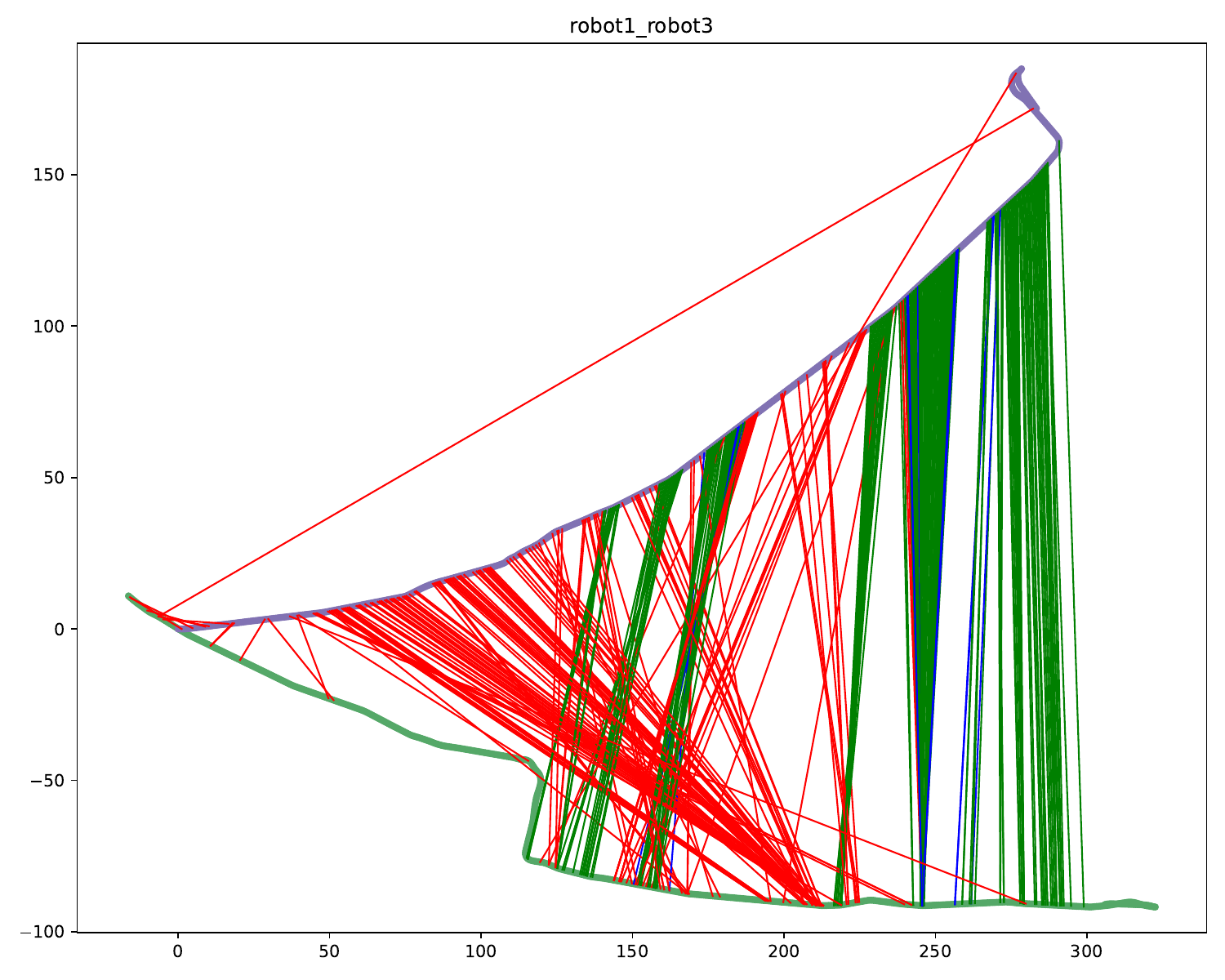}
        \par\smallskip
        \footnotesize\textbf{(a)} Initial g2o graph with all KFs
    \end{minipage}
    \hfill
    \begin{minipage}[t]{0.32\textwidth}
        \centering
        \includegraphics[width=\textwidth]{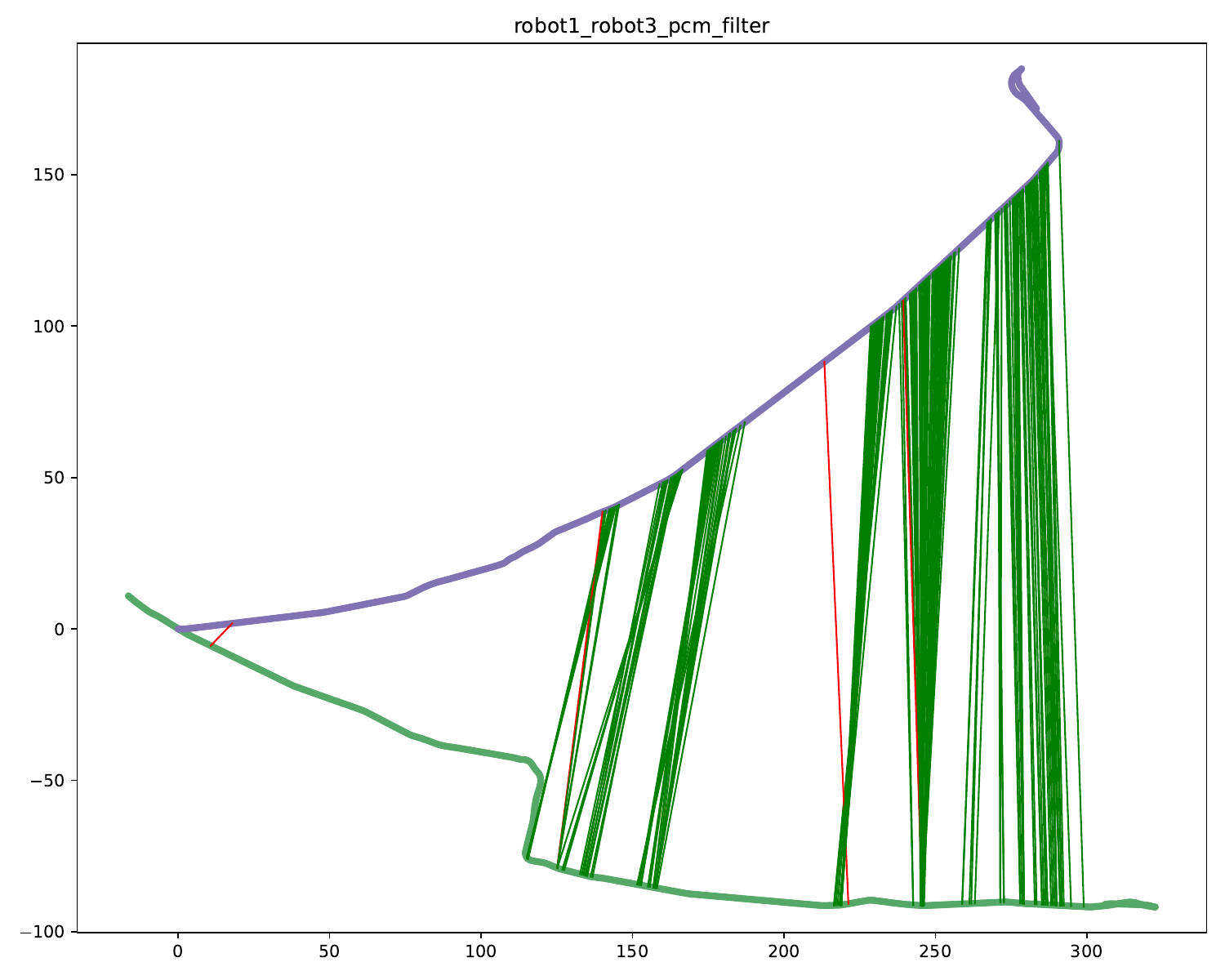}
        \par\smallskip
        \footnotesize\textbf{(b)} g2o graph with all KFs after PCM
    \end{minipage}
    \hfill
    \begin{minipage}[t]{0.32\textwidth}
        \centering
        \includegraphics[width=\textwidth]{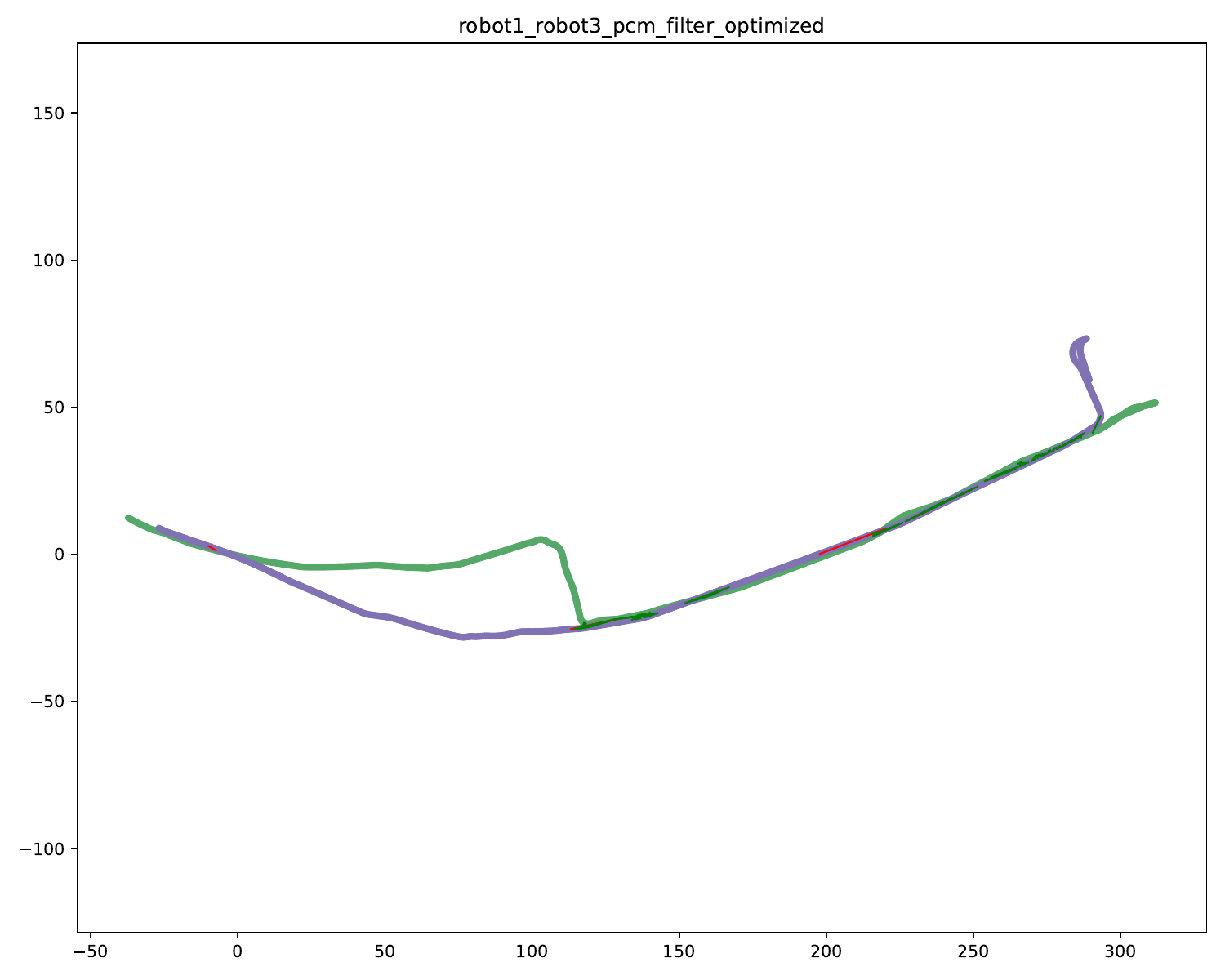}
        \par\smallskip
        \footnotesize\textbf{(c)} g2o graph with all KFs after optimization
    \end{minipage}

    % No tunnels
    \begin{minipage}[t]{0.32\textwidth}
        \centering
        \includegraphics[width=\textwidth]{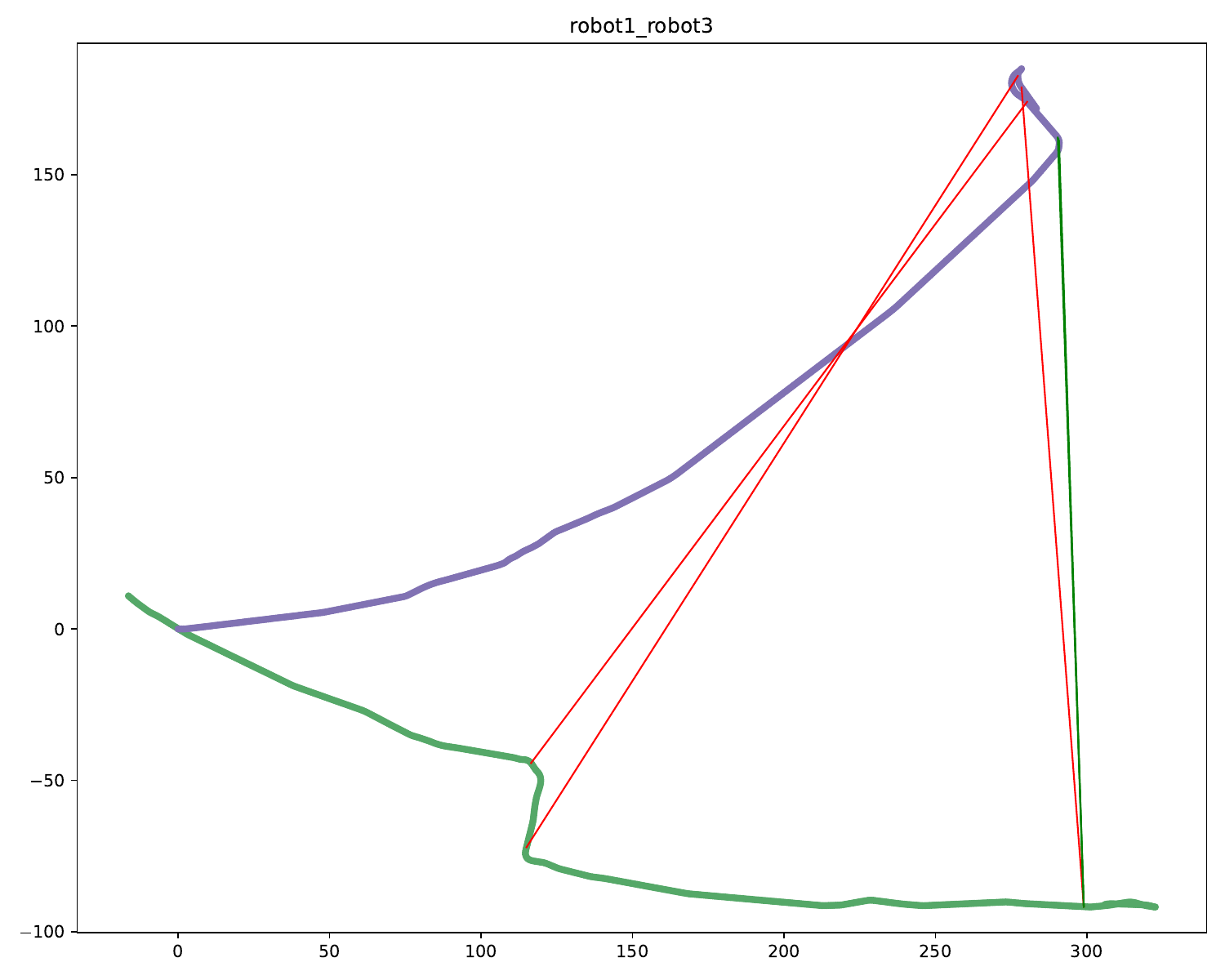}
        \par\smallskip
        \footnotesize\textbf{(d)} Initial g2o graph with tunnel-filtered KFs
    \end{minipage}
    \hfill
    \begin{minipage}[t]{0.32\textwidth}
        \centering
        \includegraphics[width=\textwidth]{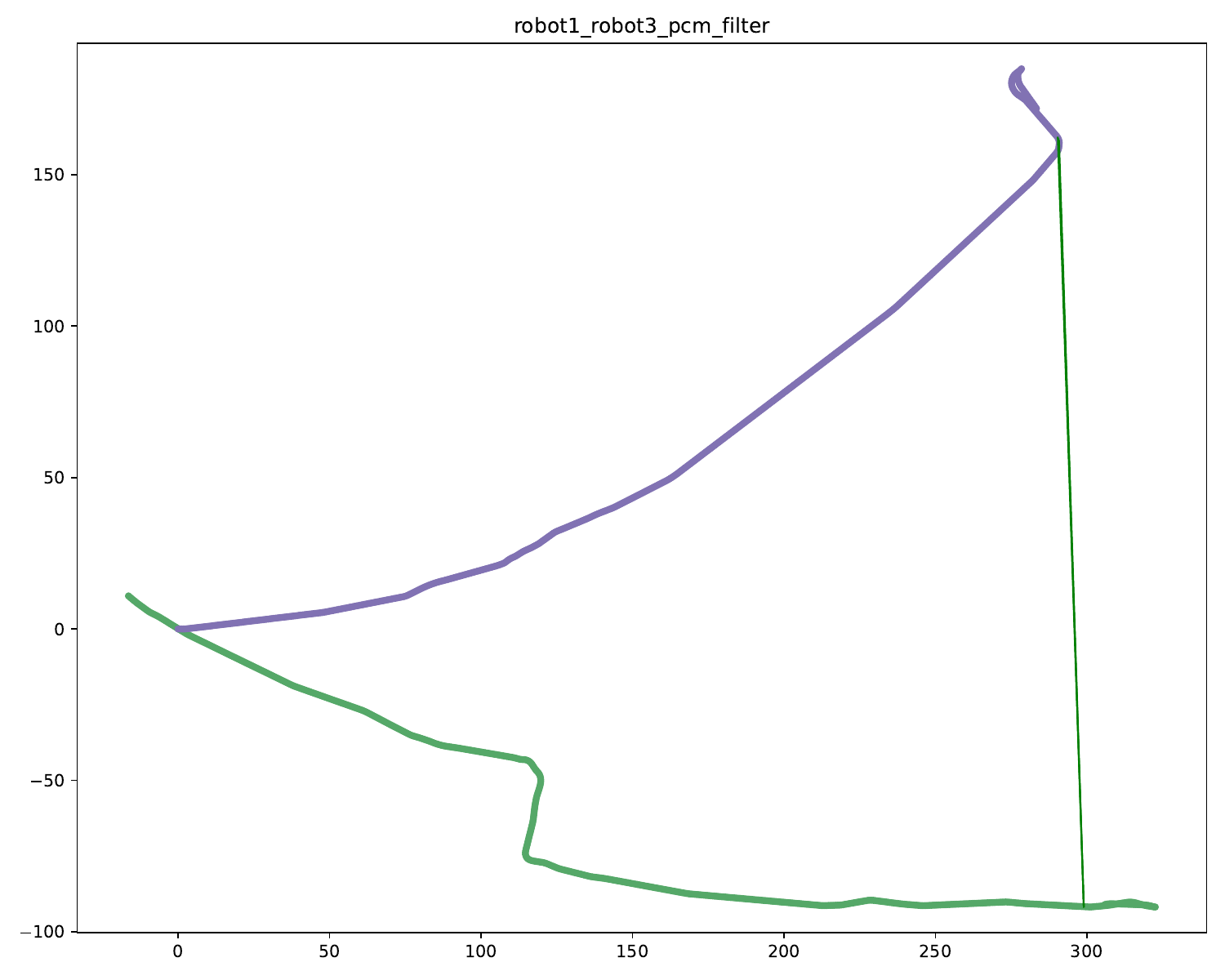}
        \par\smallskip
        \footnotesize\textbf{(e)} g2o graph with tunnel-filtered KFs after PCM
    \end{minipage}
    \hfill
    \begin{minipage}[t]{0.32\textwidth}
        \centering
        \includegraphics[width=\textwidth]{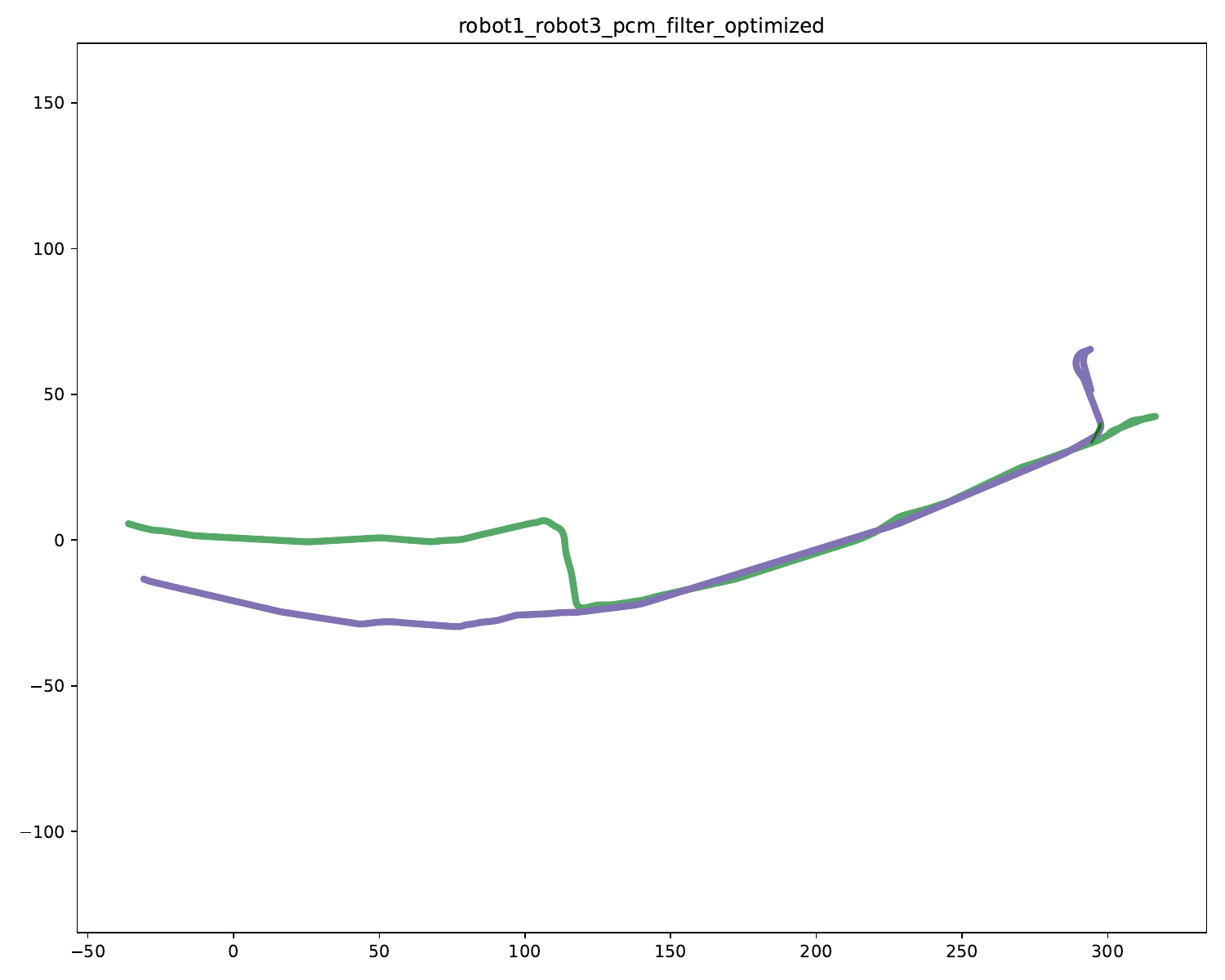}
        \par\smallskip
        \footnotesize\textbf{(f)} g2o graph with tunnel-filtered KFs after optimization
    \end{minipage}
    
    \caption{g2o graph visualization for the robot pair (1,3)}
\end{figure}

% --------- Experiment (2,3) ---------
\begin{figure}[H]
    \centering

    % Tunnels
    \begin{minipage}[t]{0.32\textwidth}
        \centering
        \includegraphics[width=\textwidth]{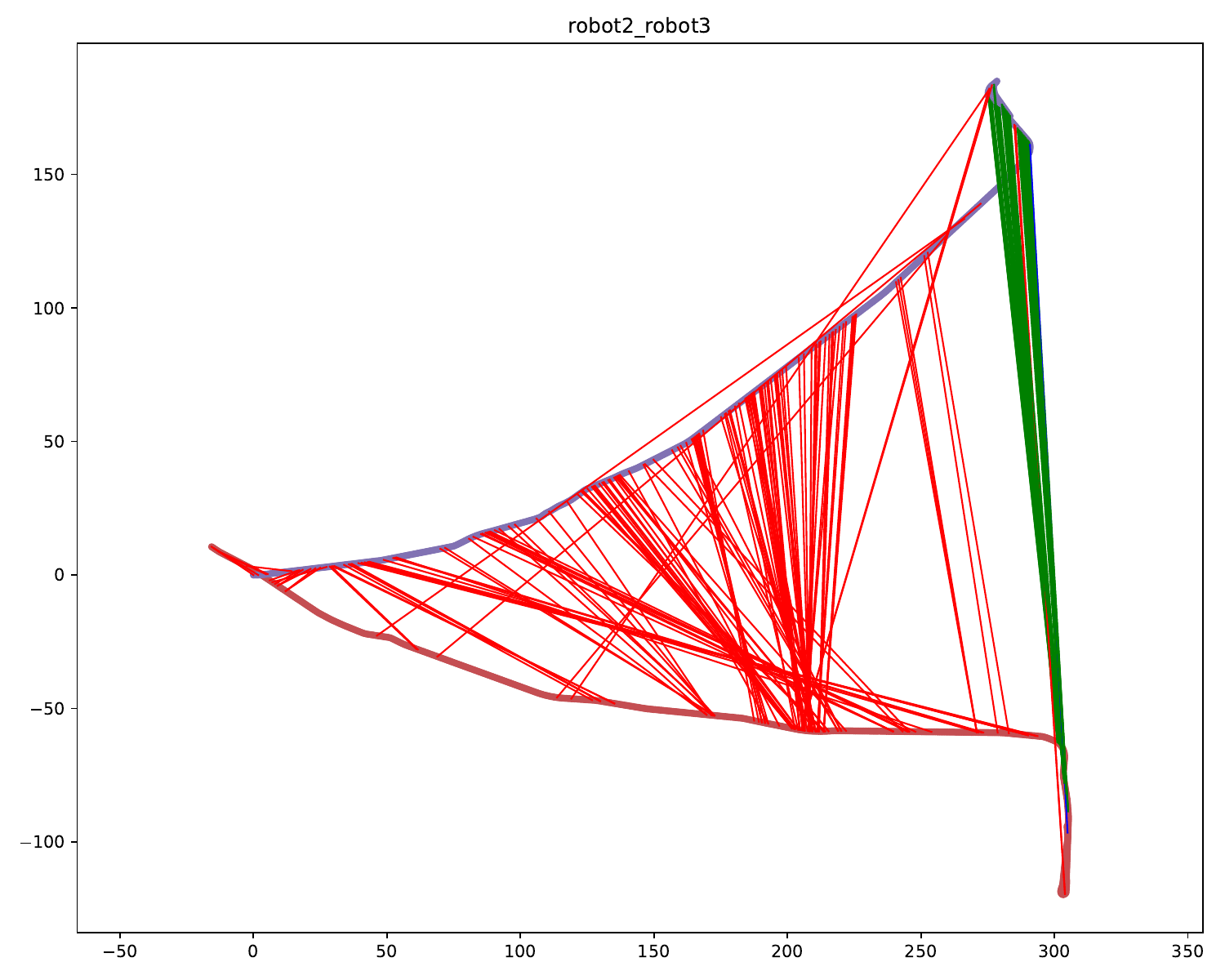}
        \par\smallskip
        \footnotesize\textbf{(a)} Initial g2o graph with all KFs
    \end{minipage}
    \hfill
    \begin{minipage}[t]{0.32\textwidth}
        \centering
        \includegraphics[width=\textwidth]{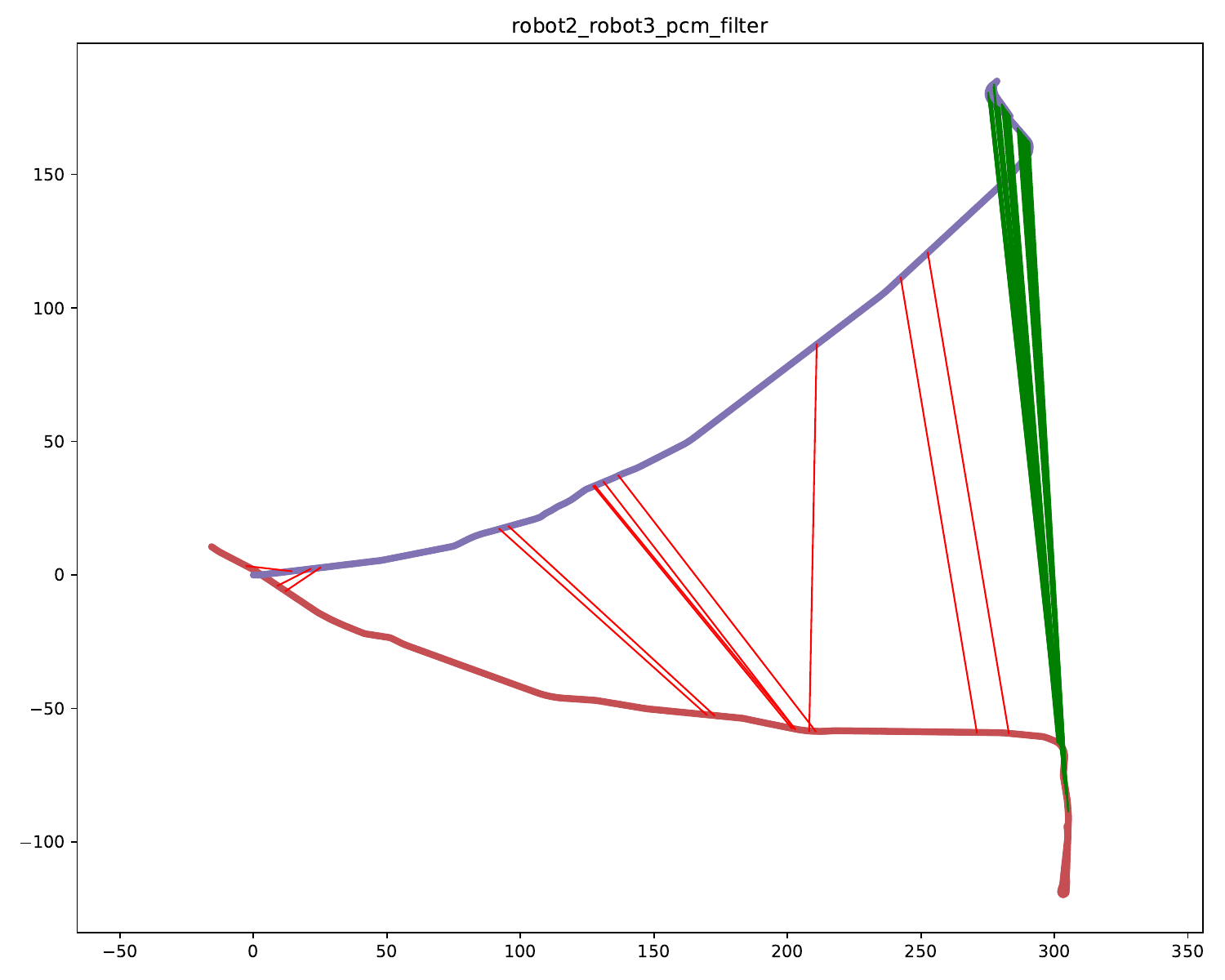}
        \par\smallskip
        \footnotesize\textbf{(b)} g2o graph with all KFs after PCM
    \end{minipage}
    \hfill
    \begin{minipage}[t]{0.32\textwidth}
        \centering
        \includegraphics[width=\textwidth]{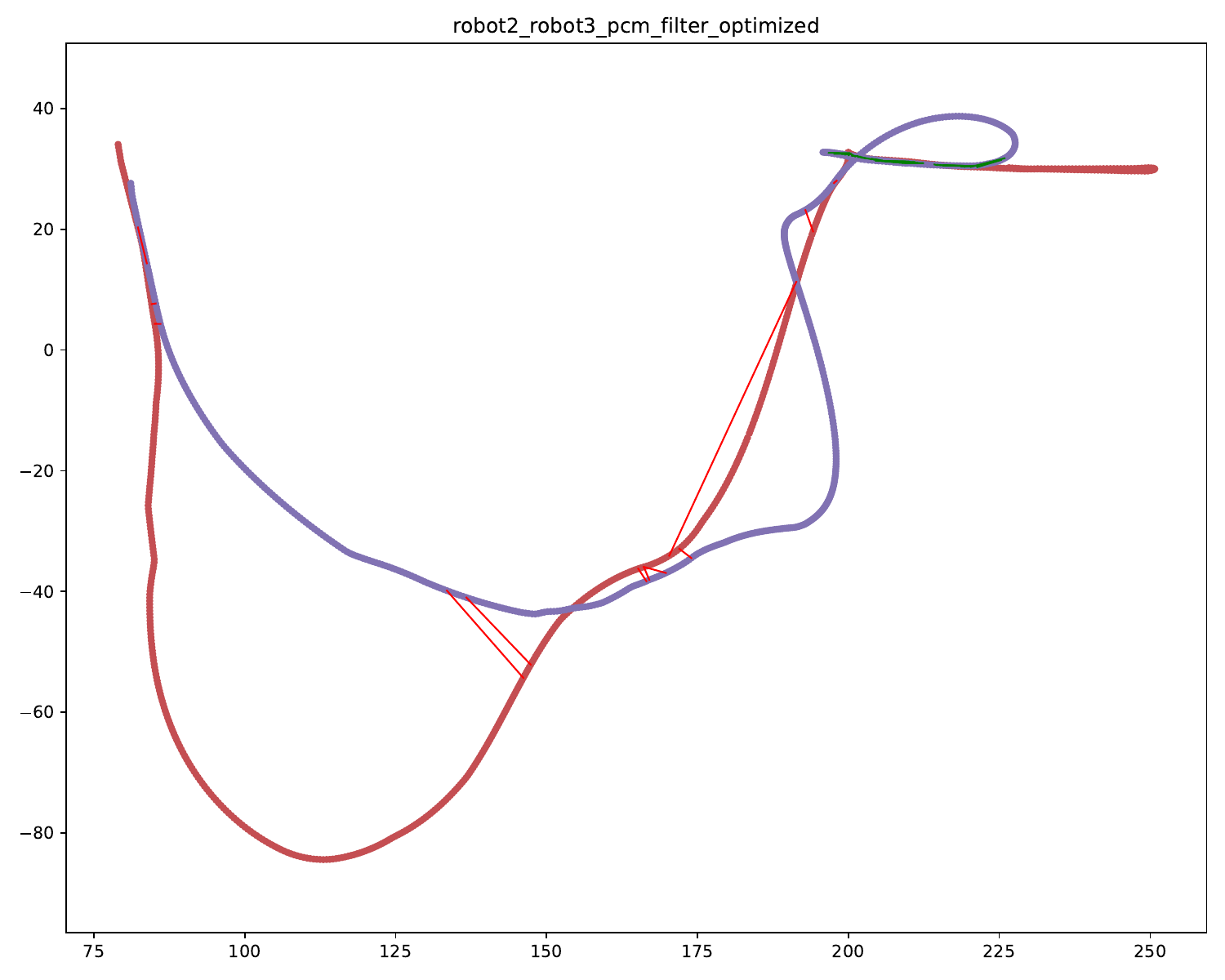}
        \par\smallskip
        \footnotesize\textbf{(c)} g2o graph with all KFs after optimization
    \end{minipage}

    % No tunnels
    \begin{minipage}[t]{0.32\textwidth}
        \centering
        \includegraphics[width=\textwidth]{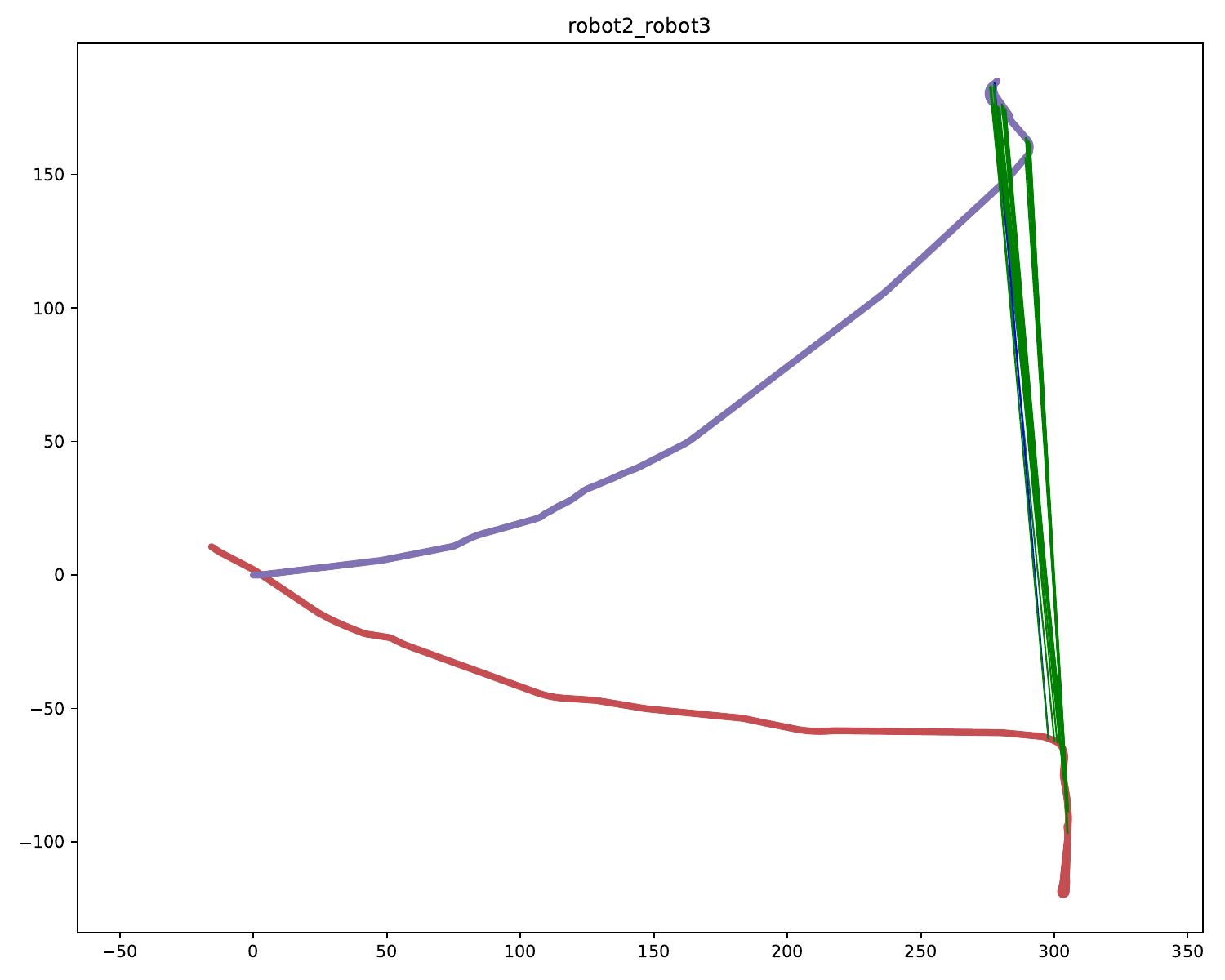}
        \par\smallskip
        \footnotesize\textbf{(d)} Initial g2o graph with tunnel-filtered KFs
    \end{minipage}
    \hfill
    \begin{minipage}[t]{0.32\textwidth}
        \centering
        \includegraphics[width=\textwidth]{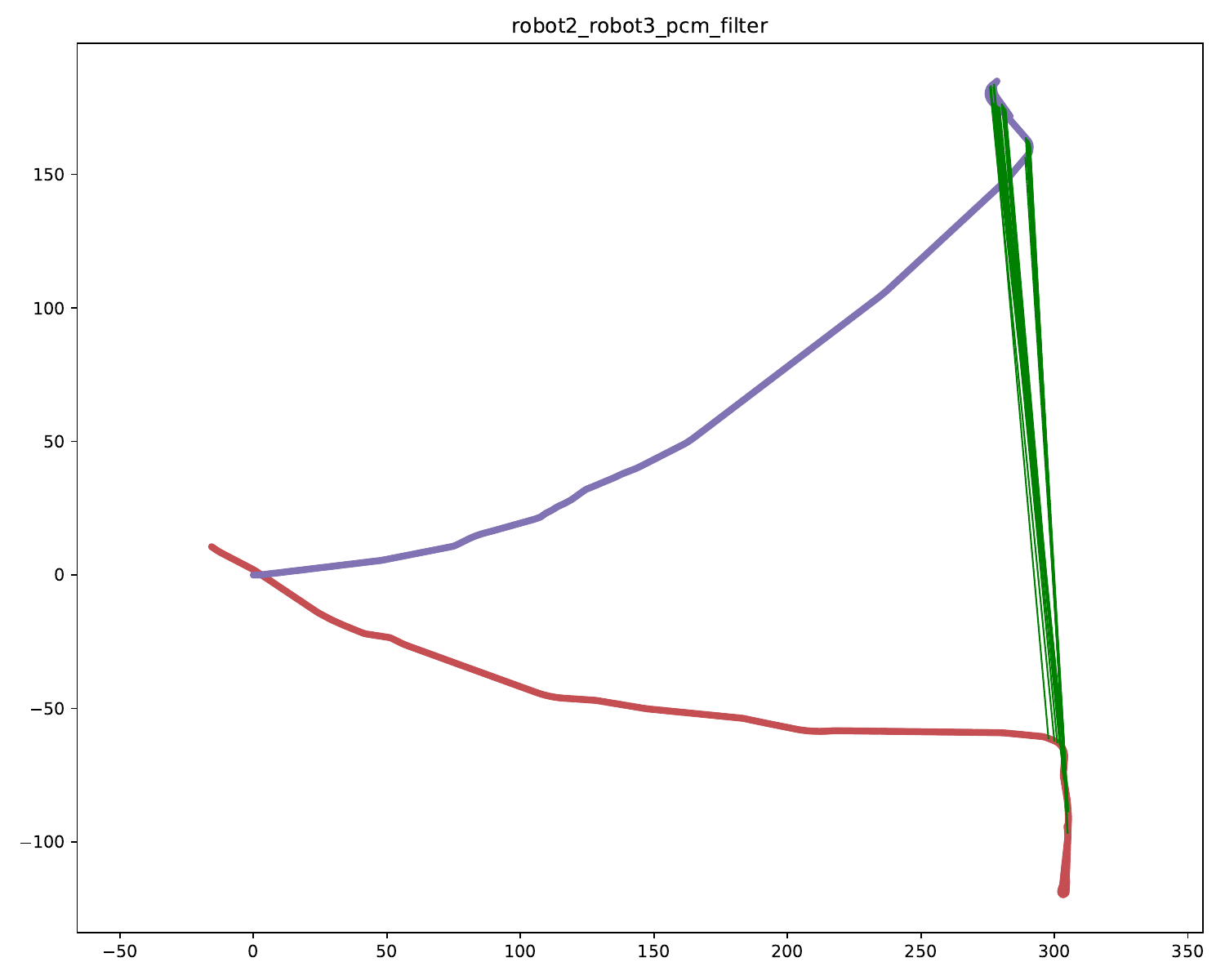}
        \par\smallskip
        \footnotesize\textbf{(e)} g2o graph with tunnel-filtered KFs after PCM
    \end{minipage}
    \hfill
    \begin{minipage}[t]{0.32\textwidth}
        \centering
        \includegraphics[width=\textwidth]{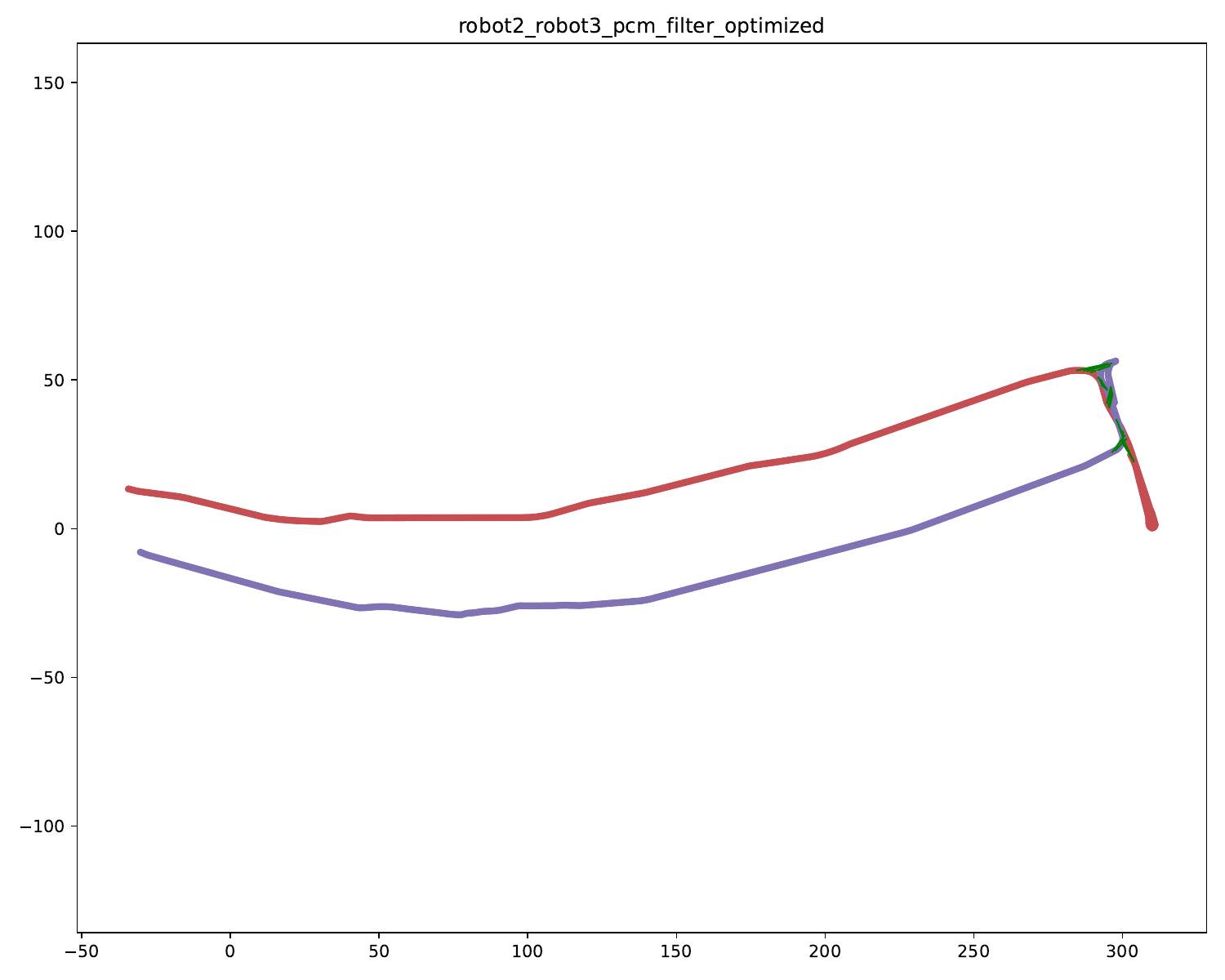}
        \par\smallskip
        \footnotesize\textbf{(f)} g2o graph with tunnel-filtered KFs after optimization
    \end{minipage}
    
    \caption{g2o graph visualization for the robot pair (2,3)}
\end{figure}

\appendix
\section[\appendixname~\thesection]{Trajectories visualization}

\begin{figure}[H]
    \centering
    \begin{minipage}[t]{0.49\textwidth}
        \centering
        \includegraphics[width=\textwidth]{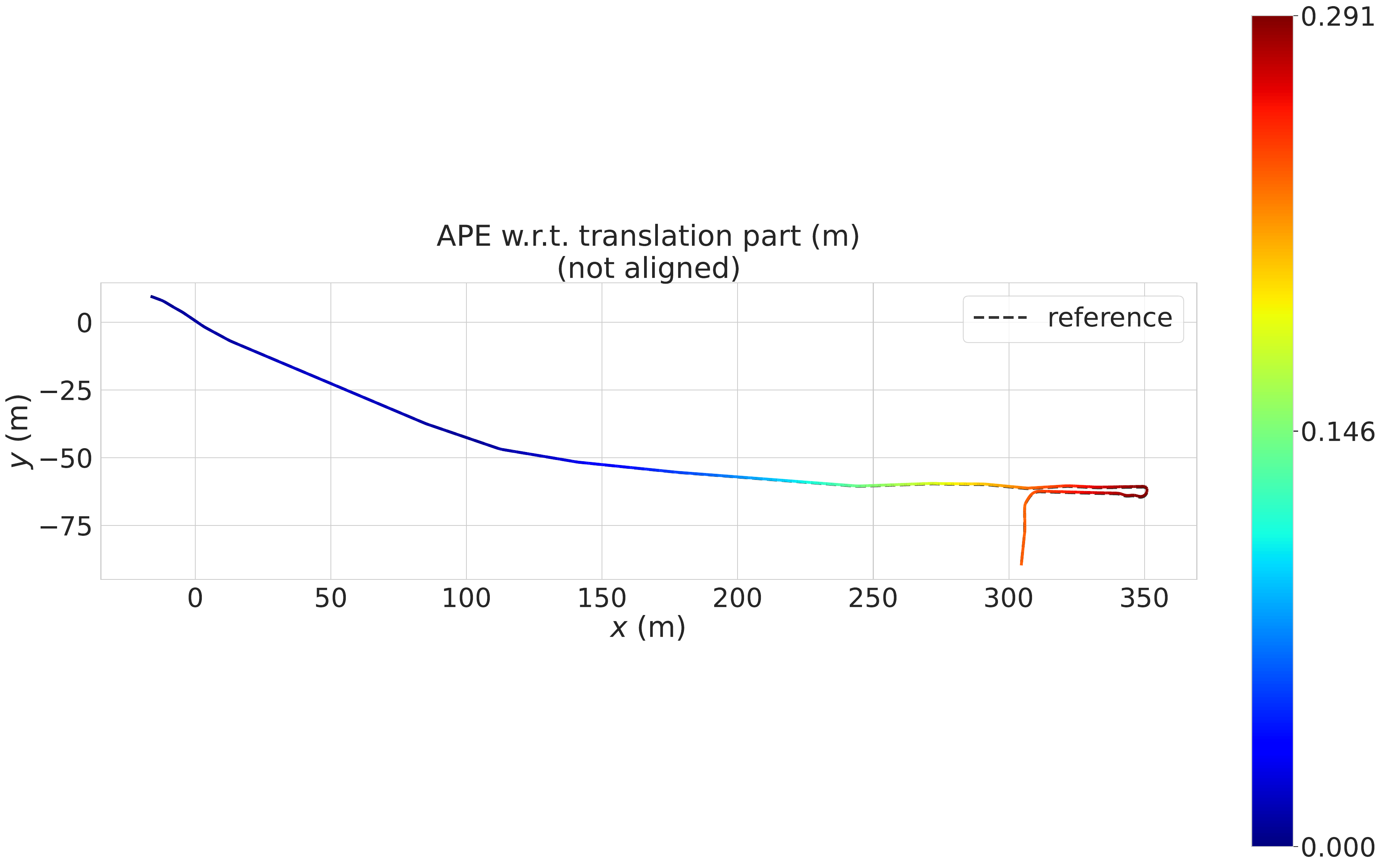}
    \end{minipage}
    \hfill
    \begin{minipage}[t]{0.49\textwidth}
        \centering
        \includegraphics[width=\textwidth]{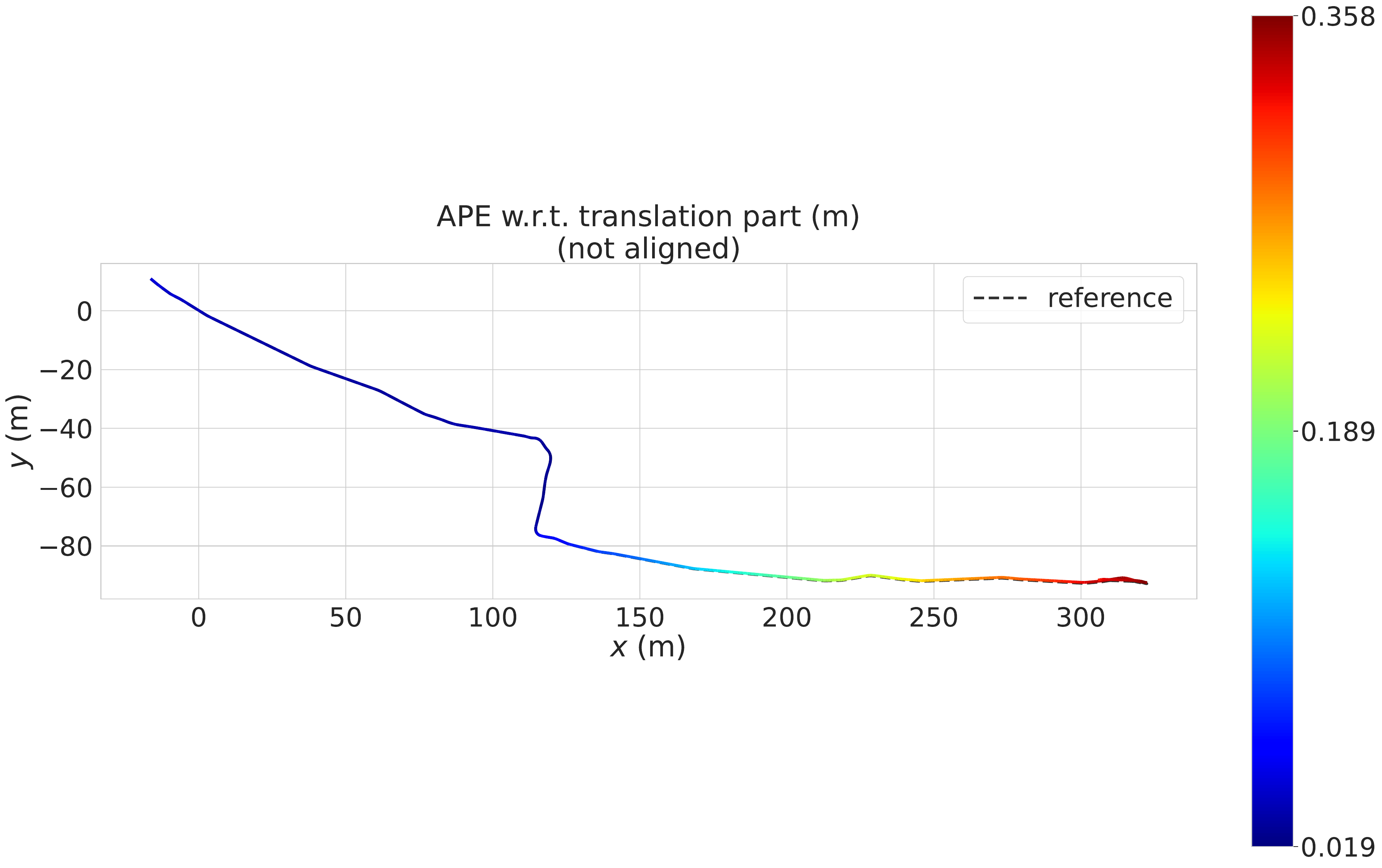}
    \end{minipage}
    \caption{Trajectories with ATE for the robot pair 0,1}
\end{figure}

\begin{figure}[H]
    \centering
    \begin{minipage}[t]{0.49\textwidth}
        \centering
        \includegraphics[width=\textwidth]{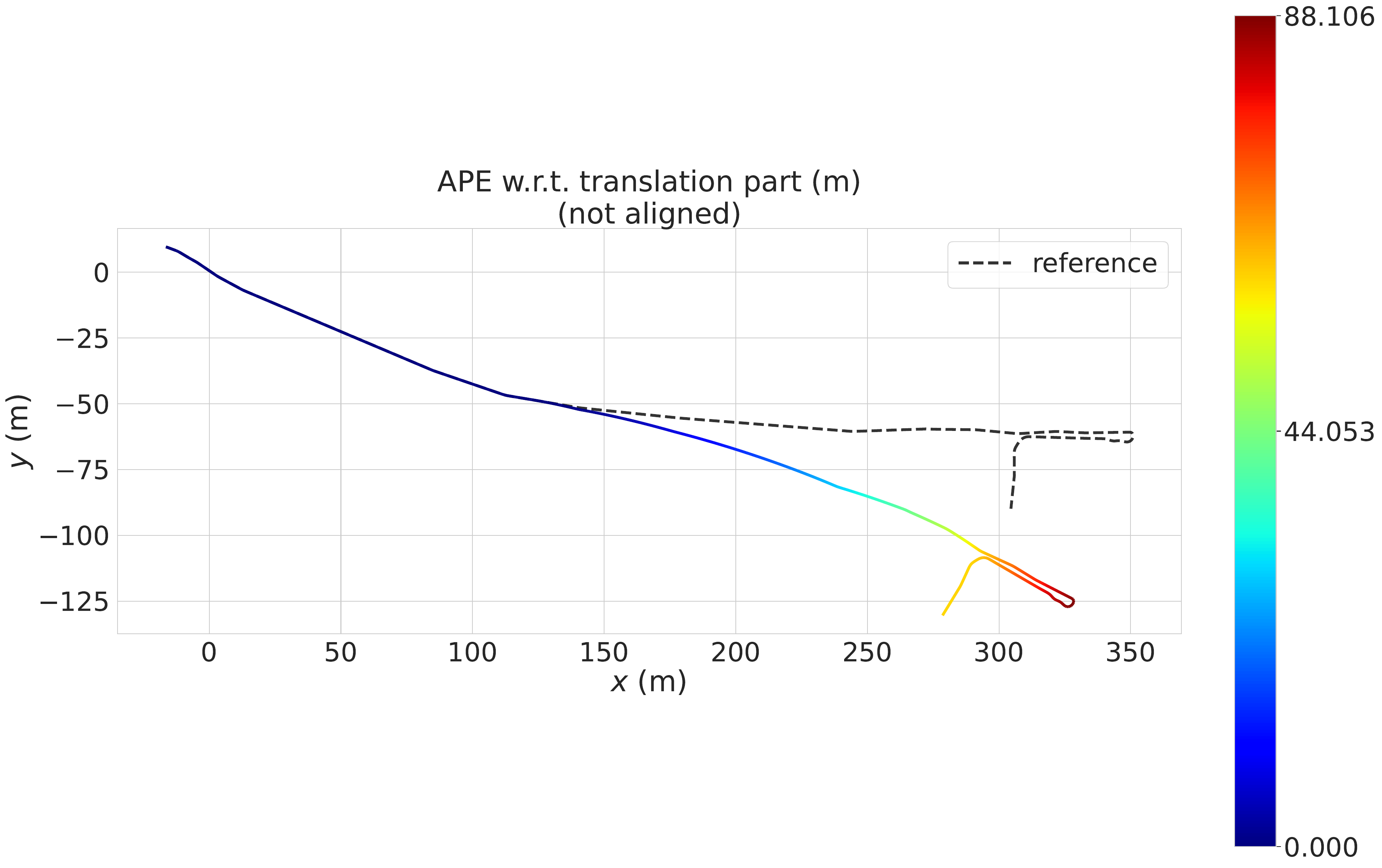}
    \end{minipage}
    \hfill
    \begin{minipage}[t]{0.49\textwidth}
        \centering
        \includegraphics[width=\textwidth]{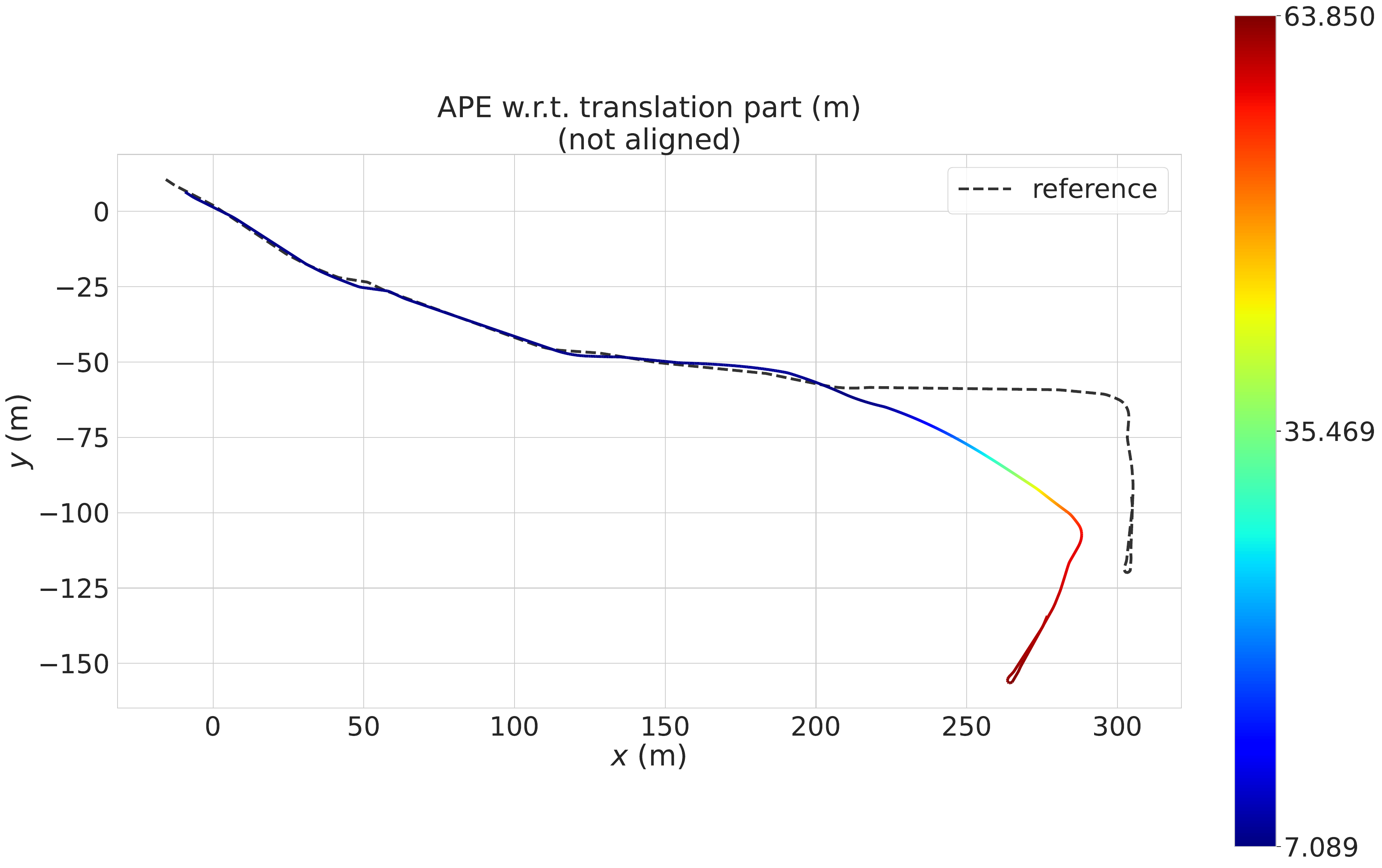}
    \end{minipage}
    \caption{Trajectories with ATE for the robot pair 0,2}
\end{figure}

\begin{figure}[H]
    \centering
    \begin{minipage}[t]{0.49\textwidth}
        \centering
        \includegraphics[width=\textwidth]{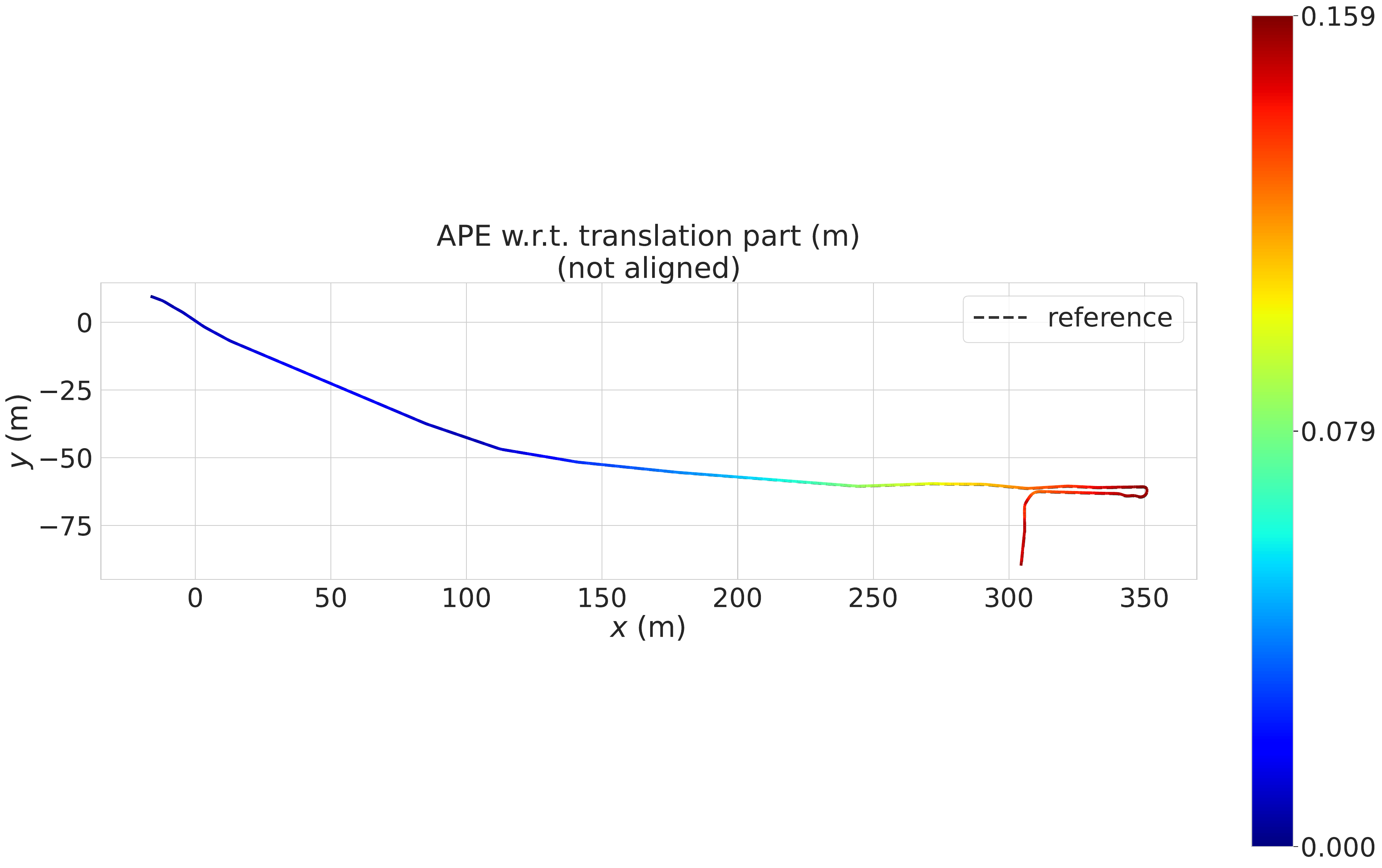}
    \end{minipage}
    \hfill
    \begin{minipage}[t]{0.49\textwidth}
        \centering
        \includegraphics[width=\textwidth]{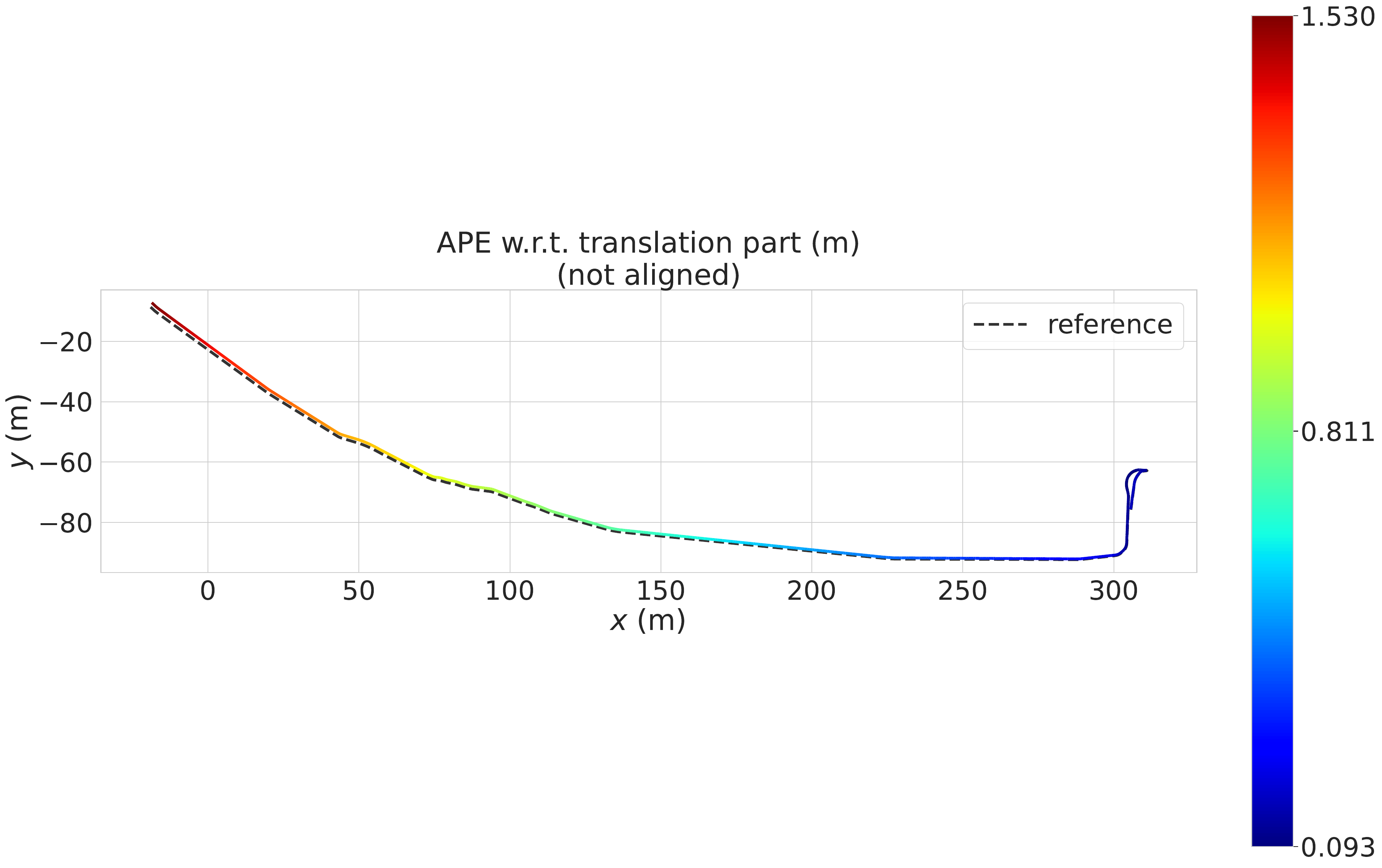}
    \end{minipage}
    \caption{Trajectories with ATE for the robot pair 0,3}
\end{figure}

\begin{figure}[H]
    \centering
    \begin{minipage}[t]{0.49\textwidth}
        \centering
        \includegraphics[width=\textwidth]{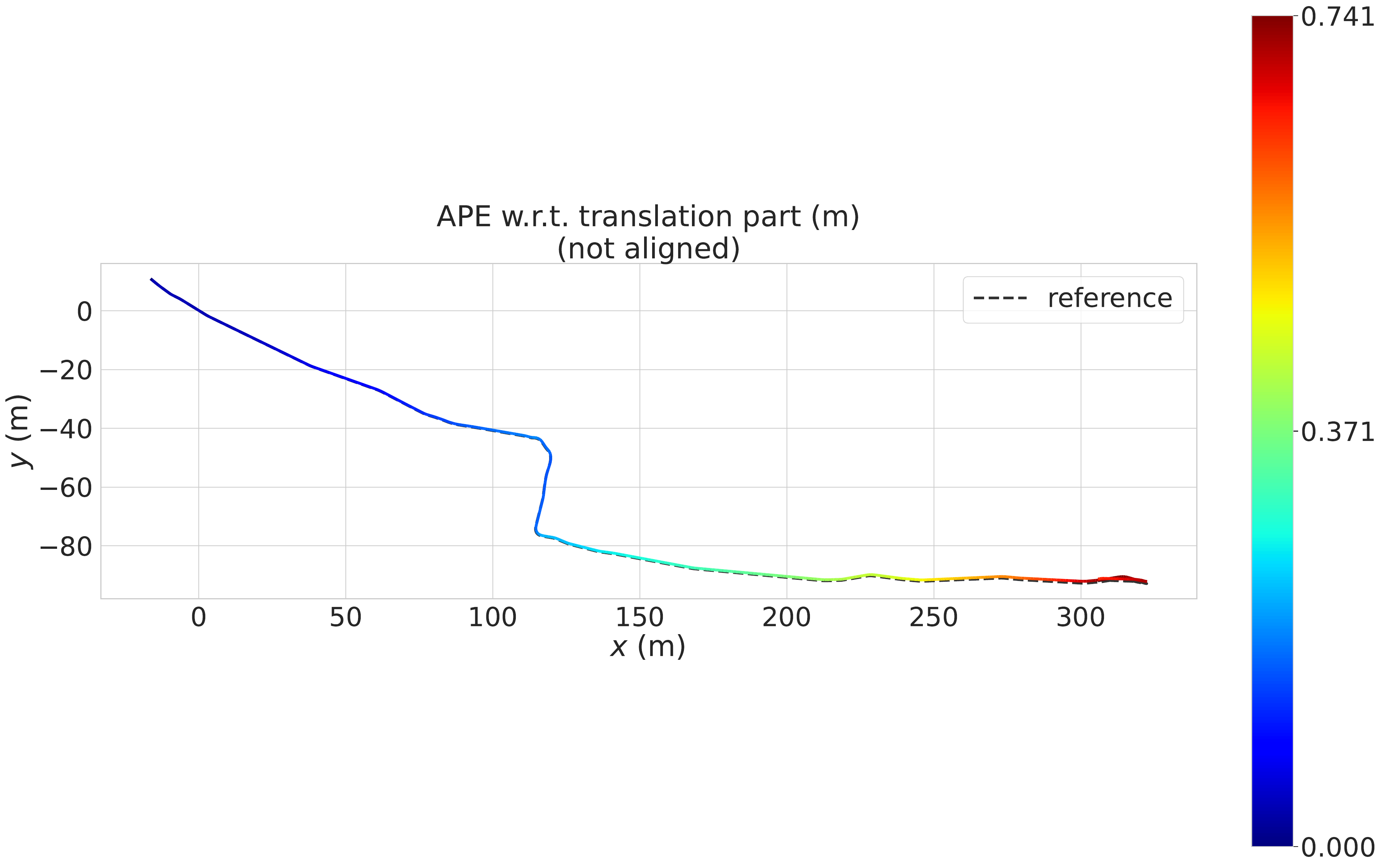}
    \end{minipage}
    \hfill
    \begin{minipage}[t]{0.49\textwidth}
        \centering
        \includegraphics[width=\textwidth]{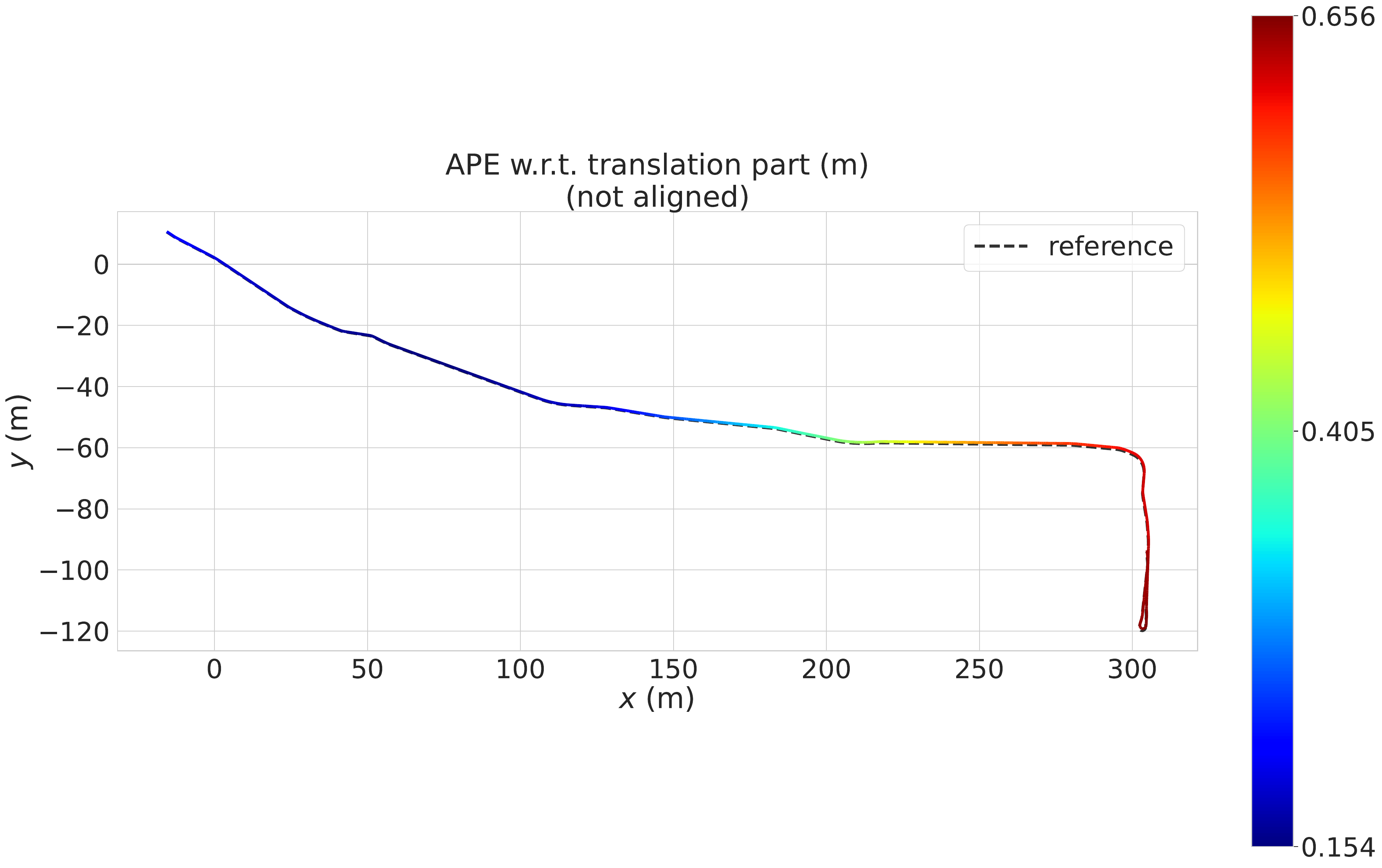}
    \end{minipage}
    \caption{Trajectories with ATE for the robot pair 1,2}
\end{figure}

\begin{figure}[H]
    \centering
    \begin{minipage}[t]{0.49\textwidth}
        \centering
        \includegraphics[width=\textwidth]{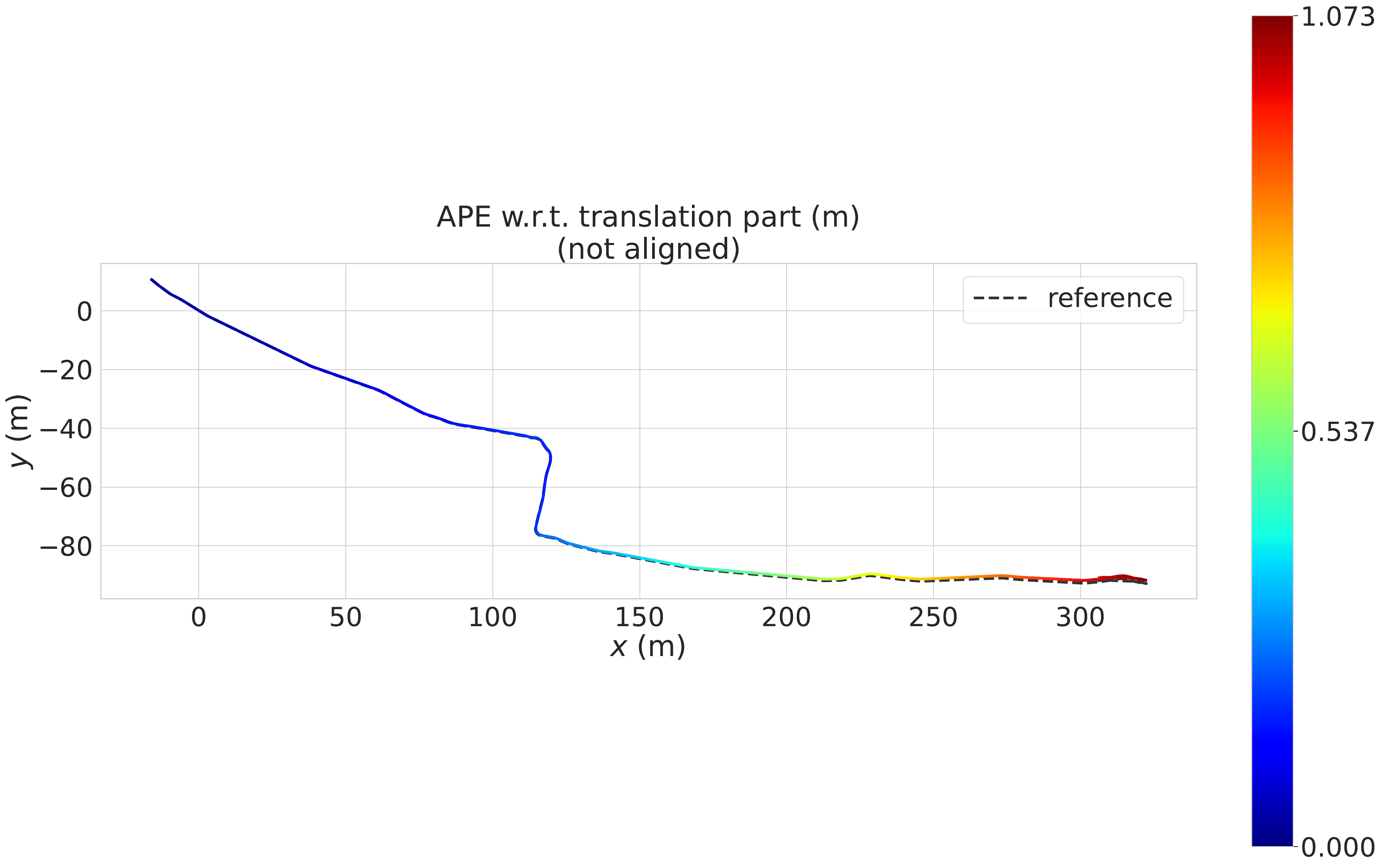}
    \end{minipage}
    \hfill
    \begin{minipage}[t]{0.49\textwidth}
        \centering
        \includegraphics[width=\textwidth]{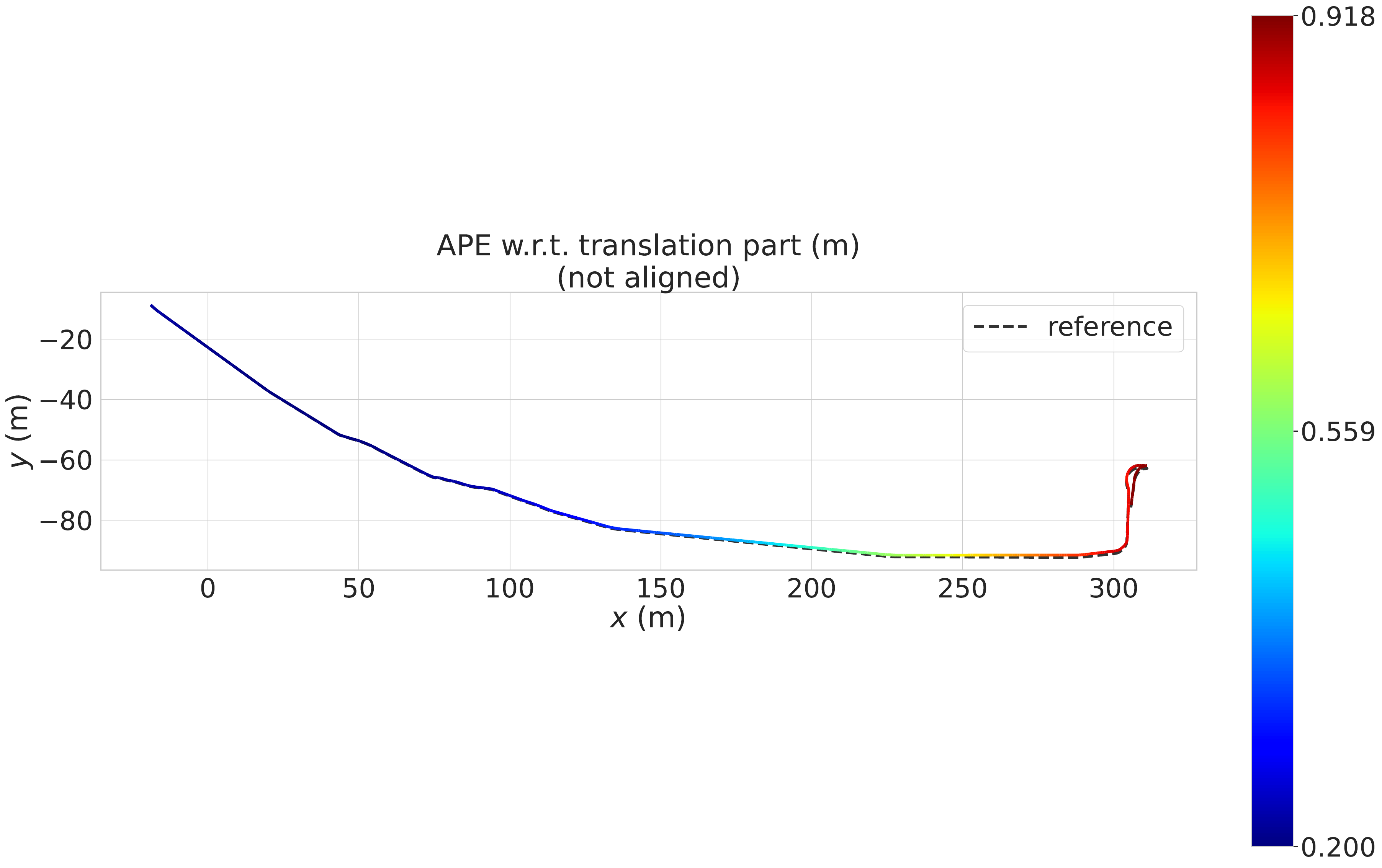}
    \end{minipage}
    \caption{Trajectories with ATE for the robot pair 1,3}
\end{figure}

\begin{figure}[H]
    \centering
    \begin{minipage}[t]{0.49\textwidth}
        \centering
        \includegraphics[width=\textwidth]{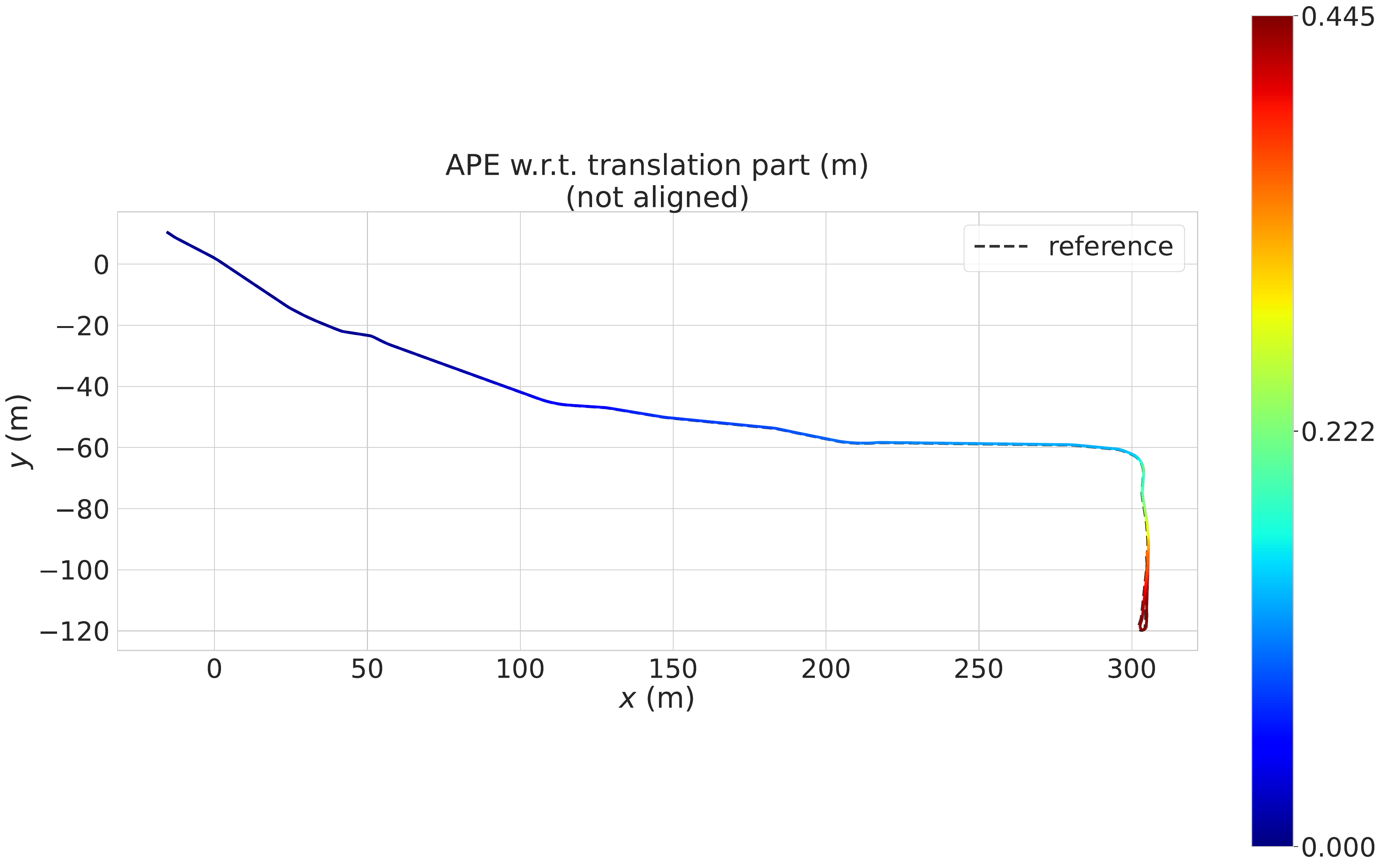}
    \end{minipage}
    \hfill
    \begin{minipage}[t]{0.49\textwidth}
        \centering
        \includegraphics[width=\textwidth]{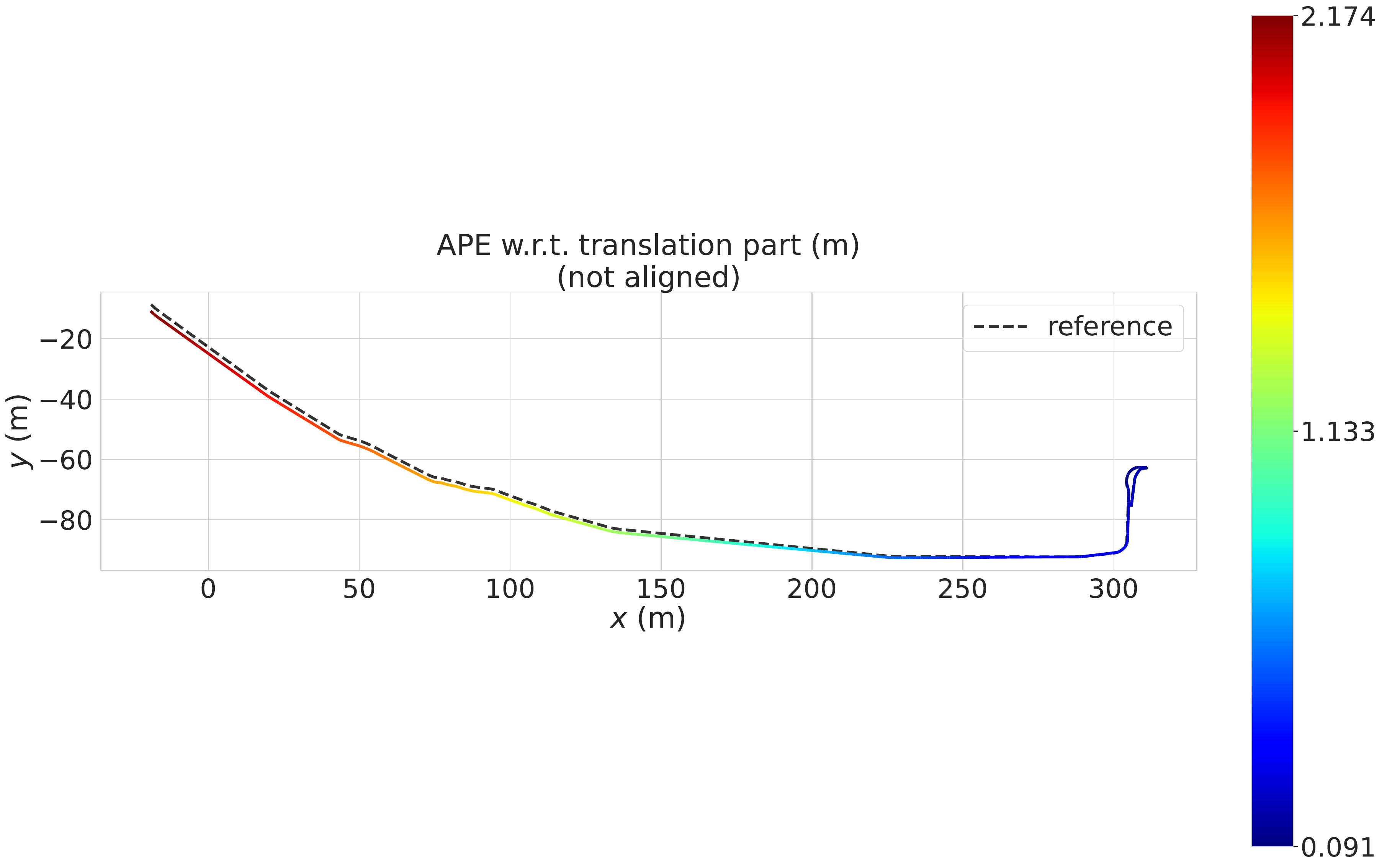}
    \end{minipage}
    \caption{Trajectories with ATE for the robot pair 2,3}
\end{figure}

%%%%%%%%%%%%%%%%%%%%%%%%%%%%%%%%%%%%%%%%%%
\isPreprints{}{% This command is only used for ``preprints''.
\begin{adjustwidth}{-\extralength}{0cm}
} % If the paper is ``preprints'', please uncomment this parenthesis.

\reftitle{References}

\bibliography{bibliography}

%\PublishersNote{}
\isPreprints{}{% This command is only used for ``preprints''.
\end{adjustwidth}
} % If the paper is ``preprints'', please uncomment this parenthesis.
\end{document}